\documentclass[manuscript,screen,a4paper,margin=1in]{acmart}

\AtBeginDocument{%
  }

\setcopyright{acmlicensed}
\copyrightyear{2025}
\acmYear{2025}

\usepackage{array}
\usepackage{booktabs}
\usepackage{enumitem} 
\usepackage{pifont}   
\usepackage{graphicx}     
\usepackage{adjustbox}    
\usepackage{colortbl}     
\usepackage{xcolor}       
\usepackage{multirow}
\usepackage{subcaption}

\usepackage{forest}
\usepackage{geometry} 
\usepackage[utf8]{inputenc}
\usepackage{tabularx}   
\usepackage{ragged2e}   

\newcolumntype{L}{>{\RaggedRight\arraybackslash}X}

\newcommand{\xmark}{\ding{55}} 

\DeclareUnicodeCharacter{00F9}{\`u} 
\DeclareUnicodeCharacter{00E1}{\'a} 
\DeclareUnicodeCharacter{1ECD}{\d{o}}
\DeclareUnicodeCharacter{0300}{\`}
\DeclareUnicodeCharacter{0301}{\'} 
\DeclareUnicodeCharacter{0198}{K} 
\DeclareUnicodeCharacter{1ECC}{\d{O}} 

\definecolor{highlightgreen}{RGB}{210, 245, 210}

\usepackage{hyperref}
\usepackage{caption}
\usepackage{tikz}
\usepackage{makecell}
\usepackage{threeparttable}
\usepackage{placeins}  
\usepackage{siunitx}    

\usepackage{fontawesome5}

\newcommand{\iconHuggingFace}{\textbf{HF}}       
\newcommand{\iconKaggle}{\faIcon{chart-bar}}        
\newcommand{\iconGitHub}{\faIcon{github}}
\newcommand{\iconLock}{\faIcon{lock}} 

\definecolor{excellent}{RGB}{0, 128, 0}
\definecolor{good}{RGB}{46, 160, 67}
\definecolor{fair}{RGB}{255, 127, 0}
\definecolor{poor}{RGB}{214, 39, 40}
\definecolor{missing}{RGB}{150, 150, 150}
\definecolor{critical}{RGB}{180, 0, 0}

\newcommand{\badgeExcellent}{\textcolor{excellent}{$\bigstar\bigstar$}}
\newcommand{\badgeGood}{\textcolor{good}{$\bigstar$}}
\newcommand{\badgeFair}{\textcolor{fair}{$\circ$}}
\newcommand{\badgePoor}{\textcolor{poor}{$\triangle$}}
\newcommand{\badgeMissing}{\textcolor{missing}{$\times$}}
 
\newcommand{\accessMissing}{\textcolor{missing}{$\times$}}
\newcommand{\accessPublic}{\textcolor{good}{\faIcon{github}}}

\hypersetup{
    colorlinks=true,
    urlcolor=blue,
    citecolor=blue,
}

\begin{document}


\title{NaijaNLP: A Survey of Nigerian Low-Resource Languages}

\author{Isa Inuwa-Dutse}
\affiliation{%
  \institution{University of Huddersfield}
  \country{United Kingdom}
}

\authorsaddresses{} 

\renewcommand{\shortauthors}{Inuwa-Dutse}


\begin{abstract} 
With over 500 languages in Nigeria, three languages - Hausa, Yorùbá and Igbo spoken by more than 175 million people, account for about 65\% of the languages. However, these languages are classed as low-resource due to insufficient digital resources to support tasks in computational linguistics. 
While several research efforts and initiatives have been presented, a coherent understanding of the state of classic Natural Language Processing (NLP) spanning grammatical formalisation to linguistic resources that support models development is lacking. 
This study presents the first comprehensive review of the state of affairs in NLP research across the three major Nigerian languages (NaijaNLP).  
We quantitatively assess the available linguistic resources and identify key challenges. 
Of the 293 reviewed studies, $27.6\%$ contributed new linguistic resources.  
This finding highlights a strong reliance on repurposing existing data rather than creating new resources. Also, language-specific challenges, such as morphological analysis and effective representation of diacritics, remain under-explored. To advance NaijaNLP and LR-NLP more broadly, we echo the need for more collaborative efforts in resource enrichment, comprehensive annotation, and increased community support.

\end{abstract}






\maketitle       

\section{Introduction}
\label{sec:introduction} 

Of the 7,170 distinct languages spoken globally \cite{globallang2025}, over 90\% are classified as low-resource (LR) from a computational stand. 
While there is no universal consensus on what constitutes a low-resource language \cite{nigatu2024zeno}, a language is generally considered low-resource if it suffers from limited parallel source-target data \cite{ortega2021love}. 
In other words, low-resource languages (LRLs) are languages for which statistical methods cannot be directly applied due to insufficient linguistic resources \cite{magueresse2020low}. 
Essentially, the term low-resource encompasses a broad spectrum of resource conditions and application domains \cite{duong2017natural,hedderich2020survey,yazar2023low,avetisyan2023large,krasadakis2024survey,maddu2024survey,masethe2024word}. 
Numerous studies have explored low-resource natural language processing (LR-NLP) across diverse domains, covering misinformation detection \cite{wang2024monolingual}, security \cite{teze2024future}, neural machine translation \cite{andrabi2021review,shi2022low}, cyberbullying detection \cite{mahmud2023cyberbullying}, and sentiment analysis \cite{ghafoor2021impact,mabokela2022multilingual,girija2023analysis,hughes2023bhattacharya_lab,yusuf2024sentiment,raychawdhary2025empowering}. 
A common denominator across these tasks is that LRLs suffer from limited linguistic resources leading to suboptimal performance in NLP downstream applications \cite{joshi2020state}.  
Low-resource languages are often treated as a homogeneous group, overlooking key variations in morphology, diacritics, data availability, linguistic tools and computational support. The binary classification of differentiating languages as merely low-resource or high-resource has been a dominant approach in NLP research \cite{hedderich2020survey,yazar2023low,krasadakis2024survey,maddu2024survey,masethe2024word}. 
However, we argue that this dichotomous stance may impede progress in language enrichment as it fails to account for the unique challenges faced by individual or regional language groups. Instead, we advocate for a more nuanced approach that examines specific linguistic and regional contexts, identifies their distinct challenges and develops tailored solutions. Against this backdrop, our work is intended to provide a comprehensive review and assessment of NLP for Nigerian languages, particularly the three major languages (Hausa, Yorùbá, and Igbo), collectively referred to as NaijaNLP.

\subsection{Background} 
Nigeria\footnote{see \url{https://nigeria.gov.ng/} and \url{https://en.wikipedia.org/wiki/Nigeria}}, the most populous country in Africa, is home to three major languages\footnote{\url{https://en.wikipedia.org/wiki/Hausa_language}; \url{https://en.wikipedia.org/wiki/Yoruba_language}; \url{https://en.wikipedia.org/wiki/Igbo_language}} - Hausa, Yorùbá and Igbo - alongside numerous other languages and ethnic groups.  
Hausa, a Chadic language \cite{britannica2014}, is predominantly spoken in northern Nigeria and southern Niger but it is widely used across West and Central Africa, including Ghana, Cameroon, Benin, Togo, Chad, Ivory Coast, and Sudan. Ranked 20th globally in terms of number of speakers\footnote{\url{https://en.wikipedia.org/wiki/List_of_languages_by_total_number_of_speakers}}, Hausa serves as a lingua franca for non-native speakers across many African countries, making it one of the most influential Chadic languages.  
With over 50 million speakers, Yorùbá is a major Niger-Congo language that is widely spoken in Nigeria and other parts of Africa, especially the Western part. 
Similarly, Igbo is another influential language spoken by over 31 million people, mostly in southeastern Nigeria. As a tonal language, Igbo words can differ in meaning solely based on tone, with high and low tones marked by acute and grave accents, respectively \cite{goldsmith1976autosegmental,clark2011tonal}.  
At least one of these three major languages is spoken by nearly every Nigerian citizen, and together, they serve as primary communication means for over 200 million people worldwide. Despite their influence, population size, and possession of substantial corpora of undigitised literature, digital resources for these languages remain limited. 
In this study, we adopt the term NaijaNLP\footnote{Derived from Naija, a term based on the etymon Niger, which gave its name to the river and, subsequently Nigeria.} to refer to NLP research activities and tasks related to the aforementioned three major Nigerian languages. \citet{caron2020methodological} was our first encounter with the term \textit{Naija} utilised in the NLP context to describe research on Nigerian Pidgin.

\subsubsection{Scope} 

This study investigates the traditional or classic NLP landscape for NaijaNLP, with focus on linguistic resources and foundational methodologies. Therefore, our primary objective is to examine and document early NLP approaches, linguistic resources, and tools that have been developed to address the unique challenges faced by the three major Nigerian languages. While generative AI and related techniques are transforming NLP, they fall outside the scope of this work due to the need to first establish the early foundational progress in the classic NLP which is lacking for NaijaNLP. Instead, we offer limited coverage of applicable embedding techniques and pre-trained models pertinent to the review. To this end, we explore the NLP landscape of Nigerian languages focusing on the three major languages to understand their state and offer targeted recommendations that can facilitate the development of LR-NLP for these languages and beyond.

\subsubsection{Research Questions \& Contributions} 
To better understand the state of progress and identify factors hindering the development of LRLs, we examine existing linguistic resources, downstream tasks, and community support related to NaijaNLP.  
To proceed, we put forward the following set of research questions:   
\begin{enumerate}[label=\ding{\numexpr171+\value*}]
    \item \textbf{State of affairs:} To understand and assess the current landscape of NaijaNLP, we ask the following questions:
    \begin{itemize}
        \item[1.1] What linguistic and NLP resources are currently available for NaijaNLP?  
        \item[1.2] What efforts are underway to expand and enhance the resources in 1.1?  
    \end{itemize}
    \item \textbf{Challenges and Future Prospects:} To identify limiting factors and opportunities for the advancement of NaijaNLP, we examine the following questions: 
    \begin{itemize}
        \item[2.1] What challenges prevent NaijaNLP from fully leveraging state-of-the-art NLP developments?
        \item[2.2] What strategies can be implemented to overcome the challenges in 1.2 and drive progress? 
    \end{itemize} 
\end{enumerate} 

Our review and synthesis of the above research questions have led to the following key contributions: 
    \begin{enumerate}[label=\ding{\numexpr171+\value*}]
        \item \emph{Comprehensive Analysis of NaijaNLP:} We present the first comprehensive review of classic NLP research on Nigeria's major languages. We consolidate discussions on language particularities covering morphological analysis, diacritic restoration and language formalisation, which have thus far remained fragmented. Moreover, we provide a structured review of available linguistic resources, NLP tools, and initiatives driving the development of NaijaNLP. 

        \item \emph{Articulating Challenges and Strategic Recommendations:} Through a closer examination of the NaijaNLP's ecosystem, we identify its limitations, prevailing challenges, and opportunities. Based on this analysis, we proffer targeted recommendations to enhance the development and resource expansion for the 3 languages within the broader context of low-resource NLP. 

        \item \emph{Development of a Tracking Resource Hub:} Finally, we establish a dedicated Github repository\footnote{\url{https://github.com/ijdutse/naija-nlp}} to track the progress of NaijaNLP, offering researchers a structured access to past and recent studies, linguistic resources, and community-driven initiatives to foster collaboration and sustained development. 
    \end{enumerate} 

The remainder of this paper is structured as follows: ~\S ~\ref{sec:methodology} details the methodology adopted in this study and ~\S ~\ref{sec:literature-review} covers a comprehensive review of related work in NaijaNLP. Our findings and recommendations are presented in \S~\ref{sec:results-discussion}, and \S~\ref{sec:conclusion} concludes the review.

\section{Methodology} 
\label{sec:methodology} 
\begin{figure} 
        \centering
        \includegraphics[width=\linewidth]{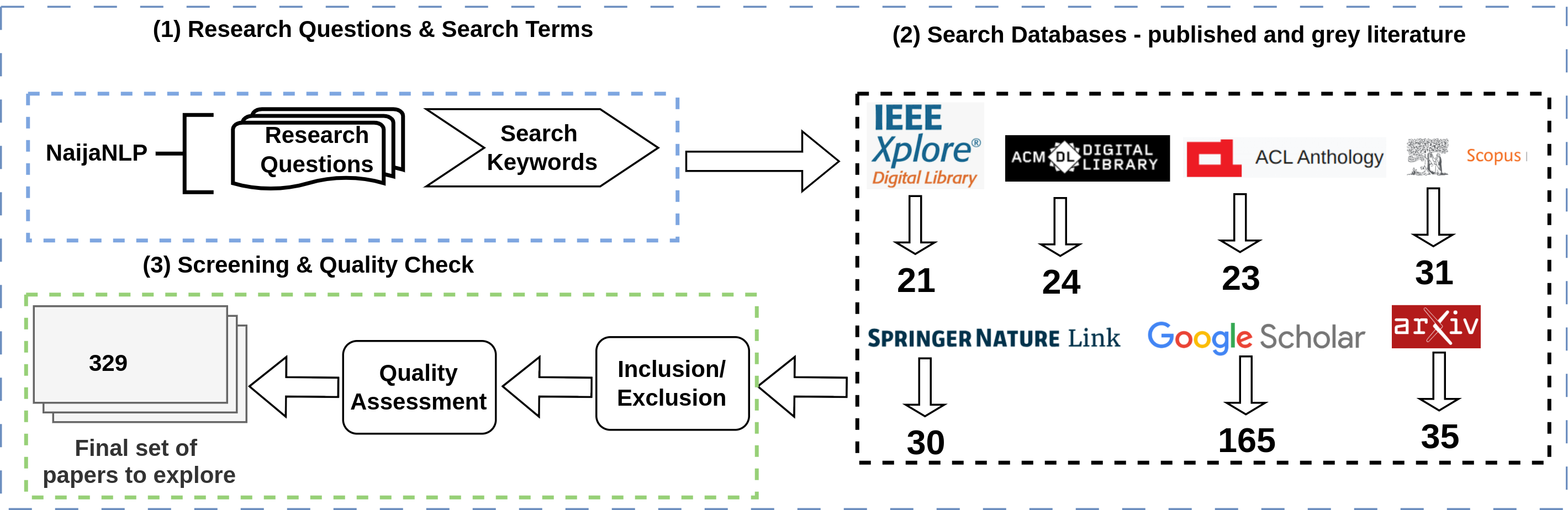}
        \caption{Overview of the review process, highlighting (1) guiding questions and search keywords (2) search databases and corresponding retrieved results, and (3) the screening process. Variations in the final article counts are due to overlapping entries retrieved from Google's search engine and other academic databases.} 
        \label{fig:literature-search}
    \end{figure}

This review explores the landscape of low-resource NLP research, with a specific focus on NaijaNLP to (1) examine existing NLP literature and linguistic resources (2) identify factors affecting linguistic resource availability and challenges (3) offer strategic recommendations for future enhancement of NaijaNLP. 
Figure~\ref{fig:literature-search} depicts an overview of the review process highlighting the research questions, search terms, consulted databases, and the screening process.

\subsection{Databases and Search Strategies}  
Our search strategy was conducted across several academic databases\footnote{comprising of \url{https://ieeexplore.ieee.org/}; \url{https://dl.acm.org/}; \url{https://www.scopus.com/}; \url{https://link.springer.com/}; \url{https://aclanthology.org/}} that include IEEE Xplore, ACM Digital Library, Scopus/Elsevier, Springer Nature\, and ACL Anthology. Additionally, grey literature\footnote{\url{https://arxiv.org/} and \url{https://scholar.google.com}} including arXiv preprint server and relevant articles accessible via Google Scholar, were incorporated into the review.  
To further expand the literature base, we employed a snowballing search strategy that iteratively identifies relevant studies through both backward (tracing references cited in key papers) and forward (identifying subsequent papers that cite the key studies) techniques. 
To ensure a comprehensive retrieval of relevant papers, we incorporated a permutation of search terms across three key categories: 
\begin{enumerate}[label=\ding{\numexpr171+\value*}]
    \item \textbf{Language-related:} These terms include Hausa, Yorùbá, and Igbo languages 
    \item \textbf{NLP-related:} This category covers terms such as low-resource languages, natural language processing (NLP), statistical modelling, rule-based, machine learning (ML), deep neural network (DNN), artificial neural network (ANN), artificial intelligence (AI), morphological analysis and diacritic restoration.  
    \item \textbf{Task-specific:} This set encompasses downstream tasks such as text classification \& summarisation, sentiment analysis, machine translation, speech recognition, part-of-speech (POS) tagging , and named-entity recognition (NER). 
\end{enumerate}  
An expanded list of the integrated search terms is provided in Table~\ref{tab:search-terms} in the appendix section. 

\emph{Inclusion and Exclusion Criteria}

In addition to high-impact publications, the selection criteria include relevant academic theses and non-seminal publications originating locally. We surmise that excluding these NaijaNLP sources could result in the omission of linguistic nuances, valuable resources and insights beneficial to the research community. Specifically, studies are excluded if: (1) they do not cover or pertain to NaijaNLP, i.e. the study falls outside the scope of the three major Nigerian languages as outlined earlier (2) they focus on languages with substantial resources, unless they explicitly address the targeted languages. 
We perform a title and abstract screening to identify studies that align with the guiding research questions. This is further reinforced by a thorough final review of the selected papers. 

\subsubsection{Preliminary Analysis} 
\begin{figure}
    \centering
    \includegraphics[width=0.85\linewidth]{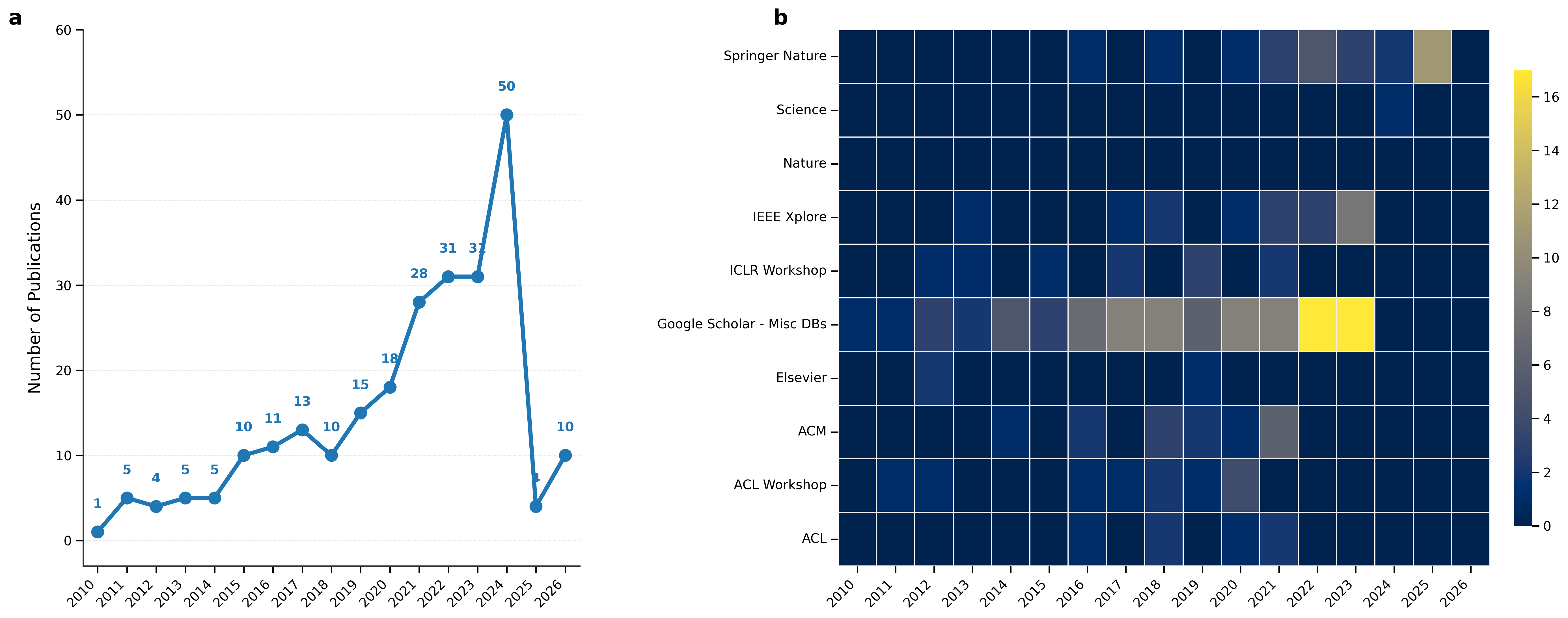}
    \caption{Aggregated publication count over time and key venues for NaijaNLP research. The trend in (a) is from the initial search before the snowbowling search. The earlier period with sparse data (1980-2011) has been truncated for clarity (b) Distribution of publication counts across primary venues and years.} 
    \label{fig:publication-trend-venues}
\end{figure}

Our initial analysis revealed significant overlap in the retrieved results, which we attribute to the frequent co-occurrence of at least 2 of the languages in the search queries.  
As a result, the total number of retrieved studies may not directly reflect the number of relevant studies specifically addressing each language. Furthermore, an examination of major academic databases highlighted a scarcity or complete absence of NaijaNLP research from high-impact venues such as Nature, NeurIPS, the main tracks of ICLR, ACL, and AAAI (see Figure~\ref{fig:publication-trend-venues}b). However, the increasing volume of NaijaNLP research output reflected in Figure~\ref{fig:publication-trend-venues}a, particularly from 2020 onward, signals a promising rise in contributions to linguistic resources and advancements in the field.

\section{Literature Review}
\label{sec:literature-review}
As a communication system, natural language comprises of structured sequences of sounds, symbols, or gestures that convey meaning in various forms. From a formalisation standpoint, languages are composed of structured strings of lexical categories (pre-terminals), forming well-defined sentences that facilitate statistical analysis of their structure \cite{culy1996formal, gunther2022language}. Progress in Machine Learning (ML) and the broader field of 
Artificial Intelligence (AI) has provided powerful tools for processing languages through NLP techniques. However, the effectiveness of NLP models depends on the availability of extensive and relevant linguistic resources, leading to a fundamental divide between high-resource and low-resource languages. High-resource languages (HRLs), such as English, French, and Spanish, benefit from sufficient digital resources and large-scale initiatives that facilitate access to standardised multilingual text resources in compatible formats \cite{graff1994multilingual,cho2014properties}. These resources, sourced from governmental bodies, international organisations, news agencies, and publishers, play a crucial role in propelling various NLP applications. 
Efforts to enrich and digitise LRLs, such as transducing texts \cite{randell1998hausar}, resource enrichment \cite{adegbola2009building}, and digitising indigenous poetry \cite{hausapoetry} are essential for their integration into computational frameworks. 

\subsubsection{Past Reviews} 
As previously noted, advancements in NLP have predominantly benefited high-resource languages, while progress in LR-NLP continues to face unique challenges owing to limited linguistic resources to drive downstream applications. Several past studies spanning language enhancement initiatives \cite{zhang2017embracing,adebara2022afrolid,robertson2022understanding,costa2022no,fan2021beyond,zhou2024character,wali2026automatic}, data augmentation and sentiment analysis \cite{fadaee2017data,raychawdhary2024optimizing,muhammad2022naijasenti} have examined the broader state of LR-NLP, highlighting these challenges to the development of low-resource languages. 
Similarly, numerous studies have investigated early NLP research involving the three major languages in Nigeria. However, these studies tend to focus on either a single language such as Hausa \cite{zakari2021systematic,muhammad2025hausanlp}, Yorùbá \cite{yusof2013review}, Igbo \cite{nwankwegu2021leveraging,maryann2021machine,orji2024igbo} or they concentrated on specific NLP tasks like text summarisation \cite{edwards2023text}, sentiment analysis \cite{raychawdhary2024enhancing,muhammad2022naijasenti}, machine translation \cite{benito2022machine,haddow2022survey}, and automatic speech recognition for tonal languages \cite{kaur2021automatic}. Other reviews adopted a broader spectrum by covering various African languages \cite{caron2020methodological,hedderich2020survey,asubiaro2021state,asubiaro2021evaluating,adelani2022natural,obiajuludigital,emmanuelcurrent,ezugwu2023machine,abdou2024review,hu2024review,issaka2026african}. While these studies are insightful, they provide a fragmented perspective on the current state of classic NLP research in the targeted languages. Our review builds upon previous studies by incorporating a wider scope that consolidates early contributions, open-access resources, and community-driven supports for low-resource languages. Thus, we offer a comprehensive analysis of the NaijaNLP landscape, examining past research efforts, available datasets, tools, language communities, and initiatives aimed at enhancing LR-NLP development. The subsequent discussions involving the three languages are organised alphabetically by language, following the order: Hausa, Igbo, Yorùbá.

\subsection{Formal Grammar and Knowledge Representation} 
\label{sec:grammar-and-knowledge-representation}
Formal grammar and knowledge representation provide crucial foundation for digital language resource development by offering structured frameworks for language processing. Computational grammar facilitates language enrichment through systematic analyses of syntax, semantics and linguistic structures, thereby enabling the creation of reusable linguistic resources for LRLs \cite{crysmann2012hag}. 
Recognising their fundamental role in language understanding and development of robust NLP tools, we begin our review with discussion on formal grammar and knowledge representation. 
In the context of NaijaNLP, relevant studies have explored grammatical formalisation \cite{abdoulaye1992aspects,culy1996formal,crysmann2009autosegmental,copestake2002implementing,crysmann2012hag,blasi2017grammars,umaraspects}, morphological analysis \cite{finkel2009computational,eludiora2018computational}, and knowledge representation \cite{eze1997aspects,aina2021ontology,ifeanyi2022semantic,aina2023human}. 
Notably, \citet{abdoulaye1992aspects} applied the Role and Reference Grammar framework\footnote{see \citet{van1980role}} to analyse sentence structure and linguistic processes in Hausa language.  
In a similar vein, \citet{crysmann2009autosegmental} demonstrated the utility of bidirectional Head-Driven Phrase Structure Grammar (HPSG) based on the Lingo Grammar Matrix \cite{bender2002grammar} to analyse vowel length, lexical and grammatical tone in Hausa. 
This effort also incorporates a Linguistic Knowledge Builder representation \cite{copestake2002implementing} that is inspired by Autosegmental Phonology \cite{crysmann2009autosegmental}. 
To facilitate the generation of grammatical structure, \citet{crysmann2012hag} implemented a formal grammar for Hausa, leading to the development of the HaG platform. Beyond these studies, \citet{blasi2017grammars} examined the transmission of grammatical features across 48 creole and 111 non-creole languages. Noting the challenge in maintaining logical consistency in knowledge representation, \citet{alade2019issues} investigated standardised knowledge representation for Yorùbá narratives in which imprecise codification of event and concept can lead to sub-optimal performance. 
In the context of Hausa NLP, \citet{abubakar2019hausa} developed Hausa WordNet, a lexical resource that extracts knowledge from Kamus (Hausa dictionary). 
Following the Lexical Markup Framework, \citet{enguehard2014computerization} offered a system for converting bilingual African language-French dictionaries (including Bambara, Hausa, Kanuri, Tamajaq, and Songhai-Zarma) from Word format to XML; such system is useful in enhancing lexicographic resources. 

Further discussions on morphological analysis and diacritic restoration, especially for Yorùbá and Igbo, are presented in \S~\ref{sec:language-particularity}.


\subsection{Language Particularities}  
\label{sec:language-particularity}

Languages exhibit distinct structural, phonetic, and orthographic characteristics that influence how they are written, pronounced, and processed. These particularities manifest in diverse orthographic systems comprising of alphabets, abjads, syllabaries, and logograms, which shape idea expression, word formation and written representation. Additionally, phonetic variations, such as differences in phonemes, stress patterns, and tonal systems, impact how languages are spoken and understood. Recognising and integrating these linguistic particularities and nuances are crucial for developing effective NLP tools that accurately represent the unique feature of each language \cite{asubiaro2015statistical,adebara2022towards,odoje201612}. 
Several Nigerian languages are tonal and employ characters beyond the standard Latin alphabet, rendering them structurally distinct from English and other high-resource languages. Consequently, failure to adequately account for these particularities will hinder the effective adaptation of existing NLP tools that have been developed primarily for English and other HRLs in the context of LRLs. Addressing this challenge, relevant studies have explored the linguistic diversity and specific needs of Nigeria's major languages \cite{amuda2010limited,oyinloye2015issues,abdulkareem2016yorcall,adewumi2020challenge,adebara2022towards}. This section expands on the existing studies on the linguistic particularities, especially morphological analysis and diacritic restoration.

\subsubsection{Morphological Analysis}
\label{sec:morphological-analysis}
Morphological analysis is concerned with examining word structure and formation, and has been documented in past studies \cite{iheanetu2017some,iheanetu2019addressing,chidiebere2020analysis,henry2020performance,frohmann2024segment}. To uncover deeper linguistic patterns, \citet{ochu2019corpus} leveraged concordance line strategies to extract multiword expressions in Igbo language. For effective representation of Igbo text, \citet{chidiebere2020analysis} examined collocation patterns, word order, compounding, and lexical ambiguity. 
In the context of Yorùbá language, \citet{oyinloye2015issues} developed a computational system (VerbMorpher) for the morphological analysis of standard Yorùbá verbs; the system incorporates linguistic resources applicable for spell-checking, syntax analysis, and dictionary functionalities. 
To enhance the effectiveness of morphological analysis, \citet{adegbola2016pattern} employed unsupervised induction approach to infer morphological rules for standard Yorùbá. Unlike conventional methods that often rely on word-segment frequency, the approach in \cite{adegbola2016pattern} is pattern-based that utilised word-pattern frequency. Although the study alluded to lexicons collected for the development of a Yorùbá speech synthesiser project, information about the data is lacking. For data curation and quality improvement, \citet{adewole2017token} applied a token validation system for automated corpus collection from online sources. The validation system is crucial in retaining key terms in the collection.  
Addressing structural complexities of Yorùbá in translation, \citet{adebara2021translating} detailed a strategy for handling language-specific morphological marking, particularly in cases where English features or cues are contextually inferred in Yorùbá. Motivated by the lack of machine-readable sense inventory for ambiguous Yorùbá words, \citet{adegokedevelopment} proposed a disambiguation component for a Yorùbá-English machine translation system. \citet{akinwonmi2024rule} introduced a declarative rule-based algorithm to address syllabification errors in Yorùbá, especially for consonant-vowel-nasal and diphthong-vowel-nasal structures. 

\subsubsection{Diacritic Restoration} 
\label{sec:diacritic-restoration}  
All the languages in NaijaNLP use diacritic marks to convey how a letter should be pronounced, however, the low-resource nature of the languages mean such marks are limited or missing. Thus, diacritic restoration involves adding or correcting diacritical marks such as accents, umlauts (two dots), and tildes associated with textual data. This restoration process is essential for accurate handling of historical texts and user-generated content where diacritics may have been omitted \cite{orife2020improving,adewumi2020challenge}. Given that many African languages utilise Latin-based scripts with diacritical marks, \citet{scannell2011statistical} propose a strategy and open-source package for automatic unicodification that converts ASCII text into its proper unicode form across multiple languages. 
In addressing orthographic and tonal diacritics, relevant studies leveraged n-gram models trained on the Igbo Bible corpus for word-level diacritic restoration \cite{ezeani2016automatic,ezeani2017lexical}, and embedding models for text similarity computation \cite{ezeani2018transferred,ezeani2018igbo}. Similarly, relevant studies documented optimal strategies for integrating diacritics and their effects on model performance for Yorùbá NLP tasks \cite{de2007automatic, tunji2011design, asahiah2017restoring, orife2018attentive, alabi2020massive, ogheneruemu2022development}. Recognising the impact of omitting tonal information in performance degradation, \citet{adegbola2012quantifying} assessed various machine learning models for handling tone-mark representation in Yorùbá text processing.   
For resource enhancement, \citet{adewumi2020challenge} introduced new analogy sets to aid evaluation and, \citet{adewole2020automatic} applied a rule-based framework to resolve the omission of tone or syllable in Yorùbá text. \citet{asahiah2023diacritic} developed a spell-checking and correction system that explicitly accounts for diacritic presence. In addressing diacritic usage and pronunciation challenges, \citet{abdulkareem2016yorcall} implemented a system (YorCALL) that is aimed at enhancing user proficiency in correctly applying diacritics in Yorùbá.  
The above studies on diacritic restoration, while beneficial in providing valuable insights and a range of methodology, they contributed limited linguistic resources, especially usable and accessible benchmark datasets.

\subsubsection{Numerical System} 
The challenges stemming from the lack of or limited standardised linguistic resources extend beyond textual to numerical data. A key area of interest is assessing whether NaijaNLP possesses the necessary linguistic resources, such as well-defined lexicons, for developing their numerical systems through published research. The earliest attempt at analysing Hausa numerals for machine processing dates back to the Automatic Text Comprehension project that surveyed and examined challenges related to converting spoken numerals into decimal representations \cite{sigurd1980numbers}. 
The project led to the development of an automated rule-based systems capable of bidirectional conversion between mathematical representations and their linguistic counterparts. 
Similarly, \citet{rhoda2017computational} developed a number-text conversion system for Igbo, contributing to the broader goal of numerical processing in LRLs. 
In a related effort, \citet{mustapha2023automated} formulated an algorithm for converting numerical figures into Hausa words, while \citet{agbeyangi2016web} introduced a numeral translation system for English-Yorùbá.  
Beyond numerical conversion, \citet{ajao2022yoruba} developed a Yorùbá handwritten character recognition to enhance effective linguistic digitisation. 
Despite these developments, research on numerical systems in NaijaNLP remains limited, highlighting the need for further studies and resource development that will ensure accurate numerical representation and processing across the languages.

\subsection{Linguistic Resources and Datasets} 
\label{sec:linguistic-resources-datasets} 

The development of relevant tools, resources, and techniques is essential for enabling various NLP downstream tasks and facilitating wider language coverage for low-resource NLP (LR-NLP) applications \cite{crysmann2009autosegmental}. On the other hand, insufficient linguistic resources and pre-processing tools presents a major challenge for LRLs when diacritic and morphological features must be considered. 
To address these challenges, several initiatives, resources, and tools have been developed to support and enrich LRLs. A case in point is the low-resource Human Language Technology (LoReHLT) project, which evaluates the DARPA\footnote{Defense Advanced Research Projects Agency (DARPA) - \url{https://www.darpa.mil/}} Low-Resource Languages for Emergent Incidents (LORELEI) research program \cite{christianson2018overview}. The LORELEI initiative includes both Hausa and Yorùbá, providing researchers with resource packages comprising of monolingual and parallel texts, a bilingual dictionary, text annotated with part-of-speech (POS) tags, named-entity-recognition (NER), and noun-phase-chunking annotations. Furthermore, language processing tools such as segmenters and morphological analysers, aimed at improving NLP capabilities for these languages have been included.

In analysing the available tools and resources for NaijaNLP, we focus on relevant research areas that closely align with the LORELEI research initiative \cite{christianson2018overview}. Additional research areas outside these categories will be considered necessary to ensure a comprehensive evaluation of available LR-NLP tools and resources. The following section provides a detailed review of available datasets for NaijaNLP.

\subsubsection{Hausa Datasets}
Past studies have documented useful language resources and strategies to support the development of Hausa NLP for various downstream tasks \cite{ibrahimnecat,inuwa2021first,mohammed2024lexicon,muhammad2025hausanlp}.  
For sentiment analysis, \citet{mohammed2024lexicon} developed a Hausa dataset comprising 14,663 instances (4,154 positive, 4,310 negative, and 6,199 neutral sentiments). The dataset was created using words and phrases from Hausa dictionary\footnote{see \textit{Kamus na Hausa zuwa Turanci} \cite{rigdon2017english}} and further expanded through data augmentation techniques. Alongside a customised stemming method to enhance model performance, \citet{shehu2024unveiling} contributed a set of sentiment lexicons in Hausa language. 
\citet{imam2022first} curated a dataset of approximately 2600 articles to support fake news detection in Hausa language and \citet{vargas2024hausahate} introduced an early collection of expert-annotated 2,000 Hausa comments from Facebook for hate speech detection. In similar efforts, \citet{zandam2023online} contributed a dataset of threatening lexicons from Twitter, and \citet{adam2024detection} curated a collection of offensive terms in Hausa languages. \citet{awwalu2021corpus} constructed a useful tagset for part-of-speech tagging in Hausa, while \citet{salifou2014design} implemented a spell-checking and correction tool for Hausa, which is accessible via LyTexEditor and includes an extension or add-on for \url{OpenOffice.org}. 

Tables~\ref{tab:contributed-hausa-datasets} and ~\ref{tab:summary-of-all-quality-ratings} summarise the respective Hausa datasets that have been introduced by various studies and the associated quality indicators. Downstream tasks such as sentiment analysis are well-represented, accounting for 36\% of the contributed and accessible resources. While a substantial proportion of the resources have been annotated (65\%), only 40\% of the studies reported inter-annotator agreement. Some studies such as \citet{parida2023havqa} and \citet{vargas2024hausahate} demonstrate commendable annotation and effective validation. Of the introduced datasets, about 60\% of resources have been made publicly available. 

\subsubsection{Igbo Dataset} 
Several studies have reported useful datasets to facilitate computational tasks in Igbo language. 
\cite{babatunde2026advanced} contributed a large scale annotated dataset consisting of 750,000 user expressions and responses for developing Chatbot systems in Igbo language. Native speakers and linguistic experts have been consulted in developing the corpus. 
\citet{chiomaweb} compiled a comprehensive set of Igbo words and their meanings, while \citet{nganga2020spoken} is focused on designing a methodological framework for creating digital dialectal dictionary that relies on a corpus of spoken-word and oral traditions in Igbo. In a different study, \citet{ajao2023recurrent} implemented an RNN-based handwritten character recognition system for Igbo using data from native speakers. 
\citet{onuora2024machine} and \citet{ana2024ai} reported the development of a collection of hate speech content, however, information about the data was not reported. Moreover, datasets for detecting cyberbullying and related content in Igbo and Pidgin English have been discussed in \cite{okoloegbomultilingual,okoloegbo2022multilingual}, however, access to such datasets is not publicly available. To aid with named-entity recognition, \citet{onyemaechi2023some} presented a database of 158 Chi-prefixed Igbo personal names, categorised into praise, thanksgiving, testimony, prayer, and declarative groups.

For the Igbo resources in Table~\ref{tab:contributed-igbo-datasets}, POS/NER coverage is detailed with relevant resources. However, only 20\% of the studies detailed the annotation and validation processes (see Table~\ref{tab:summary-of-all-quality-ratings}).  
Moreover, sufficient information about the associated annotator agreement and accessibility of the contributed resources will be useful in bias mitigation and enhancing reproducibility.


\FloatBarrier
\begin{table*}[!htb]
\centering
\small 
        \caption{An overview of Hausa NLP datasets and language resources. Quality ratings: \badgeExcellent{} rigorous; \badgeGood{} adequate; \badgeFair{} partial; \badgePoor{} limited; \badgeMissing{} not reported.}   
        \label{tab:contributed-hausa-datasets}
        \begin{threeparttable}
        \begin{tabular}{@{} l p{5.5cm} c c c c @{}}
        \toprule
        \multirow{2}{*}{\textbf{Task}} & \multirow{2}{*}{\textbf{Key Contribution}} & \multicolumn{3}{c}{\textbf{Quality Indicators}} & \multirow{2}{*}{\textbf{Access}} \\
        \cmidrule(lr){3-5}
        & & \textbf{Annot.} & \textbf{Valid.} & \textbf{IAA} & \\
        \midrule
        \multicolumn{6}{c}{\textbf{Sentiment Analysis}} \\
        
        Sentiment - \cite{rakhmanov2022sentiment} &  40,000 comments; MT-based translation from English; validated by skilled users & \badgeGood & \badgeGood & \badgeGood & \accessPublic \tnote{a} \\ 
        
        Sentiment - \cite{sani2022sentiment,abdullahi2024twitter} & Both utilised data from @BBCHAUSA; \cite{abdullahi2024twitter} is focused on shortened words for ambiguity resolution & \badgePoor & \badgeMissing & \badgeMissing & \iconLock \\ 
         
        Sentiment - \cite{mohammed2024lexicon} & Data augmented from English/Hausa dictionary; expert linguist validation  & \badgeExcellent & \badgeExcellent & \badgeGood & \iconGitHub \tnote{b} \\        
         Sentiment - \cite{shehu2024unveiling} & Lexicon dictionary for sentiment analysis; translated from English & \badgeFair & \badgePoor & \badgeMissing & \accessPublic \tnote{c} \\
         
         Sentiment - \cite{mohammed2023building}  & Lexicon-based resources for Hausa & \badgeGood & \badgeGood & \badgeFair & \accessPublic\tnote{d} \\
        
        \midrule
        \multicolumn{6}{c}{\textbf{POS / NER}} \\
        POS/NER -- \cite{abubakar2019hausa}  &  Hausa WordNet with pronunciation \& utilises close class categories; BUK dictionary-based & \badgeGood & \badgeGood & \badgeMissing & \iconLock \\ 
        
        POS -- \cite{tukurcorpus} &  developed a Hausa language POS tagset (HTS) consisting of 3000 words from various online sources & \badgeFair & \badgeGood & \badgeMissing & \accessPublic \tnote{j} \\ 
        
        \midrule
        \multicolumn{6}{c}{\textbf{Diacritic Restoration}} \\
        
        Diacritic -- \cite{abubakar2024development} & Voice data v15.0; 13h transcripts & \badgeFair & \badgeFair & \badgeMissing & \iconGitHub \tnote{e} \\ 
        
        \midrule
        \multicolumn{6}{c}{\textbf{Machine Translation / ASR}} \\
        MT -- \cite{akinfaderin2020hausamt} &  Hausa–English parallel corpus from JW300, Tanzil, Tatoeba, Wikimedia public datasets & \badgeMissing & \badgeMissing & \badgeMissing & \accessPublic \tnote{f} \\  
        \midrule
        \multicolumn{6}{c}{\textbf{Miscellaneous}} \\
        Visual -- \cite{abdulmumin2022hausa} & Hausa Visual Genome (HaVG); image descriptions & \badgeExcellent & \badgeExcellent & \badgeGood & \accessPublic \tnote{g} \\ 
        Offensive -- \cite{adam2024detection}  & Hausa offensive content detection; user study conducted & \badgeGood & \badgeGood & \badgeMissing & \iconLock \\ 
        Fake News -- \cite{imam2022first} & Selection from 6 topics; real/fake news classification & \badgeFair & \badgeFair & \badgeMissing & \iconLock \\
        Corpus -- \cite{inuwa2021first} &  Collection of formal/informal Hausa corpus; parallel data & \badgeGood & \badgeFair & \badgeMissing & \accessPublic \tnote{h} \\ 
        VQA -- \cite{parida2023havqa} & Data for visual QA; translated from English & \badgeExcellent & \badgeExcellent & \badgeExcellent & \iconLock \\ 
        Hate Speech -- \cite{vargas2024hausahate} &  Facebook comments; expert-annotated & \badgeExcellent & \badgeExcellent & \badgeGood & \accessPublic \tnote{i} \\ 

         Corpus -- \cite{wali2026automatic} &  400k sentence pairs; addressing textual anomalies in Hausa posts & \badgeGood & \badgeGood & \badgeMissing & \accessPublic \tnote{k} \\  

         Sexism -- \cite{muhammad2025dataset} &  translated from English; user study & \badgeGood & \badgeGood & \badgeGood & \accessPublic \tnote{l} \\

        Threat -- \cite{zandam2023online} & Threatening content from Twitter; detailed annotation & \badgeGood & \badgeGood & \badgeGood & \iconLock \\ 
        \addlinespace
        \multicolumn{6}{c}{\textbf{Cross-Language Tools}} \\
        Tools -- \cite{chiarcos2011information} &  Introduced ANNIS tool and resources for 25 African languages; unified annotation access & \badgeExcellent & \badgeExcellent & \badgeGood & \iconLock \\          
        \bottomrule
        \end{tabular}
        \begin{tablenotes}[flushleft]
        \footnotesize
        \item[] \textbf{Quality Indicators}: Annot. = Annotation rigour; Valid. = Validation process; IAA = Inter-Annotator Agreement reported. \textbf{Access}: \iconHuggingFace{} = HuggingFace; \iconKaggle{} = Kaggle; \iconGitHub{} = Github ;  ~\iconLock{} = Restricted/unavailable. 
        \tnote{a} \url{https://github.com/MrLachin/HESAC}
        \tnote{b} \url{https://drive.google.com/file/d/1OttyNIkxyfFYrWoIF2L_41J4TTR4KRcg/view?usp=sharing}
        \tnote{c} \url{https://github.com/harisushehu/Hausa-deep-sentiment-analysis}
        \tnote{d} \url{https://huggingface.co/datasets/mangaphd/hausaBERTdatatrain} 
        \tnote{e} \url{https://commonvoice.mozilla.org/en/datasets}
        \tnote{f} \url{https://github.com/WalePhenomenon/Hausa-NMT}
        \tnote{g} \url{http://hdl.handle.net/11234/1-4749}
        \tnote{h} \url{https://github.com/ijdutse/hausa-corpus}
        \tnote{i} \url{https://github.com/franciellevargas/HausaHate}    
        \tnote{j} The provided link in the study is broken. 
        \tnote{k} \url{https://github.com/ahmadmwali/HausaSeq2Seq}
        \tnote{l} \url{https://github.com/FatimaMuhammadAdam/Dataset-for-Sexism-Detection-in-Hausa-Language}
        \end{tablenotes}
        \end{threeparttable}
\end{table*}
\FloatBarrier


\FloatBarrier
\begin{table*}[!htb]
        \centering
        \small
        \caption{An overview of Igbo NLP resources. 
        Quality ratings: \badgeExcellent{} rigorous; \badgeGood{} adequate; \badgeFair{} partial; \badgePoor{} limited; \badgeMissing{} not reported.}  
        
        \label{tab:contributed-igbo-datasets}
        \begin{threeparttable}
        \begin{tabular}{@{} l p{5.5cm} c c c c @{}}
        \toprule
        
         \multirow{2}{*}{\textbf{Task \& Source}} & \multirow{2}{*}{\textbf{Key Contribution}} & \multicolumn{3}{c}{\textbf{Quality Indicators}} & \multirow{2}{*}{\textbf{Access}} \\
        \cmidrule(lr){3-5}
        
        & & \textbf{Annot.} & \textbf{Valid.} & \textbf{IAA} & \\
        \midrule
        \multicolumn{6}{c}{\textbf{Sentiment Analysis}} \\ 
        Sentiment - \cite{ogbuju2020development} & IgboSentilex: sentiment lexicons via Google Translate + manual correction & \badgePoor & \badgePoor & \badgeMissing & \iconLock \\ 

         \midrule
        \multicolumn{6}{c}{\textbf{POS / NER}} \\
        
        POS/NER - \cite{onyenwe2018basic} &  BLARK: 1M token corpus; 70-tag POS tagset; 300K tagged subcorpus & \badgeExcellent & \badgeExcellent & \badgeExcellent & \iconLock \\ 
 
         \midrule
        \multicolumn{6}{c}{\textbf{TC / TS}} \\
        Text Summarisation - \cite{mbonu2022igbosum1500} & IgboSum1500: translated news summaries; validated by native speakers & \badgeFair & \badgeFair & \badgeMissing & \iconLock \\

         \midrule
        \multicolumn{6}{c}{\textbf{Machine Translation / ASR}} \\
        Speech/ASR - \cite{ana2024ai,onuora2024machine} & Hate speech dataset; data creation is unclear & \badgePoor & \badgePoor & \badgeMissing & \iconLock \\ 

        Speech/ASR - \cite{babatunde2026advanced} & ASR \& sentiment data; 750k user expressions and responses & \badgeGood & \badgeGood & \badgeMissing & \iconLock \\  

        \midrule
        \multicolumn{6}{c}{\textbf{Miscellaneous}} \\

        Miscellaneous - \cite{chiomaweb} & Web-based Igbo thesaurus; the tool is based on previous work & \badgePoor & \badgeMissing & \badgeMissing & \iconLock \\
 
       Miscellaneous - \cite{ajao2023recurrent} & data collected via scanning handwritten Igbo alphabets; 3,600 images collected & \badgePoor & \badgeMissing & \badgeMissing & \iconLock \\

       Miscellaneous - \cite{okoloegbo2022multilingual} & 1800 collection of popular pidgin phrases and slangs; Igbo and Pidgin data translated to English & \badgeFair & \badgeMissing & \badgeMissing & \badgeMissing \\

         Miscellaneous - \cite{nganga2020spoken} & Audio databank for Igbo dialects; Igbo lexicographical work toproceed in keeping with international standards & \badgeFair & \badgeMissing & \badgeMissing & \badgeMissing \\  
       
        \bottomrule
        \end{tabular}
        \begin{tablenotes}[flushleft]
        \footnotesize 
        \item[] \textbf{Quality Indicators}: Annot. = Annotation rigour; Valid. = Validation process; IAA = Inter-Annotator Agreement. \textbf{Access}: \iconHuggingFace{} = HuggingFace; \iconKaggle{} = Kaggle; \iconGitHub{} = Github ; ~\iconLock{} = Restricted/unavailable.
        \end{tablenotes}
        \end{threeparttable}
    \end{table*}
\FloatBarrier


\subsubsection{Yoruba Datasets} 
For the Yorùbá language, researchers have contributed a range of linguistic resources and strategies to support various downstream applications \cite{orife2018attentive,ahia2024voices}. 
\citet{fagbolu2015digital} introduced one of the earliest corpora to support machine translation and other NLP downstream tasks. 
\citet{ademusire2023development} contributed a collection of annotated Yorùbá data for event extraction, and \citet{akpobi2024yankari} developed a large-scale monolingual Yorùbá dataset (Yankari). 
\citet{afolabi2013implementation} developed an audio database for Yorùbá consonant-vowel syllables, the alphabet, and a text-to-speech system. 
In a related work, \citet{iyanda2015statistical} compiled a Yorùbá corpus for speech generation, sourced from textbooks, newspapers, and online materials. 
Similarly, \citet{gutkin2020developing} contributed a speech synthesis dataset made of over 4 hours of 48 kHz recordings in Yorùbá, while \citet{akinwonmi2021development} contributed annotated prosodic read-speech syllabic data, extracted from fictional books and online scriptures. 
To further standardise Yorùbá linguistic resources, \citet{oluwatoyinstochastic} introduced YoTEx (Yoruba Text Lexicon), a repository that learns from English text to enhance Yorùbá text processing. \citet{okediya2019building} utilised various online resources consisting of books, blogs, social websites, and Yorùbá dictionaries, to construct a general-purpose lexical ontology for Yorùbá NLP applications.

A detailed breakdown of the contributed Yoruba datasets is presented in Table~\ref{tab:contributed-yoruba-datasets}.  
About 58\% of the studies reported detailed annotation of the contributed resources, however, only 25\% included the applicable inter-annotator agreement score. The validation process is relatively low compared to the overall datasets. Moreover, only 33\% of the reported resources are publicly accessible. Studies such as YORÙLECT \cite{ahia2024voices}, YOSM \cite{shode2022yosm}, YANKARI \cite{akpobi2024yankari}, and \cite{adegbola2016pattern} set commendable quality standards with public access and clear annotation protocols. 

\FloatBarrier
\begin{table*}[!htb]
        \centering
        \small
        \caption{An overview of Yorùbá NLP datasets and resources contributed by past studies. 
        Quality ratings: \badgeExcellent{} rigorous; \badgeGood{} adequate; \badgeFair{} partial; \badgePoor{} limited; \badgeMissing{} not reported.} 
        
        \label{tab:contributed-yoruba-datasets}
        \begin{threeparttable}
        \begin{tabular}{@{} l p{6.1cm} c c c c @{}}
        \toprule
        \multirow{2}{*}{\textbf{Task \& Source}} & \multirow{2}{*}{\textbf{Key Contribution}} & \multicolumn{3}{c}{\textbf{Quality Indicators}} & \multirow{2}{*}{\textbf{Access}} \\
        \cmidrule(lr){3-5}
        & & \textbf{Annot.} & \textbf{Valid.} & \textbf{IAA} & \\
        \midrule
        \multicolumn{6}{c}{\textbf{MA / DR}} \\ 
        MA/DR\tnote{a} - \cite{adebara2021translating}  & Bare nouns disambiguation in Yorùbá & \badgeFair & \badgePoor & \badgeMissing & \iconLock \\
        
        DR - \cite{oluwatoyinstochastic} & YoTEx framework: 2.8M word corpus; 450K word database & \badgeFair & \badgeFair & \badgeMissing & \iconLock  \\ 

        DR - \cite{adewole2020automatic}  & YorubaÉlision, an elision resolution tool for Yorùbá text & \badgeFair & \badgePoor & \badgeMissing & \iconLock \\   
        DR - \cite{scannell2011statistical}  & Automatic unicodification; support many languages & \badgeGood & \badgeGood & \badgeGood & G\tnote{b} \\    

        DR - \cite{adegbola2016pattern}  & diacritised \& undiacritided data + analogy test sets for evaluating Yorùbá embeddings & \badgeGood & \badgeGood & \badgeGood & \accessPublic \tnote{i} \\ 
        \midrule
        
        \multicolumn{6}{c}{\textbf{Machine Translation \& ASR}} \\
        
        TTS - \cite{afolabi2013development}  & Audio database for Yoruba consonant-vowels & \badgeFair & \badgeMissing & \badgeMissing & \iconLock \\ 

        TTS/ASR - \cite{ahia2024voices}  & YORÙLECT data covering 4 dialects; parallel text/speech across 3 domains & \badgeGood & \badgeGood & \badgeGood & \accessPublic \tnote{c} \\
         
        TTS - \cite{akinwonmi2021development}  & 7,376 phrases/sentences; phonetically balanced corpus & \badgeGood & \badgeFair & \badgeMissing & \iconLock \\ 
        ASR - \cite{gutkin2020developing}  & 4 hours of 48kHz recordings from 36 volunteers; fully diacritised for TTS & \badgeGood & \badgeGood & \badgeFair & \accessPublic \tnote{d} \\ 
        ASR - \cite{abolade2026owe}  & 6 hours recording of 5000 Yoruba proverbs (Òwe-voice) datasets & \badgeGood & \badgeGood & \badgeGood & \accessPublic \tnote{h} \\ 

        Corpus - \cite{fagbolu2015digital} & Yorùbá corpus for translation + POS tagging lexicon & \badgeFair & \badgeFair & \badgeMissing & \iconLock \\
 
        Corpus - \cite{iyanda2015statistical}  & large scaled data; 291,392 words from various sources & \badgePoor & \badgePoor & \badgeMissing & \iconLock \\

        Corpus - \cite{akande2026low} & 1846 manually annotated data; parallel dataset of Yoruba sentences + Nigerian Pidgin translations & \badgeGood & \badgeFair & \badgeMissing & \iconLock \\

        \midrule
        \multicolumn{6}{c}{\textbf{Sentiment Analysis / Event Extraction}} \\

        Event Extraction - \cite{ademusire2023development}  & Annotated text corpus from Yoruba folktale books & \badgeGood & \badgeFair & \badgeMissing & \iconLock \\ 
        Sentiment - \cite{adeniji2024framework}  & Sentiment lexicon from Bible, comments, Yorùbá dictionary & \badgeFair & \badgeFair & \badgeMissing & \iconLock \\ 
       
        Sentiment - \cite{shode2022yosm}  & YOSM data: 1,500 reviews translated from English; MT translation + validation by natives & \badgeGood & \badgeGood & \badgeGood & \accessPublic \tnote{e} \\

        Sentiment - \cite{babatunde2026sentiment}  & 2000 movie reviews data accounting for diacritic & \badgeFair & \badgeFair & \badgeFair & \iconLock \\

        \midrule
        \multicolumn{6}{c}{\textbf{POS / NER}} \\
        
        POS - \cite{adedjouma2013part}  & 312,562 word corpus for POS tagging and translation & \badgeGood & \badgeFair & \badgeMissing & \iconLock \\
 
        POS - \cite{toyin2024hidden}  & 1,000 manually tagged sentences; HMM-based tagger & \badgeGood & \badgeGood & \badgeMissing & \iconLock \\

       \midrule
        \multicolumn{6}{c}{\textbf{Miscellaneous}} \\
        Corpus - \cite{akpobi2024yankari}  & YANKARI: 51,407 documents; 30M+ tokens; Native speaker review & \badgeGood & \badgeFair & \badgeMissing & \accessPublic \tnote{f} \\  
        Tools - \cite{chiarcos2011information}  & ANNIS tool; 25 African languages, including Yoruba & \badgeExcellent & \badgeExcellent & \badgeGood & \iconLock \\ 
        
       Ontology - \cite{okediya2019building}  & General purpose ontology from multiple sources & \badgeFair & \badgePoor & \badgeMissing & \iconLock \\ 
       DR - \cite{orife2018attentive}  & Sequence-to-Sequence model; evaluation dataset & \badgeGood & \badgeGood & \badgeFair & \accessPublic \tnote{h} \\    
       DR - \cite{adegbola2011localising}  & MS Vista localisation glossary; Yorùbá terminologies & \badgeFair & \badgePoor & \badgeMissing & \iconLock \\ 

       \bottomrule
        \end{tabular}
        \begin{tablenotes}[flushleft]
        \footnotesize
        \item[] \textbf{Quality Indicators}: Annot. = Annotation rigour; Valid. = Validation process; IAA = Inter-Annotator Agreement. \textbf{Access}: \iconHuggingFace{} = HuggingFace; \iconKaggle{} = Kaggle; \iconGitHub{} = Github ; ~\iconLock{} = Restricted/unavailable. 
        \tnote{a} MA/DR = Morphological Analysis/Diacritic Restoration.
        \tnote{b} \url{http://sourceforge.net/projects/lingala/}
        \tnote{c} \url{https://github.com/orevaahia/yorulect}
        \tnote{d} \url{https://repo.sadilar.org/}
        \tnote{e} \url{https://github.com/IyanuSh/YOSM}
        \tnote{f} \url{https://huggingface.co/datasets/acflp/YANKARI}
        \tnote{g} \url{https://github.com/Niger-Volta-LTI/yoruba-adr}
        \tnote{h} \url{https://github.com/Holuwasege/OWE-Voice-Evaluation}
        \tnote{i} \url{https://github.com/tosingithub/ydesk}
        \end{tablenotes}
        \end{threeparttable}
\end{table*}
\FloatBarrier


\subsubsection{Multilingual Datasets}
To address the challenges of LRLs and mitigate reliance on costly annotation, several initiatives and strategies have contributed to low-resource multilingual datasets development \cite{kolak2005ocr,varab2021massivesumm,fan2021beyond,hedderich2022weak,godslove2024trilingual}.  
\citet{adelani2023masakhanews} contributed one of the largest dataset and resources for news classification (MasakhaNEWS) in 16 African languages.  
In a related effort, \cite{adelani2021masakhaner} curated a large-scale, publicly available dataset for named-entity recognition (NER) across ten African languages. Moreover, \citet{adelani2022natural} introduced a sizeable human-annotated datasets for NER and machine translation across 21 African languages. 
To support the provision of high-quality human-annotated cross-lingual information retrieval resources for African languages, \citet{adeyemi2024ciral} presented a publicly available test collection for cross-lingual information retrieval (CIRAL) consisting of English queries with passages in Hausa, Somali, Swahili, and Yorùbá languages. 
Motivated by the need to factor in Nigeria’s linguistic and cultural diversity in news coverage, \cite{ogunrinde2026culturally} curated a collection 10,000 Nigerian news articles in English, Nigerian Pidgin, Yoruba, Hausa, and Igbo languages.   
\citet{muhammad2022naijasenti} contributed another large-scale human-annotated Twitter sentiment dataset for Hausa, Igbo, Nigerian Pidgin, and Yorùbá. The collection consists of approximately 30,000 annotated tweets with code-mixed examples. 
\citet{ogunleye2023using} utilised 346,000 Nigerian banking-related tweets to develop SentiLeye, a lexicon-based sentiment analysis dataset for Nigerian Pidgin. 
Multilingual machine translation is another area that is witnessing active contributions. \citet{adewumi2023afriwoz} present useful resources for developing dialogue agents in Swahili, Wolof, Hausa, Nigerian Pidgin, Kinyarwanda, and Yorùbá languages.  
\citet{ajagbe2024developing} utilised  six antenatal clinics to develop a multilingual Nigerian speech dataset in English, Yorùbá, and Pidgin. 
\citet{amuda2010limited} curated an audio-visual dataset (UISpeech) of Nigerian-accented English, recorded from native speakers of Hausa, Igbo, Yorùbá, Tiv, and Fulfulde. 
\citet{aremunaijarc} developed NaijaRC, a multi-choice reading comprehension dataset for Hausa, Igbo, and Yorùbá.  
Recognising the need to address language-related flaws in the benchmark (FLORES) for evaluating NLP performance in low-resource languages \cite{goyal2022flores}, \citet{abdulmumin2024correcting} engaged with native speakers to identify and correct inaccurate and inconsistent cases pertaining to Hausa, Northern Sotho (Sepedi), Xitsonga, and isiZulu languages.  

Relevant studies have contributed to the development of useful corpora to tackle the spread of hateful and offensive content \cite{ferroggiaro2018social,tonneau2024naijahate,muhammad2025afrihate}. Essentially, \citet{tonneau2024naijahate} presented the first annotated data for detecting hateful content, and \citet{muhammad2025afrihate} developed a multilingual corpus of annotated hate speech and abusive language (AfriHate) for 15 African languages. \citet{aliyubeyond} introduced a similar collection of hate speech in the 3 major languages, and \citet{udeze2026language} examined how a combination of Nigerian Pidgin and indigenous languages are strategically utilised in propagating offensive content. 
In addition to contributing a new collection based on Instagram comments, the work in \cite{udeze2026language} utilised the existing NaijaHate corpus \cite{tonneau2024naijahate} to investigate the use and proliferation of offensive language in Nigerian Pidgin. 
For semantic and contextual alignment with low-resource languages, \citet{oduwole2026africaption} curated a multilingual collection of image-based datasets (Flickr8k) to enable image captioning in 20 African languages (including Igbo, Hausa, Yoruba). 

The datasets and strategies in Tables~\ref{tab:contributed-multilingual-datasets1} and ~\ref{tab:contributed-multilingual-datasets2} serve as essential resources for advancing NaijaNLP. The aspect of resource creation has accelerated since 2020 (Figure~\ref{fig:publication-trend-venues}), which we largely attribute to the increasing community initiatives such as Masakhane \cite{adelani2022natural}, cross-lingual benchmarks \cite{adelani2021masakhaner,muhammad2022naijasenti,adelani2023masakhanews,adelani2024irokobench,muhammad2025afrihate}, and funding bodies that encourage collaborative research.  
Table~\ref{tab:summary-of-all-quality-ratings} summarises the quality ratings across the 81 surveyed datasets spanning Hausa, Yorùbá, Igbo, and multilingual resources. While annotation (63\%), validation (62\%), and publicly accessible data (51\%) are reported in most studies, inter-annotator agreement is relatively low (36\%). This points to the need for improved and standardised data development process. Essentially, we recommend adoption of the strategies detailed in datasets for datasets \cite{gebru2021datasheets} to be adhered to for submissions that introduce new resources for NaijaNLP.


\FloatBarrier
    \begin{table*}[!htb]
        \centering
        \small
        \caption{Multilingual Nigerian NLP datasets \& resources for machine translation, ASR, sentiment analysis, POS/NER, and QA. Quality ratings: \badgeExcellent{} rigorous; \badgeGood{} adequate; \badgeFair{} partial; \badgePoor{} limited; \badgeMissing{} not reported.}        
        
        \label{tab:contributed-multilingual-datasets1}
        \begin{threeparttable}
        \begin{tabularx}{\textwidth}{@{} l  X l c c c c @{}}
        \toprule
        
        \multirow{2}{*}{\textbf{Task \& Source}} & \multirow{2}{*}{\textbf{Key Contribution}} & \multirow{2}{*}{\textbf{Langs}} & \multicolumn{3}{c}{\textbf{Quality Indicators}} & \multirow{2}{*}{\textbf{Access}} \\
        \cmidrule(lr){4-6}
        
        & & & \textbf{Annot.} & \textbf{Valid.} & \textbf{IAA} & \\
        \midrule
        \multicolumn{7}{c}{\textbf{Machine Translation \& Automatic Speech Recognition}} \\ 
        MT/NER - \cite{adelani2022natural} & one of the largest human-annotated corpora for 21 African languages for NER \& MT & 21 & \badgeExcellent & \badgeExcellent & \badgeGood & \accessPublic\tnote{a} \\ 
        
        ASR - \cite{ajagbe2024developing} & Antenatal speech dataset; Yoruba \& Pidgin & Y,P & \badgeFair & \badgeFair & \badgeMissing & \accessPublic\tnote{b} \\ 
        
        Embeddings - \cite{alabi2020massive}  & Contextual embeddings; NER annotations & Y,Twi & \badgeGood & \badgeGood & \badgeFair & \accessPublic\tnote{c} \\
         
        ASR - \cite{amuda2010limited}  & Nigerian-accented English audio-visual data (UISpeech) & H,Y,I,Tiv,Ful & \badgeFair & \badgeFair & \badgeMissing & \accessMissing \\ 
        
        ASR - \cite{dossou2021okwugb}  & OkwuGbé ASR for Fon \& Igbo & Fon,I & \badgeGood & \badgeGood & \badgeFair & \accessPublic\tnote{d} \\   
         
        MT - \cite{fan2021beyond}  & M2M-100: supports direct translation between any pair of 100 languages & H, I, Y + others & \badgeGood & \badgeGood & \badgeFair & \accessPublic\tnote{e} \\        
         
        MT/QA - \cite{godslove2024trilingual}  & Trilingual conversation model & E,Hindi,Y & \badgeGood & \badgeGood & \badgeGood & \accessPublic\tnote{f} \\
        
        Dialogue - \cite{adewumi2023afriwoz}  & Afriwoz: dialogue datasets for 6 African languages; 9,000 turns, 1,500 turns/language & Multiple & \badgeExcellent & \badgeExcellent & \badgeExcellent & \accessPublic\tnote{g} \\ 
        
        \midrule
        \multicolumn{7}{c}{\textbf{Sentiment Analysis}} \\
        
        Sentiment - \cite{muhammad2022naijasenti}  & NaijaSenti: 30K labelled tweets for 4 languages & H,Y,I,P & \badgeExcellent & \badgeExcellent & \badgeExcellent & \accessPublic\tnote{h} \\ 
                
        Sentiment - \cite{ogunleye2023using}  & SentiLeye: 346K tweets from bank customers & E,P & \badgeFair & \badgeFair & \badgeMissing & \accessPublic\tnote{i} \\  
        \midrule
        
        \multicolumn{7}{c}{\textbf{POS / Named Entity Recognition}} \\
        
        NER - \cite{adelani2021masakhaner}  & MasakhaNER: NER for 10 African languages & 10 & \badgeExcellent & \badgeExcellent & \badgeGood & \accessPublic\tnote{j} \\

        \midrule
        \multicolumn{7}{c}{\textbf{Question Answering}} \\
        RC - \cite{aremunaijarc}  & NaijaRC: Multi-choice reading comprehension data & H,Y,I & \badgeGood & \badgeGood & \badgeGood & \accessPublic\tnote{k} \\ 
          
        \bottomrule
        \end{tabularx}
        \begin{tablenotes}[flushleft]
        \footnotesize
        \item[] \textbf{Language Codes}: H = Hausa, Y = Yorùbá, I = Igbo, P = Pidgin, E = English, Twi = Twi, Ful = Fulfulde, Fon = Fon. \textbf{Access}: \iconHuggingFace{} = HuggingFace; \iconKaggle{} = Kaggle; \iconGitHub{} = Github ; ~\iconLock{} = Restricted/unavailable.
        \tnote{a} \url{https://github.com/dadelani/africanlp-resources}
        \tnote{b} \url{https://github.com/Adeola80/Antenatal-Speech}
        \tnote{c} \url{https://github.com/ajesujoba/YorubaTwi-Embedding}
        \tnote{d} \url{https://github.com/chrisemezue/IgboASR}
        \tnote{e} \url{https://github.com/pytorch/fairseq/tree/master/examples/m2m_100}
        \tnote{f} \url{https://github.com/juliusco/trilingual-ConvQA}
        \tnote{g} \url{github.com/masakhane-io/chatbots-african-languages} 
        \tnote{h} \url{https://github.com/hausanlp/NaijaSenti}
        \tnote{i} \url{https://www.kaggle.com/batoog/datasets}
        \tnote{j} \url{https://huggingface.co/Davlan/xlm-roberta-large-masakhaner}
        \tnote{k} \url{https://github.com/AremuAdeolaJr/NaijaRC} 
        
        \end{tablenotes}
        \end{threeparttable}
    \end{table*}
\FloatBarrier

\subsection{Sequence Labelling} 
Building on existing corpora, various linguistic tools have been developed to enrich and ensure the accurate representation of LRL. Some of these tools address the specific linguistic characteristics of individual languages, while others focus on enhancing linguistic tools and resources more broadly \cite{abiola2014web,ibrahim2021convolutional,scholar2022development,ekpenyong2022towards,adewumi2022itakuroso,omotayo2024state}. Other related studies include automatic spell checkers and correctors for Yorùbá \cite{oluwaseyi2024automatic} and Hausa \cite{mijinguini2003ƙaramin,salifou2014design}, an automatic Hausa stopword constructor \cite{bichi2022automatic}, and text analyser for Igbo \cite{azubuike2024design}. 
The following section covers linguistic tools including part-of-speech tagging and named entity recognition tools within NaijaNLP.


\FloatBarrier
    \begin{table*}[!htb]
        \centering
        \small
        \caption{Multilingual Nigerian NLP datasets \& resources for text classification, summarisation, and another set of miscellaneous tasks. Quality ratings: \badgeExcellent{} rigorous; \badgeGood{} adequate; \badgeFair{} partial; \badgePoor{} limited; \badgeMissing{} not reported.}  
        
        \label{tab:contributed-multilingual-datasets2}
        \begin{threeparttable}
        \begin{tabularx}{\textwidth}{@{} l  X l c c c c @{}}
        \toprule
 
        \multirow{2}{*}{\textbf{Task \& Source}} & \multirow{2}{*}{\textbf{Key Contribution}} & \multirow{2}{*}{\textbf{Langs}} & \multicolumn{3}{c}{\textbf{Quality Indicators}} & \multirow{2}{*}{\textbf{Access}} \\
        \cmidrule(lr){4-6}
        
        & & & \textbf{Annot.} & \textbf{Valid.} & \textbf{IAA} & \\
        \midrule
        \multicolumn{7}{c}{\textbf{Text Classification (TC) \& Summarisation (TS)}} \\
        
        TC - \cite{adelani2023masakhanews} & MasakhaNEWS: News topic classification for 16 languages & H, I, Y + others & \badgeExcellent & \badgeExcellent & \badgeGood & \accessPublic\tnote{a} \\  
        
        TC - \cite{olalekan2022machine}  & Multilingual text classification & E,Y,H & \badgeFair & \badgeFair & \badgeMissing & \accessMissing \\ 
        
        TS - \cite{varab2021massivesumm}  & MassiveSumm: covers 28.8M articles in 92 languages & H, I, Y + others & \badgeGood & \badgeGood & \badgeFair & \accessPublic\tnote{b} \\
          
        \midrule
        \multicolumn{7}{c}{\textbf{Miscellaneous}} \\
        
        Benchmark - \cite{adelani2024irokobench}  & IrokoBench: translated benchmark for NLI, math, QA for 17 languages & H, I, Y + others & \badgeExcellent & \badgeExcellent & \badgeExcellent & \accessPublic\tnote{c} \\ 
         
        Dialogue - \cite{adewumi2022itakuroso}  & Cross-lingual dialogue generation for 6 languages & H, Y, P + Others  & \badgeGood & \badgeGood & \badgeGood & \accessMissing \\
         
        CLIR - \cite{adeyemi2024ciral}  & CIRAL: Cross-lingual retrieval in many languages & H, Y + Others & \badgeGood & \badgeGood & \badgeGood & \accessPublic\tnote{d} \\
         
        Hate Speech - \cite{aliyubeyond} & Hate speech/offensive tweets & H,Y,I & \badgeFair & \badgeFair & \badgeMissing & \accessMissing \\
        
        Hate Speech - \cite{muhammad2025afrihate}  & AFRIHATE: large scale offensive content in 15 languages & H, I, Y, P + Others & \badgeExcellent & \badgeExcellent & \badgeGood & \accessPublic\tnote{e} \\ 
        
        Lexicon - \cite{ferroggiaro2018social}  & Hate speech lexicon & --- & \badgeGood & \badgeFair & \badgeMissing & \accessMissing \\   
        
        FLORES - \cite{abdulmumin2024correcting}  & Corrected FLORES dataset for 4 African languages & H + Others & \badgeExcellent & \badgeExcellent & \badgeGood & \accessPublic\tnote{f} \\ 
        
        OCR - \cite{kolak2005ocr}  & Weighted finite state machine OCR suitable for low-resource languages & Multi & \badgeGood & \badgeGood & \badgeMissing & \accessMissing \\  

        Offensive Content - \cite{tonneau2024naijahate}  & 35,976 annotated dataset for hate speech detection & H, I, Y & \badgeExcellent & \badgeExcellent & \badgeExcellent & \accessPublic \tnote{g} \\   

        Corpus - \cite{ogundepo2023enabling}  & Cross-lingual retrieval dataset (AfriCLIRMatrix) for many African languages & H, I, Y + Others & \badgeGood & \badgeGood & \badgeMissing & \accessPublic \tnote{j} \\  

        Image Captioning - \cite{oduwole2026africaption}  & image-based datasets (Flickr8k) for semantic and contextual alignment & Multi & \badgeGood & \badgeFair & \badgeFair & \accessMissing \\

        News Classification - \cite{ogunrinde2026culturally}  & 10,000 online news articles & H, I, Y, P & \badgeGood & \badgeGood & \badgeGood & G\tnote{h \& i} \\ 
        
        \bottomrule 
        \end{tabularx}
        \begin{tablenotes}[flushleft]
        \footnotesize
        \item[]   \textbf{Language Codes}: H = Hausa, Y = Yorùbá, I = Igbo, P = Pidgin, E = English, Som = Somali, Swa = Swahili. \textbf{Access}: \iconHuggingFace{} = HuggingFace; \iconKaggle{} = Kaggle; \iconGitHub{} = Github ; ~\iconLock{} = Restricted/unavailable. 
        \tnote{a} \url{https://github.com/masakhane-io/masakhane-news}
        \tnote{b} \url{https://github.com/danielvarab/massive-summ}
        \tnote{c} \url{https://github.com/masakhane-io/masakhane-nlu}
        \tnote{d} \url{https://github.com/ciralproject/ciral}
        \tnote{e} \url{https://github.com/AfriHate/AfriHate}
        \tnote{f} \url{https://github.com/dsfsi/flores-fix-4-africa}
        \tnote{g} \url{https://github.com/worldbank/NaijaHate}
        \tnote{h} \url{https://drive.google.com/file/d/1KGP0kuMvqu8emlECPpxONAZslyXmD-p0/view} 
        \tnote{i} \url{https://docs.google.com/spreadsheets/d/1zIuirE2m7Sk2OWuY9o2doanNCpmCzDoLyekoEb96ee4/edit?gid=0#gid=0} 
        \tnote{j} \url{https://github.com/castorini/africlirmatrix} 
        \end{tablenotes}
        \end{threeparttable}
    \end{table*}
\FloatBarrier


\FloatBarrier
\begin{table*}[!htb]
        \centering
        \small
        \caption{Summary of quality indicators across the contributed datasets for NaijaNLP.}
        \label{tab:summary-of-all-quality-ratings}
        \begin{tabular}{@{} l c c c c c @{}}
        \toprule
        \multirow{2}{*}{\textbf{Quality Indicator}} & \multicolumn{4}{c}{\textbf{Category}} &  \multirow{2}{*}{\textbf{Overall (81)}} \\
        \cmidrule(lr){2-5}
        & \textbf{Hausa (20)}  & \textbf{Igbo (10)} & \textbf{Yorùbá (24)} & \textbf{Multilingual (27)}  & \\
        \midrule
        Annotation Reported & 13 (65\%) & 2 (20\%) & 14 (58\%) & \textbf{22 (81\%)} & 51 (63\%) \\
        Validation Performed & 12 (60\%) & 2 (20\%) & 9 (38\%) & \textbf{20 (74\%)} & 50 (62\%) \\
        Inter-Annotator Agreement & 8 (40\%) & 1 (10\%) & 6 (25\%) & \textbf{14 (52\%)} & 29 (36\%) \\
        Accessible Data & 12 (60\%) & 1 (10\%) & 8 (33\%) & \textbf{20 (74\%)} & 41 (51\%) \\
        \bottomrule
        \end{tabular}
        \vspace{2mm}
\end{table*}
\FloatBarrier

\subsubsection{Part-of-Speech Tagging and Named Entity Recognition}  
Depending on its context and definition, each word in a text belongs to a grammatical category, such as noun, verb, adjective, adverb and preposition. Thus, Part-of-Speech (POS) tagging facilitates the structural analysis of sentences by assigning words to their corresponding grammatical groups. Similarly, some words in a text may refer to specific entities such as people, organisations, locations, and dates. Thus, Named Entity Recognition (NER) identifies and classifies such entities accordingly. To that end, several studies have explored both POS tagging and NER related to NaijaNLP and beyond \cite{maitama2014text,ren2016automatic,lu2016multi,adelani2021masakhaner,mehari2024semi,gasparini2026automated}.   
\citet{adelani2020distant} examined NER in Hausa and Yorùbá, while \citet{model2020accent} explore accent classification for Nigerian-Accented English, covering Hausa, Yorùbá, and Igbo.  
\citet{tukur2024towards} introduced a POS tagger for Kanuri\footnote{A Nilo-Saharan language spoken in the Lake Chad Basin of West and Central Africa} and corresponding corpus. 
\citet{adewumi2022itakuroso} explored cross-lingual transfer learning to optimise Hausa and Yorùbá POS and NER tasks. 

\paragraph{POS \& NER for NaijaNLP}
Given the limitations of classic stemmers based on Porter’s algorithm \cite{porter1980algorithm}, researchers have developed Hausa-specific stemming techniques that account for language particularities \cite{bashir2015word,bimba2016stemming,musa2022improved}. 
Essentially, \citet{bashir2015word} introduced an automatic Hausa word stemming system to optimise text processing, while \citet{bimba2016stemming} developed a stemming strategy using 78 affix-stripping rules and reference lookup based on a database of 1500 Hausa root words.   
To minimising noise in supervised learning methods, \citet{zhang2017embracing} introduced a low-resource POS tagging strategy that integrates diverse linguistic sources comprising of the World Atlas of Linguistic Structure, CIA names, PanLex, and survival guides.   
\citet{tukurcorpus,tukur2019tagging,tukur2020parts} examined part-of-speech tagging techniques to facilitate sentiment analysis of Hausa web content.   
A group of related studies have contributed to the development of POS \& NER in Igbo language \cite{onyenwe2015use, onyenwe2016predicting, onyenwe2014part, onyenwe2019toward, olamma2019hidden, anbootstrapping}.  
For NER in Igbo, \citet{chukwuneke2023igboner} utilised a cross-language projection method that leverages a parallel English-Igbo corpora and \citet{soronnadienhancing} apply a transformer-based model (IgboBERTa) optimised for Igbo NLP tasks, including NER, text classification, and sentiment analysis. 

Similarly, appreciable efforts have been made to advance Yorùbá POS tagging and NER \cite{ugwu2024part}. 
\citet{kumolalo2010development} designed a rule-based approach to enhance Yorùbá syllabification, mapping words to their corresponding syllables, while \citet{adedjouma2013part} curated a Yorùbá corpus focused on part-of-speech tagging.  
\citet{adegunlehin2019investigation} improved Yorùbá NER by incorporating contextual information such as surrounding words and POS tags. 
\citet{omolaoye2020proverb} introduced a computational approach for the representation of Yorùbá proverbs, contributing to indigenous knowledge preservation. 
\citet{toyin2024hidden} implemented a Hidden Markov Model (HMM)-based POS tagger, contributing a manually annotated dataset of 1,000 Yorùbá sentences drawn from various domains. 
The development of POS tagging and NER tools for Hausa, Igbo, and Yorùbá has contributed to NaijaNLP in addressing language-specific challenges, enhancing computational linguistic resources, and facilitating various downstream tasks. 
While existing resources have proven useful, further research and dataset expansion are needed to bridge gaps and improve linguistic tool development for NaijaNLP.

\subsection{Downstream Tasks} 
\label{sec:downstream-tasks}
In NLP, downstream tasks such as text classification, sentiment analysis, machine translation, and speech recognition rely fundamentally on rich annotated datasets and effective computational models. In this section, we review the existing downstream tasks and applicable resources that have been developed for Hausa, Igbo, and Yorùbá languages. 

\subsubsection{Text Classification and Summarisation} 
\label{sec:downstream-task-sentiment-analysis} 
Text classification (TC) involves assigning textual data to predefined classes based on a similarity or dissimilarity metric informed by a range of methodologies from rule-based approaches to deep learning models.   
As a specialised form of text classification, sentiment analysis (SA) deals with identifying subjective information in a text collection to determine its polarity or sentiment. In addressing the unique linguistic characteristics of LRLs, several studies have contributed to advancing SA in NaijaNLP \cite{muhammad2022naijasenti,raychawdhary2023transformer,abegunde1832design,abdullahi2023hausanlp,raychawdhary2024optimizing,shehu2024unveiling,raychawdhary2025empowering}.  
\citet{abdou2025monitoring} designed a framework for assessing online geopolitical news based on sentiment and public attention across multiple African countries including Kenya, Nigeria, Senegal, and South Africa. Complementary approaches involve the use of an enhanced feature acquisition method for multilingual sentiment analysis involving English and Hausa tweets \cite{abubakar2021enhanced}.   
\citet{sani2022sentiment} developed a system for sentiment analysis on Hausa data extracted from BBC Hausa's Twitter (now X) handle\footnote{\url{https://www.bbc.com/hausa} and \url{https://x.com/bbchausa}}. In related studies, \citet{akande2022tweerify} and \citet{ibrahim2024deep} utilised Hausa movies review data to train models for detecting aspect-level sentiment. Moreover, \citet{abdullahi2024twitter} is focused on identifying sentiment associated with abbreviated Hausa terms \cite{abdullahi2024twitter}, and \citet{rakhmanov2022sentiment} utilised a collection of 40,000 students' comments to develop a system for analysing sentiment. 
For resource enrichment, \citet{mohammed2023building} curated a collection of sentiment lexicons for sentiment analysis in Hausa. Moreover,  \cite{hughes2023bhattacharya_lab} developed a useful strategy based on transformer model to analyse sentiment in Nigerian Pidgin and Yorùbá languages. As one of the early studies leveraging pre-trained models, the study demonstrated the potential for NLP techniques to be applied to low-resource languages and improve downstream tasks. 

In the Igbo context, \citet{ogbuju2020development} introduced general-purpose sentiment lexicons (IgboSentilex), and \citet{okoloegbo2022multilingual} implemented an interactive system for detecting, monitoring, and regulating cyberbullying content in Igbo and Pidgin English. Supported by a collection of sentiment lexicons derived from diverse sources, \citet{adeniji2024framework} proposed a system for disambiguating sentiment lexicons in Yorùbá. 
\citet{shode2022yosm} utilised reviews from movies to develop the YOSM dataset for sentiment analysis. 
Although relevant datasets have contributed useful datasets (see Tables~\ref{tab:contributed-hausa-datasets},~\ref{tab:contributed-igbo-datasets},~\ref{tab:contributed-yoruba-datasets}, ~\ref{tab:contributed-multilingual-datasets1} and ~\ref{tab:contributed-multilingual-datasets2}), there is a compelling need for more high-quality annotated data to train models that are better equipped at capturing complex linguistic patterns, including sarcasm, irony, and context-dependent sentiments in NaijaNLP. 

Complementing text classification, text summarisation (TS) aims to generate concise and coherent summaries that retain the essential information of longer texts.  
\citet{varab2021massivesumm} designed a multilingual text summarisation framework by enriching existing resources to facilitate machine translation (MT) on a large scale. \citet{olalekan2022machine} implemented multilingual text classification models for English, Yorùbá, and Hausa, while \citet{bashir2017automatic} utilised a collection of Hausa documents to train summarisation models. Additional contributions include the development of automatic summarisation strategy for Hausa text \cite{bichi2023graph,bichi2024integrating}. 
For Igbo, \citet{ifeanyi2020comparative, ifeanyin} proposed classification systems based on n-gram models and k-nearest neighbours techniques to improve text representation and classification. In parallel, \citet{mbonu2022igbosum1500} offered a detailed discussion about the development of the IgboSum1500 dataset, a dedicated resource for Igbo text summarisation. Meanwhile, \citet{adegokeestimating} explored models designed to compute semantic similarity between Yorùbá sentences. For text summarisation, in particular, the development of more purposive datasets and models capable of better contextual understanding, semantics, and inter-textual relationships will further enrich NaijaNLP.

\subsubsection{Machine Translation and Speech Recognition} 
Machine Translation (MT) focuses on the automatic translation of text or speech from one language to another, and it has evolved from rule-based to transformer-based systems. Effective MT systems are heavily dependent on large volumes of parallel data to generate accurate translations \cite{mahata2020performance}.  
In addressing the needs of low-resource languages, \citet{akinfaderin2020hausamt} develop a translation system for English–Hausa trained on the JW300, Tanzil, Tatoeba and Wikimedia public datasets. Additionally, \citet{tresner2023intent} propose a system for intent recognition in user messages directed to a chatbot (askNivi), which is designed to facilitate discussions on health-related topics in Hausa, Hindi, and Swahili. In another useful approach, \citet{brugnone2024ought} leverage the wisdom of the crowd and unsupervised learning strategies to identify subpopulations based on value-laden statements extracted from Hausa narratives related to maternal and child healthcare. Responding to the need for on-demand translation, \citet{watt2023edge} present an approach for deploying MT on the edge, thereby enabling an embedded system for English–Hausa translation.  

For performance improvement, \citet{hatami2024english} utilise a multimodal strategy integrating visual and textual information for translation across language pairs consisting of English, Hindi, Malayalam, Bengali, and Hausa.   
\citet{robertson2022understanding} propose strategies centred around transparency to reduce translation error for chat agents in LRLs. 
\citet{maryann2022enhanced} and \citet{usip2023text} develop MT systems for English–Igbo using both rule-based and corpus-based approaches.  
To lessen the impact of harmful English-Igbo translation errors emanating from non-compositional multiword expressions and polysemy, \citet{ohuoba2024quantifying} present diagnostic tests involving high-stake domains.  
\citet{adegbola2011localising} explore a useful strategy for localising the MS Vista operating system by translating relevant English lexicons into Yorùbá terminologies.  
\citet{eludiora2015development} develop a rule-based system to address tone changes in Yorùbá verbs and \citet{fagbolu2016applying} design an accessible MT system for English–Yorùbá translation via mobile and web platforms. 
\citet{eludiora2016development} develop an English–Yorùbá translator and \citet{sundaydevelopment} investigate a multi-layer hybrid approach that combines data-driven and rule-based methods. 
Finally, \citet{timothybilingual} focus on mitigating challenges such as vanishing gradients, translation accuracy, and computational efficiency in English–Yorùbá MT systems.  
Although MT is one of the most widely utilised NLP applications, its performance on LRL pairs remains suboptimal compared to that of HRL \cite{robertson2022understanding}, primarily due to insufficient parallel data. There is a high proportion of studies focused more on training and optimisation strategies rather than contributing new resources. Future research on MT should emphasise the development of more accessible richer parallel data to train models that can effectively capture complex linguistic patterns, manage long-range dependencies, accommodate domain-specific terminology, and preserve cultural nuances for NaijaNLP. 

\paragraph{Automatic Speech Recognition} 
Automatic speech recognition (ASR), interchangeably referred to as speech recognition, involves converting spoken language into text using methods spanning pattern matching to more effective transformer-based models. 
To mitigate the challenge of limited data, several studies have reported useful techniques and frameworks \cite{odelobi2008recognition,luka2012neural,gibbonmarketspeak,aliero10taxonomy,ibrahim2022graphic}. 
\citet{ngueajio2022hey} presented strategies to mitigate biases in speech recognition systems, especially those related to gender, race, and disabilities.  
\citet{zhou2024character} utilised meta-learning via pre-training on speech data from ten languages; for application, dataset from the target language is fine-tuned.  
Noting that the removal of diacritics can increase homography and degrade recognition accuracy, \citet{abubakar2024development} developed an ASR system for Hausa that emphasises diacritised word forms.  
\citet{ibrahim2022framework} propose an integrated framework for Hausa ASR that encompasses data creation, as well as the development of both acoustic and language models. 
With focus on reflecting cultural nuances, \cite{abolade2026owe} curated speech datasets for oral Yoruba literature and and proverbs. 
Further contributions for Yorùbá include the development of speech-to-text recognition systems \cite{afolabi2013development,akintola2017machine, babatunde2024speech}, and a localised Google navigation voice system in fluent Yorùbá \cite{oyesanmi2024towards}. 
From application point of view, relevant dialogue systems for handling information retrieval and frequently asked questions have been reported. 
For instance, \citet{mabrouk2021multilingual} developed a multilingual chat agent that supports African dialects using a general-purpose neural embedding model capable of addressing a broad range of tasks \cite{wu2018starspace}. 
To enhance COVID-19 understanding in Northeast Nigeria, Translators without Borders\footnote{\url{https://translatorswithoutborders.org/chatbot-release-northeast-nigeria/}} designed a chatbot, \textit{Shehu}, that engaged with users in English, Hausa, or Kanuri. 
\citet{haruna2021hausa} developed a Hausa-language chatbot intended to engage users in storytelling and facilitate language learning among native speakers. \citet{aremu2024utilising} explored the potential of AI-powered chatbots in supporting the learning of endangered Nigerian languages. 

Tables~\ref{tab:contributed-multilingual-datasets1} and ~\ref{tab:contributed-multilingual-datasets2} suggest that the overall progress and volume of studies in ASR remain limited. So far, it is a recurring theme that only a handful of studies contributed new resources with majority of the studies reusing existing resources. Thus, additional resources to develop models capable of managing complex acoustic and linguistic patterns, including diverse accents, diacritic preservation, and variable speech patterns are required. 

\section{Discussion and Findings}
\label{sec:results-discussion} 
Following a review of the related studies in the previous sections, this section synthesises the current state of NaijaNLP, prevailing challenges and prospects for future development. Our findings are organised around key dimensions derived from the research questions outlined in \S~\ref{sec:introduction}. 

\subsection{State of Affairs in NaijaNLP} 
This section articulates key development in resources for NaijaNLP, alongside ongoing efforts to expand and enhance them. Tables~\ref{tab:contributed-hausa-datasets},~\ref{tab:contributed-yoruba-datasets},~\ref{tab:contributed-igbo-datasets},~\ref{tab:contributed-multilingual-datasets1} and ~\ref{tab:contributed-multilingual-datasets2} provide aggregated overview of the available resources and related studies across NaijaNLP languages. 
Additionally, Table~\ref{tab:datasets-and-downstream-tasks} outlines specific datasets and their corresponding downstream tasks.  
A key insight in this regard is the growing body of literature addressing various NLP downstream tasks for Hausa, Igbo, and Yorùbá languages. However, significant gaps remain pertaining to adequate labelled data and effective handling of language-specific particularities.  
Of the reviewed studies, only about $27.6\%$ of the reviewed studies have contributed novel resources, with around 81 new resources emerging from a final set of approximately 293 papers (refer to Figure~\ref{fig:literature-search}). This suggests a disproportionate reliance on existing linguistic resources. Such over-dependence on repurposing existing data, rather than creating new, high-quality, and well-annotated datasets, underscores the need for increased collaborative efforts in developing high quality and accessible linguistic resources.

\subsubsection{State of Affairs in Linguistic Resources and NLP Tools} 
We discuss the state of linguistic resources for NaijaNLP in relation to grammar, knowledge representation, language particularities, datasets, and downstream tasks. 

\paragraph{1. Grammar and Knowledge Representation} 
Despite their significance in providing a critical foundation for the development of digital resources through structured frameworks for language processing, research in grammar and knowledge representation (detailed in \S~\ref{sec:grammar-and-knowledge-representation}), remains limited within NaijaNLP. Formal grammar and its analysis is prominent in research involving Hausa, focusing on grammatical formalisation \cite{abdoulaye1992aspects,culy1996formal,crysmann2009autosegmental,copestake2002implementing,crysmann2012hag,blasi2017grammars}. 
Expanding research in these areas is crucial for enhancing the representation of LRLs and improving the performance of generative AI models for these languages. Moreover, digitalisation of language resources will strengthen NaijaNLP and narrow the divide between theoretical linguistics and practical applications.

\paragraph{2. Language Particularities}
As previously noted, Nigerian languages often exhibit distinct orthographic, phonetic, and syntactic characteristics that set them apart from English and other high-resource languages. Consequently, inadequate resources and consideration of these unique characteristics affect the adaptation and performance of NLP tools that have been developed primarily for HRLs when applied to low-resource contexts. Numerous studies have addressed the linguistic diversity and specific needs of Nigeria's major languages \cite{amuda2010limited,oyinloye2015issues,abdulkareem2016yorcall,adewumi2020challenge}. Languages within NaijaNLP, especially Igbo and Yorùbá, exhibit rich morphological structures and extensive use of diacritics that challenge conventional techniques. Thus, various studies have examined these challenges pertaining to Igbo \cite{iheanetu2017some,iheanetu2019addressing,ochu2019corpus,chidiebere2020analysis,henry2020performance} and Yorùbá \cite{oyinloye2015issues,adegbola2016pattern,abdulkareem2016yorcall,adewole2017token,oluwatoyinstochastic,okediya2019building}, and \cite{alade2019issues,oyekanmi2013intelligent,adebara2021translating,adegokedevelopment,akinwonmi2024rule,frohmann2024segment}.  
Diacritic restoration plays a crucial role in processing historical texts, user-generated content, and datasets where diacritics may have been omitted due to technical constraints.  
Although some insightful theoretical and analytical studies pertaining to orthographic and tonal diacritics for Igbo \cite{ezeani2016automatic,ezeani2017lexical,ezeani2018transferred,ezeani2018igbo} and Yorùbá \cite{de2007automatic,tunji2011design,adegbola2012quantifying,asubiaro2015statistical,asahiah2017restoring,orife2018attentive,alabi2020massive,adewumi2020challenge} languages have been undertaken, advancing research in this domain requires more comprehensive datasets, appropriate tools and greater community involvement to account for cultural nuances and accurate processing of diacritics in NaijaNLP. 
For additional details on morphological analysis and diacritic restoration, refer to \S~\ref{sec:morphological-analysis} and \S~\ref{sec:diacritic-restoration}, respectively. 
 
Furthermore, there has been limited research focused on enriching linguistic resources related to numerical systems in NaijaNLP. The earliest attempts at analysing Hausa numerals for machine processing date back to \cite{sigurd1980numbers}. Subsequent efforts have explored the conversion between spoken numerals and decimal representations \cite{sigurd1980numbers,agbeyangi2016web,mustapha2023automated,rhoda2017computational,ajao2022yoruba}. Overall, research on numerical systems within NaijaNLP is limited, underscoring the need for further studies and resource development to ensure accurate numerical representation and processing. 

\paragraph{3. Monolingual and Multilingual Data}
Alongside the curated corpora for Hausa, Igbo, and Yorùbá languages (see Section~\ref{sec:linguistic-resources-datasets}), a range of strategies, including data augmentation techniques, have been proposed to improve both the quality and quantity of the linguistic resources.  
From Tables~\ref{tab:contributed-multilingual-datasets1}, ~\ref{tab:contributed-multilingual-datasets2} and \ref{tab:summary-of-all-quality-ratings}, it is evident that the majority of high-volume, high-quality data employed across the downstream tasks come from the cross-lingual category, a result of effective collaborative initiatives.  
The shortfall associated with monolingual datasets (Tables~\ref{tab:contributed-hausa-datasets},~\ref{tab:contributed-igbo-datasets} and ~\ref{tab:contributed-yoruba-datasets}) underscores the need for further refinement and systematic annotation to advance state-of-the-art NLP research in realising the full potential of NaijaNLP \cite{adegbola2009building, bigi2017developing, adeyemi2024ciral}. 

\subsubsection{State of Affairs in Downstream Tasks}  
Relevant studies have been dedicated to developing various downstream tasks within the NaijaNLP ecosystem. This section discusses key findings regarding such tasks and applicable tools. Discussion about the state of affairs in resource enhancement is covered in \S~\ref{sec:mitigation-strategies-and-prospects}. 

\paragraph{1. POS and NER}
Numerous studies contributed to POS tagging and NER within the context of NaijaNLP, addressing language-specific challenges and nuances. Essentially, studies have been reported in monolingual contexts for Hausa \cite{bashir2015word,bimba2016stemming,lu2016multi,tukurcorpus,tukur2019tagging,tukur2020parts,awwalu2021corpus}, Igbo \cite{onyenwe2015use,onyenwe2016predicting,onyenwe2014part,onyenwe2019toward,olamma2019hidden,anbootstrapping,chukwuneke2023igboner,soronnadienhancing}, Yorùbá \cite{kumolalo2010development,adedjouma2013part,abiola2014web,adegunlehin2019investigation,omolaoye2020proverb,ugwu2024part,toyin2024hidden} languages, and in multilingual settings \cite{ren2016automatic,zhang2017embracing,adelani2020distant,model2020accent,oyewusi2021naijaner,tukur2024towards,adewumi2022itakuroso,mehari2024semi}. These contributions are instrumental in addressing language-specific challenges and in enhancing the computational linguistic resources underlying various NLP applications. However, while existing resources have proven valuable, further research and rich dataset expansion are required to advance and keep pace with prevailing progress in the field.

\paragraph{2. Text Classification and Sentiment Analysis} 
While numerous studies have focused on text classification and summarisation \cite{varab2021massivesumm,olalekan2022machine,bashir2017automatic,bichi2023graph,bichi2024integrating,ifeanyi2020comparative,ifeanyin,mbonu2022igbosum1500,adegokeestimating}, linguistic resources and models capable of effectively capturing context, semantics, and intertextual relationships are still required.  
In addition to text classification, significant research has been conducted on sentiment analysis using datasets from social media and other public sources (see \S~\ref{sec:downstream-task-sentiment-analysis}, Tables~\ref{tab:contributed-hausa-datasets},~\ref{tab:contributed-igbo-datasets},~\ref{tab:contributed-yoruba-datasets},~\ref{tab:contributed-multilingual-datasets1}, and ~\ref{tab:contributed-multilingual-datasets2}). The approaches range from traditional bag-of-words models to state-of-the-art transformer-based techniques, each addressing the nuances of sentiment expression in low-resource contexts \cite{abubakar2021enhanced,shehu2024unveiling,akande2022tweerify,ibrahim2024deep,sani2022sentiment,mohammed2023building,abdullahi2024twitter,ogbuju2020development,adeniji2024framework,abegunde1832design,shode2022yosm}. 
While some related studies based on deep learning and transformer-based models aimed at addressing sentiment analysis in NaijaNLP have been conducted \cite{yusuf2023fine,raychawdhary2023transformer,raychawdhary2024optimizing,abdullahi2023hausanlp,abdou2025monitoring,rakhmanov2022sentiment}, pretrained models, specifically those trained on indigenous data remain limited \cite{muhammad2022naijasenti,muhammad2025afrihate}. There is a need for more representative data and models that can effectively capture complex linguistic phenomena, including sarcasm, irony, and context-dependent sentiments in NaijaNLP.

\paragraph{3. Machine Translation and Automatic Speech Recognition}
Many studies have focused on machine translation for Hausa \cite{akinfaderin2020hausamt,tresner2023intent,hatami2024english}, Igbo \cite{robertson2022understanding,maryann2022enhanced,usip2023text,ohuoba2024quantifying}, and Yorùbá \cite{adegbola2011localising,folajimi2012using,odoje2014investigating,eludiora2015development,fagbolu2016applying,eludiora2016development,sundaydevelopment,timothybilingual} languages. 
Furthermore, research in speech technologies has explored both automatic speech recognition and text-to-speech (TTS) systems, with particular emphasis on addressing challenges such as diacritic preservation and tonal variation inherent to Hausa \cite{luka2012neural,aliero10taxonomy,ibrahim2022framework,abubakar2024development,ibrahim2022graphic}, Igbo \cite{gibbonmarketspeak,ngueajio2022hey}, and Yorùbá \cite{dossou2021okwugb,odelobi2008recognition,aoga2016integration,akintola2017machine,babatunde2024speech,afolabi2013development,oyesanmi2024towards} languages.   
Future research should emphasise the development of parallel data and pretrained models specifically trained on indigenous data.  


    \begin{table}[h!]
            \centering
            \renewcommand{\arraystretch}{1.3}
            \adjustbox{max width=\textwidth}{
            \begin{tabular}{|l|cccccc|cc|c|}
                \hline
                \rowcolor{gray!20} 
                \multicolumn{9}{|c|}{\cellcolor{gray!30} \textbf{Datasets and Application Areas}} \\ \hline

                \rowcolor{gray!20} 
                \textbf{Dataset} & \textbf{Proportion (\%)} & \textbf{MT/ASR} & \textbf{SA} & \textbf{TC/TS} & \textbf{POS/NER} & \textbf{QA} & \textbf{IR} & \textbf{Misc} \\ 
                \hline
                
                Hausa Only & \cellcolor{blue!25} 24.7\% & \cellcolor{green!25}\checkmark & \cellcolor{green!25}\checkmark & \cellcolor{green!25}\checkmark & \cellcolor{green!25}\checkmark & \cellcolor{green!25}\checkmark & \cellcolor{green!25}\checkmark & \cellcolor{green!25}\checkmark \\
    
                Hausa-Igbo-Yorùbá & \cellcolor{blue!25} 5.6\% & \cellcolor{green!25}\checkmark & \cellcolor{green!25}\checkmark & \cellcolor{red!25}\xmark & \cellcolor{green!25}\checkmark & \cellcolor{green!25}\checkmark & \cellcolor{red!25}\xmark & \cellcolor{green!25}\checkmark \\
    
                Hausa-Yorùbá-Others & \cellcolor{blue!25} 2.3\% & \cellcolor{red!25}\xmark & \cellcolor{red!25}\xmark & \cellcolor{red!25}\xmark & \cellcolor{green!25}\checkmark & \cellcolor{red!25}\xmark & \cellcolor{red!25}\xmark & \cellcolor{green!25}\checkmark  \\
    
                Hausa-English & \cellcolor{blue!25} 4.5\% & \cellcolor{green!25}\checkmark & \cellcolor{green!25}\checkmark & \cellcolor{red!25}\xmark & \cellcolor{red!25}\xmark & \cellcolor{red!25}\xmark & \cellcolor{red!25}\xmark & \cellcolor{red!25}\xmark  \\
    
                Hausa-Others & \cellcolor{blue!25} 5.6\% & \cellcolor{green!25}\checkmark & \cellcolor{red!25}\xmark & \cellcolor{green!25}\checkmark & \cellcolor{red!25}\xmark & \cellcolor{red!25}\xmark & \cellcolor{red!25}\xmark & \cellcolor{red!25}\xmark  \\
                
                Igbo Only & \cellcolor{blue!25} 10.1\% & \cellcolor{green!25}\checkmark & \cellcolor{green!25}\checkmark & \cellcolor{green!25}\checkmark & \cellcolor{green!25}\checkmark & \cellcolor{red!25}\xmark & \cellcolor{red!25}\xmark & \cellcolor{green!25}\checkmark \\
    
                Igbo-Yorùbá-English & \cellcolor{blue!25} 2.3\% & \cellcolor{green!25}\checkmark & \cellcolor{red!25}\xmark & \cellcolor{red!25}\xmark & \cellcolor{red!25}\xmark & \cellcolor{red!25}\xmark & \cellcolor{red!25}\xmark & \cellcolor{red!25}\xmark  \\
    
                Igbo-Others & \cellcolor{blue!25} 3.4\% & \cellcolor{green!25}\checkmark & \cellcolor{red!25}\xmark & \cellcolor{red!25}\xmark & \cellcolor{red!25}\xmark & \cellcolor{red!25}\xmark & \cellcolor{red!25}\xmark & \cellcolor{green!25}\checkmark  \\
                
                Yorùbá Only & \cellcolor{blue!25}14.6\% & \cellcolor{green!25}\checkmark & \cellcolor{green!25}\checkmark & \cellcolor{green!25}\checkmark & \cellcolor{green!25}\checkmark & \cellcolor{green!25}\checkmark & \cellcolor{green!25}\checkmark & \cellcolor{green!25}\checkmark \\
    
                Yorùbá-English & \cellcolor{blue!25} 3.4\% & \cellcolor{green!25}\checkmark & \cellcolor{red!25}\xmark & \cellcolor{red!25}\xmark & \cellcolor{red!25}\xmark & \cellcolor{red!25}\xmark & \cellcolor{red!25}\xmark & \cellcolor{red!25}\xmark  \\
    
                Yorùbá-Others & \cellcolor{blue!25} 4.5\% & \cellcolor{green!25}\checkmark & \cellcolor{red!25}\xmark & \cellcolor{red!25}\xmark & \cellcolor{green!25}\checkmark & \cellcolor{red!25}\xmark & \cellcolor{red!25}\xmark & \cellcolor{red!25}\xmark \\
    
                African Languages & \cellcolor{blue!25} 7.9\% & \cellcolor{green!25}\checkmark & \cellcolor{green!25}\checkmark & \cellcolor{red!25}\xmark & \cellcolor{green!25}\checkmark & \cellcolor{red!25}\xmark & \cellcolor{red!25}\xmark & \cellcolor{green!25}\checkmark \\
                
                Hausa-Igbo-Yorùbá-English-Others & \cellcolor{blue!25} 11.2\% & \cellcolor{green!25}\checkmark & \cellcolor{green!25}\checkmark & \cellcolor{green!25}\checkmark & \cellcolor{green!25}\checkmark & \cellcolor{green!25}\checkmark & \cellcolor{green!25}\checkmark & \cellcolor{green!25}\checkmark \\
                \hline
            \end{tabular}
            }
            \caption{Overview of reported datasets across the languages documented in the reviewed literature. The above acronyms refer to the following: MT/ASR – Machine Translation/Automatic Speech Recognition; SA – Sentiment Analysis; TC/TS – Text Classification/Text Summarisation; POS/NER – Part-of-Speech Tagging \& Named Entity Recognition; QA – Questions and Answers; IR – Information Retrieval; Misc – Miscellaneous downstream tasks. A green checkmark (\checkmark) indicates the availability of the resource, whereas a red $X$ (\xmark) signifies its absence.}  
            \label{tab:datasets-and-downstream-tasks}
        \end{table}        


\subsection{Challenges and Future Prospects for NaijaNLP}  
In this section, we discuss the challenges hindering NaijaNLP from fully leveraging the state-of-the-art NLP advances and recommend mitigation strategies. 

\subsubsection{Limited Foundational Frameworks and Data Processing Challenges} 
Computational grammar and knowledge representation are foundational to the development of digital language resources because the offer structured frameworks for language processing \cite{crysmann2012hag}.   
The linguistic particularities and diversity of Nigerian languages present significant challenges for standard NLP techniques because these languages feature unique orthographic systems, pronunciation patterns, and syntactic structures, including the use of diacritics and tonal variations.   
Additionally, complex morphological structures, such as rich affixation and non-concatenative morphology, further complicate processing. The frequent omission or misapplication of diacritics, as well as issues like mis-syllabification and elision, emphasise the need for specialised linguistic resources tailored to Nigerian Languages. However, studies in these areas remain sparse within NaijaNLP \cite{abdoulaye1992aspects,culy1996formal,crysmann2009autosegmental,copestake2002implementing,blasi2017grammars,eze1997aspects,aina2021ontology,ifeanyi2022semantic,aina2023human}. 
 
\paragraph{1. Limited Digital and Standardised Resources}  
Despite the population size and increasing number of Internet users \cite{internetusengn2024}, the languages under NaijaNLP lack sufficient digital resources to effectively support computational tasks.  
For context, these languages collectively have fewer than 200,000 Wikipedia articles (see Table~\ref{tab:wiki_articles_language} for a subset of the 353 languages with official Wikipedia editions), which play key role in training NLP models.  
As highlighted in the rows of sub-tables (b) and (c) of Table~\ref{tab:wiki_articles_language}, low-resource languages such as Hausa, Igbo, and Yorùbá have minimal digital presence. Expanding these resources with quality data is essential for enhancing the digital footprint, research potential, and NLP capabilities of these languages. 

    \begin{table}[h]
        \centering
        \renewcommand{\arraystretch}{1.2} 
        \setlength{\tabcolsep}{4pt} 
        \resizebox{\textwidth}{!}{ 
        \begin{tabular}{lccr | lccr | lccr}
            \toprule
            \multicolumn{4}{c}{\textbf{(a) High-Article Languages (1.5M+)}} & \multicolumn{4}{c}{\textbf{(b) Low-Article Languages (54K-170K)}} & \multicolumn{4}{c}{\textbf{(c) Very Low-Article Languages (38K-49K)}} \\
            \cmidrule(lr){1-4} \cmidrule(lr){5-8} \cmidrule(lr){9-12}
            \textbf{Language} & \textbf{\#Articles} & \textbf{\#Active Users}  & \textbf{Ratio} & \textbf{Language} & \textbf{\#Articles} & \textbf{\#Active Users} & \textbf{Ratio} & \textbf{Language} & \textbf{Articles} & \textbf{\#Active Users} & \textbf{Ratio} \\
            \midrule
            
            1. en   & 7,221,777 & 252,454 & 29   & 63. hi  & 170,759 & 890 & 192  & 107. tl  & 49,197 & 285 & 173  \\
            2. ceb  & 6,115,421 & 204     & 30k  & 64. mk  & 163,394 & 286 & 571  & 108. glk & 48,314 & 22  & 2196 \\
            3. de   & 3,142,055 & 32,658  & 96  & 65. zh-yue & 151,396 & 844 & 179  & 109. wuu & 48,292 & 90  & 537  \\
            4. fr   & 2,773,619 & 31,748  & 87   & 66. lv  & 144,792 & 413 & 351  & \cellcolor{highlightgreen} 110. ig  & \cellcolor{highlightgreen} 48,144 & \cellcolor{highlightgreen} 100 & \cellcolor{highlightgreen} 481  \\
            5. sv   & 2,626,957 & 3,951   & 665  & 67. la  & 142,119 & 199 & 714  & 111. diq & 42,814 & 82  & 522 \\
            6. nl   & 2,224,528 & 6,799   & 327  & 68. ast & 139,211 & 173 & 805  & \cellcolor{highlightgreen} 112. yo  & \cellcolor{highlightgreen} 38,847 & \cellcolor{highlightgreen} 73  & \cellcolor{highlightgreen} 532  \\
            7. es   & 2,130,236 & 35,720  & 60   & 69. af  & 130,084 & 277 & 470  & 113. ban & 38,609 & 72  & 536  \\
            8. ru   & 2,113,126 & 16,911  & 125  & 70. te  & 127,313 & 394 & 323  & 114. kn  & 35,537 & 164 & 217  \\
            9. it   & 1,981,280 & 25,331  & 78   & 71. sw  & 122,616 & 235 & 522  & 115. sco & 34,233 & 121  & 283 \\
            10. pl  & 1,704,089 & 8,715   & 196  & 72. tg  & 118,822 & 113 & 1052   & 116. als  & 31,774 & 91  & 349  \\
            11. arz & 1,633,164 & 284     & 5751 & 73. my  & 111,691 & 286 & 391  & 117. shn  & 31,624 & 27 & 1208  \\

            12. zh  & 1,548,047 & 13,958  & 111  & \cellcolor{green!18} 74. \cellcolor{green!18} ha  & \cellcolor{green!18} 106,959 & \cellcolor{green!18} 253 & \cellcolor{green!18} 423 & 118. gu  & 30,907 & 107  & 289 \\
            
            13. ja  & 1,513,046 & 24,086  & 63   & 75. sq & 106,194  & 341  & 311 & 119. ia  & 30,519 & 50  & 610 \\
            
            \bottomrule 
         \end{tabular}
        } 
        \caption{Summary of the languages with the (a) highest (b) low and (c) very low number of digital articles as tracked by Wikipedia (\url{https://meta.wikimedia.org/wiki/List_of_Wikipedias} accessed 09/08/26). In sub-table $b$ Hausa (ha) has just above 106K articles while both Igbo (ig) and Yorùbá (yo) in sub-table $c$ have just over 48K and about 38k articles, respectively. These numbers are very small compared to the population size and number of Internet users. Based on the ratio of the number of articles relative to active users contributing to the number of articles, there are few users contributing to digital content for the 3 languages.}
        \label{tab:wiki_articles_language}
    \end{table}


\paragraph{2. Linguistic Resource Enhancement}
This section discusses initiatives aimed at enhancing digital resources, tools, and collaborative efforts, particularly community-driven projects designed to advance NaijaNLP.  
Laudable initiatives such as the Alt-i, which aims to enhance NLP support for Nigerian languages \cite{adegbola2009building}, and comprehensive tools for facilitating downstream tasks across sub-Saharan languages \cite{chiarcos2011information} are required. 
Efforts to advance lexicographic resources include bilingual African language-French dictionaries \cite{enguehard2014computerization}, optimising stopwords extraction \citet{turki2024text}, tokeniser and an automatic speech system for predicting word pronunciation and segmentation for Nigerian Pidgin \cite{bigi2017developing}.  
Addressing the broader low-resource NLP landscape, \citet{costa2022no} explored strategies to narrow the performance gap between high-resource and low-resource languages. Their approach involves exploratory interviews with native speakers, data creation, and model development.  

\subsubsection{Mitigation Strategies and Future Prospects}
\label{sec:mitigation-strategies-and-prospects} 
Noting the aforementioned limitations, this section discusses strategies to mitigate challenges facing NaijaNLP and explores future prospects for enhancing the NLP tools.

\paragraph{1. Repurposing LLMs to Suit LRLs}
Pretrained models act as crucial starting points for fine-tuning on tasks such as text classification, machine translation, question answering, text generation, and summarisation. Together, embeddings and pretrained models underpin both language understanding and generation. However, base embeddings and language generation models have HRL-centric training data and evaluation protocols, which led researchers to resort to transfer learning and task-adaptive fine-tuning strategies \cite{gururangan2020don,muller2020being,alabi2022adapting,wang2023noise,oladipo2024backbones,abdou2024multilingual}. 
Moreover, benchmark studies often overlook the complexities inherent in NaijaNLP tasks \cite{adewumi2022vector,srivastava2022beyond,dossou2022afrolm,chowdhery2023palm,ojo2023good,liu2023graphmax,lawalcontextual,orok2024pharmacy,remy2024trans}. 
In response to this limitation, \citet{yousuf2024improving} propose a contrastive learning strategy to better align embeddings by optimising semantic distances between similar and dissimilar sentences.  
To address out-of-distribution scenarios, \citet{jiang2023low} propose a non-parametric approach that combines compression algorithms with k-nearest-neighbour classifiers, using compressor-based distance metrics \cite{fadaee2017data}.     
To alleviate for resource limitation, \citet{ogueji2021small} develop a multilingual language model (AfriBerta) trained on 11 low-resource languages. Models such as AfriBERTa demonstrate that small but high quality data approach based on similar languages has the potential of superior performance compared to joint training on large datasets with high-resource languages \cite{conneau2019unsupervised,ogueji2021small}.  
Despite such efforts, studies specifically addressing embeddings for low-resource Nigerian languages remain limited \cite{abdulmumin2019hauwe,alabi2020massive,yousuf2024improving}. 


\paragraph{Community Initiatives for Developing LRLs} 
As evidenced in Table~\ref{tab:contributed-multilingual-datasets1} and ~\ref{tab:contributed-multilingual-datasets2}, community-driven initiatives and collaborative efforts are crucial for advancing linguistic resources in NaijaNLP. Relevant projects, such as the African Languages Technology Initiative (Alt-i) \cite{adegbola2009building}, Low-Resource Human Language Technology (LoReHLT) \cite{christianson2018overview}, African Languages Lab \cite{issaka2026african}, BULB project \cite{adda2016breaking}, AfroLID \cite{adebara2022afrolid}, Masakhane\footnote{isiZulu's word translated as 'We build together' - \url{https://www.masakhane.io/}}, and Lacuna fund\footnote{\url{https://lacunafund.org/datasets/language/}} play a pivotal role in enriching low-resource languages. Major technology groups seldom prioritise the enrichment of linguistic resources for individual or regional LRLs, frequently addressing them as a homogeneous group or solely for evaluation \cite{dabre2020survey,muller2020being,jiang2023low,fadaee2017data}.   
To ensure that LRLs are not marginalised in the AI revolution, increased innovative strategies such as comprehensive digitisation of existing resources are critical. 
Recent global commitments, including AI for Development (AI4D) Funders Collaborative \cite{AI4D2024}, underscore the importance of enhancing the representation of African languages in LLMs. Efforts to transduce texts \cite{randell1998hausar}, enrich resources \cite{adegbola2009building}, and digitisation of indigenous poetry \cite{hausapoetry} are essential steps toward integrating LRLs into modern computational frameworks. 

Furthermore, high-impact conferences like ACL and ICLR\footnote{\url{https://www.aclweb.org/} and \url{https://iclr.cc/}} support these efforts by hosting workshops and shared tasks that unite researchers from academia, industry, and research laboratories to address specific NLP challenges. For instance, \emph{SemEval Shared Tasks} offers useful avenue for in-depth exploration of language semantics, with contributions in semantic textual relatedness \cite{salahudeen2024hausanlp,ousidhoum-etal-2024-semeval}, translation \cite{chen2021university,nowakowski2021adam,tran2021facebook,yousuf2024improving}, and sentiment analysis \cite{benlahbib2023nlp,raychawdhary2023seals_lab,ramanathan2023techssn}. Moreover, \emph{AfricaNLP Workshops}, co-located with prominent conferences like ICLR, foster focused dialogue and collaborative research on thematic issues in low-resource African languages.



\section{Conclusion} 
\label{sec:conclusion} 
Low-resource NLP continues to confront enduring challenges that require targeted, and innovative strategies. This study offered a comprehensive review of the state of NLP in Nigeria, focusing on its three major languages - Hausa, Igbo, and Yorùbá (collectively termed NaijaNLP). By examining the unique linguistic characteristics and resource constraints of these languages, our review provides actionable recommendations to advance and sustain NaijaNLP within the broader realm of low-resource NLP (LR-NLP) research. 
Despite the numerical predominance of LRLs, languages within NaijaNLP are often treated as a homogeneous group. We surmise that this practice obscures significant variations in resource availability and quality and may impede targeted technological development. To that effect, we argue for a more nuanced, language-specific approach to address the unique challenges faced by individual languages. The prevalent reliance on HRL-based models further highlights the need to invest in bespoke, indigenous models.  
Essentially, our investigation spans available linguistic resources, language particularities, computational tools, and collaborative initiatives, revealing significant progress alongside challenges. These challenges underscore the need for greater investment in high-quality data, robust computational tools, and enhanced community collaboration to fully harness the potential of LRLs in NLP applications. To address these challenges, we recommend: 

    \begin{enumerate}[label=\ding{\numexpr171+\value*}]
        \item Emphasising the critical role of formal grammar and linguistic analysis in developing reusable linguistic resources. 
        \item Prioritising the collection and annotation of larger, more diverse datasets that capture language particularities such as morphological and dialectal variations. 
        \item Investing in the creation of models tailored to the unique needs of each language, including the development of large language models built from scratch using indigenous data, rather than relying solely on models repurposed from high-resource languages. 
        \item Fostering collaborative research efforts and community engagement to ensure the sustained development of NLP resources for LRLs. 
    \end{enumerate} 
In summary, advancing NaijaNLP demands dedicated efforts to overcome linguistic, resource, and sociopolitical challenges. By prioritising language-specific approaches and fostering robust collaborative research, we can pave the way for more inclusive and effective NLP solutions for low-resource languages. 


\bibliographystyle{ACM-Reference-Format}
\bibliography{main}


\begin{thebibliography}{314}


\ifx \showCODEN    \undefined \def \showCODEN     #1{\unskip}     \fi
\ifx \showISBNx    \undefined \def \showISBNx     #1{\unskip}     \fi
\ifx \showISBNxiii \undefined \def \showISBNxiii  #1{\unskip}     \fi
\ifx \showISSN     \undefined \def \showISSN      #1{\unskip}     \fi
\ifx \showLCCN     \undefined \def \showLCCN      #1{\unskip}     \fi
\ifx \shownote     \undefined \def \shownote      #1{#1}          \fi
\ifx \showarticletitle \undefined \def \showarticletitle #1{#1}   \fi
\ifx \showURL      \undefined \def \showURL       {\relax}        \fi
\providecommand\bibfield[2]{#2}
\providecommand\bibinfo[2]{#2}
\providecommand\natexlab[1]{#1}
\providecommand\showeprint[2][]{arXiv:#2}

\bibitem[Abdou et~al\mbox{.}(2025)]%
        {abdou2025monitoring}
\bibfield{author}{\bibinfo{person}{Mohamed~Naira Abdou}, \bibinfo{person}{Imade Benelallam}, {and} \bibinfo{person}{Youcef Rahmani}.} \bibinfo{year}{2025}\natexlab{}.
\newblock \showarticletitle{Monitoring african geopolitics: a multilingual sentiment and public attention framework}.
\newblock \bibinfo{journal}{\emph{Applied Intelligence}} \bibinfo{volume}{55}, \bibinfo{number}{2} (\bibinfo{year}{2025}), \bibinfo{pages}{1--19}.
\newblock


\bibitem[Abdou~Mohamed et~al\mbox{.}(2024a)]%
        {abdou2024multilingual}
\bibfield{author}{\bibinfo{person}{Naira Abdou~Mohamed}, \bibinfo{person}{Anass Allak}, \bibinfo{person}{Kamel Gaanoun}, \bibinfo{person}{Imade Benelallam}, \bibinfo{person}{Zakarya Erraji}, {and} \bibinfo{person}{Abdessalam Bahafid}.} \bibinfo{year}{2024}\natexlab{a}.
\newblock \showarticletitle{Multilingual speech recognition initiative for African languages}.
\newblock \bibinfo{journal}{\emph{International Journal of Data Science and Analytics}} (\bibinfo{year}{2024}), \bibinfo{pages}{1--16}.
\newblock


\bibitem[Abdou~Mohamed et~al\mbox{.}(2024b)]%
        {abdou2024review}
\bibfield{author}{\bibinfo{person}{Naira Abdou~Mohamed}, \bibinfo{person}{Imade Benelallam}, \bibinfo{person}{Anass Allak}, {and} \bibinfo{person}{Kamel Gaanoun}.} \bibinfo{year}{2024}\natexlab{b}.
\newblock \showarticletitle{A Review on NLP Approaches for African Languages and Dialects}. In \bibinfo{booktitle}{\emph{International Conference on Advanced Technologies for Humanity}}. Springer, \bibinfo{pages}{207--213}.
\newblock


\bibitem[Abdoulaye(1992)]%
        {abdoulaye1992aspects}
\bibfield{author}{\bibinfo{person}{Mahamane~Laoualy Abdoulaye}.} \bibinfo{year}{1992}\natexlab{}.
\newblock \emph{\bibinfo{title}{Aspects of Hausa morphosyntax in role and reference grammar}}.
\newblock \bibinfo{thesistype}{Ph.\,D. Dissertation}. \bibinfo{school}{State Univ. of New York Buffalo}.
\newblock


\bibitem[Abdulkareem and Effiong(2016)]%
        {abdulkareem2016yorcall}
\bibfield{author}{\bibinfo{person}{Zainab Abdulkareem} {and} \bibinfo{person}{Emmanuel~E Effiong}.} \bibinfo{year}{2016}\natexlab{}.
\newblock \showarticletitle{YorCALL: Improving and Sustaining Yoruba Language through a Practical Iterative Learning Approach.}. In \bibinfo{booktitle}{\emph{OcRI}}. \bibinfo{pages}{1--5}.
\newblock


\bibitem[Abdullahi et~al\mbox{.}(2024)]%
        {abdullahi2024twitter}
\bibfield{author}{\bibinfo{person}{Habeeba~Ibraheem Abdullahi}, \bibinfo{person}{Muhammad~Aminu Ahmad}, {and} \bibinfo{person}{Khalid Haruna}.} \bibinfo{year}{2024}\natexlab{}.
\newblock \showarticletitle{Twitter sentiment analysis for Hausa abbreviations and acronyms}.
\newblock \bibinfo{journal}{\emph{Science World Journal}} \bibinfo{volume}{19}, \bibinfo{number}{1} (\bibinfo{year}{2024}), \bibinfo{pages}{101--104}.
\newblock


\bibitem[Abdullahi et~al\mbox{.}(2023)]%
        {abdullahi2023hausanlp}
\bibfield{author}{\bibinfo{person}{Saheed~Salahudeen Abdullahi}, \bibinfo{person}{Falalu~Ibrahim Lawan}, \bibinfo{person}{Ahmad~Mustapha Wali}, \bibinfo{person}{Amina~Abubakar Imam}, \bibinfo{person}{Aliyu~Rabiu Shuaibu}, \bibinfo{person}{Yusuf Aliyu}, \bibinfo{person}{Nur~Bala Rabiu}, \bibinfo{person}{Musa Bello}, \bibinfo{person}{Shamsuddeen~Umar Adamu}, \bibinfo{person}{Saminu~Mohammad Aliyu}, {et~al\mbox{.}}} \bibinfo{year}{2023}\natexlab{}.
\newblock \showarticletitle{HausaNLP at SemEval-2023 Task 12: Leveraging African Low Resource TweetData for Sentiment Analysis}. In \bibinfo{booktitle}{\emph{4th Workshop on African Natural Language Processing}}.
\newblock


\bibitem[Abdulmumin et~al\mbox{.}(2022)]%
        {abdulmumin2022hausa}
\bibfield{author}{\bibinfo{person}{Idris Abdulmumin}, \bibinfo{person}{Satya~Ranjan Dash}, \bibinfo{person}{Musa~Abdullahi Dawud}, \bibinfo{person}{Shantipriya Parida}, \bibinfo{person}{Shamsuddeen~Hassan Muhammad}, \bibinfo{person}{Ibrahim~Sa'id Ahmad}, \bibinfo{person}{Subhadarshi Panda}, \bibinfo{person}{Ond{\v{r}}ej Bojar}, \bibinfo{person}{Bashir~Shehu Galadanci}, {and} \bibinfo{person}{Bello~Shehu Bello}.} \bibinfo{year}{2022}\natexlab{}.
\newblock \showarticletitle{Hausa visual genome: A dataset for multi-modal English to Hausa machine translation}.
\newblock \bibinfo{journal}{\emph{arXiv preprint arXiv:2205.01133}} (\bibinfo{year}{2022}).
\newblock


\bibitem[Abdulmumin and Galadanci(2019)]%
        {abdulmumin2019hauwe}
\bibfield{author}{\bibinfo{person}{Idris Abdulmumin} {and} \bibinfo{person}{Bashir~Shehu Galadanci}.} \bibinfo{year}{2019}\natexlab{}.
\newblock \showarticletitle{hauwe: Hausa words embedding for natural language processing}. In \bibinfo{booktitle}{\emph{2019 2nd International Conference of the IEEE Nigeria Computer Chapter (NigeriaComputConf)}}. IEEE, \bibinfo{pages}{1--6}.
\newblock


\bibitem[Abdulmumin et~al\mbox{.}(2024)]%
        {abdulmumin2024correcting}
\bibfield{author}{\bibinfo{person}{Idris Abdulmumin}, \bibinfo{person}{Sthembiso Mkhwanazi}, \bibinfo{person}{Mahlatse~S Mbooi}, \bibinfo{person}{Shamsuddeen~Hassan Muhammad}, \bibinfo{person}{Ibrahim~Said Ahmad}, \bibinfo{person}{Neo Putini}, \bibinfo{person}{Miehleketo Mathebula}, \bibinfo{person}{Matimba Shingange}, \bibinfo{person}{Tajuddeen Gwadabe}, {and} \bibinfo{person}{Vukosi Marivate}.} \bibinfo{year}{2024}\natexlab{}.
\newblock \showarticletitle{Correcting FLORES Evaluation Dataset for Four African Languages}.
\newblock \bibinfo{journal}{\emph{arXiv preprint arXiv:2409.00626}} (\bibinfo{year}{2024}).
\newblock


\bibitem[Abegunde et~al\mbox{.}(2019)]%
        {abegunde1832design}
\bibfield{author}{\bibinfo{person}{O Abegunde}, \bibinfo{person}{AR Iyanda}, {and} \bibinfo{person}{DO Ninan}.} \bibinfo{year}{2019}\natexlab{}.
\newblock \showarticletitle{Design Issues in Sentiment Analysis for Yor{\`u}b{\'a} Written Text}.
\newblock  (\bibinfo{year}{2019}).
\newblock


\bibitem[Abiola et~al\mbox{.}(2014)]%
        {abiola2014web}
\bibfield{author}{\bibinfo{person}{OB Abiola}, \bibinfo{person}{AO Adetunmbi}, \bibinfo{person}{AI Fasiku}, {and} \bibinfo{person}{KA Olatunji}.} \bibinfo{year}{2014}\natexlab{}.
\newblock \showarticletitle{A web-based English to Yoruba noun-phrases machine translation system}.
\newblock \bibinfo{journal}{\emph{International Journal of English and Literature}} \bibinfo{volume}{5}, \bibinfo{number}{3} (\bibinfo{year}{2014}), \bibinfo{pages}{71--78}.
\newblock


\bibitem[Abolade(2026)]%
        {abolade2026owe}
\bibfield{author}{\bibinfo{person}{Daud Abolade}.} \bibinfo{year}{2026}\natexlab{}.
\newblock \showarticletitle{{\`O}WE-Voice: An Evaluation of Monolingual and Multilingual ASR Model Using Yoruba Proverb Speech Dataset}. In \bibinfo{booktitle}{\emph{Proceedings of the 7th Workshop on African Natural Language Processing (AfricaNLP 2026)}}. \bibinfo{pages}{82--87}.
\newblock


\bibitem[Abubakar et~al\mbox{.}(2021)]%
        {abubakar2021enhanced}
\bibfield{author}{\bibinfo{person}{Amina~Imam Abubakar}, \bibinfo{person}{Abubakar Roko}, \bibinfo{person}{Aminu~Muhammad Bui}, {and} \bibinfo{person}{Ibrahim Saidu}.} \bibinfo{year}{2021}\natexlab{}.
\newblock \showarticletitle{An enhanced feature acquisition for sentiment analysis of english and hausa tweets}.
\newblock \bibinfo{journal}{\emph{International Journal of Advanced Computer Science and Applications}} \bibinfo{volume}{12}, \bibinfo{number}{9} (\bibinfo{year}{2021}).
\newblock


\bibitem[Abubakar et~al\mbox{.}(2019)]%
        {abubakar2019hausa}
\bibfield{author}{\bibinfo{person}{Amina~Imam Abubakar}, \bibinfo{person}{Abubakar Roko}, \bibinfo{person}{AB Muhammad}, {and} \bibinfo{person}{Ibrahim Saidu}.} \bibinfo{year}{2019}\natexlab{}.
\newblock \showarticletitle{Hausa WordNet: An Electronic Lexical Resource}.
\newblock \bibinfo{journal}{\emph{Saudi Journal of Engineering and Technology}} \bibinfo{volume}{4}, \bibinfo{number}{8} (\bibinfo{year}{2019}), \bibinfo{pages}{279--285}.
\newblock


\bibitem[Abubakar et~al\mbox{.}(2024)]%
        {abubakar2024development}
\bibfield{author}{\bibinfo{person}{Abdulqahar~Mukhtar Abubakar}, \bibinfo{person}{Deepa Gupta}, {and} \bibinfo{person}{Susmitha Vekkot}.} \bibinfo{year}{2024}\natexlab{}.
\newblock \showarticletitle{Development of a diacritic-aware large vocabulary automatic speech recognition for Hausa language}.
\newblock \bibinfo{journal}{\emph{International Journal of Speech Technology}} \bibinfo{volume}{27}, \bibinfo{number}{3} (\bibinfo{year}{2024}), \bibinfo{pages}{687--700}.
\newblock


\bibitem[Adam et~al\mbox{.}(2024)]%
        {adam2024detection}
\bibfield{author}{\bibinfo{person}{Fatima~Muhammad Adam}, \bibinfo{person}{Abubakar~Yakubu Zandam}, {and} \bibinfo{person}{Isa Inuwa-Dutse}.} \bibinfo{year}{2024}\natexlab{}.
\newblock \showarticletitle{Detection and Analysis of Offensive Online Content in Hausa Language}.
\newblock \bibinfo{journal}{\emph{arXiv preprint arXiv:2501.08284}} (\bibinfo{year}{2024}).
\newblock


\bibitem[Adda et~al\mbox{.}(2016)]%
        {adda2016breaking}
\bibfield{author}{\bibinfo{person}{Gilles Adda}, \bibinfo{person}{Sebastian St{\"u}ker}, \bibinfo{person}{Martine Adda-Decker}, \bibinfo{person}{Odette Ambouroue}, \bibinfo{person}{Laurent Besacier}, \bibinfo{person}{David Blachon}, \bibinfo{person}{H{\'e}lene Bonneau-Maynard}, \bibinfo{person}{Pierre Godard}, \bibinfo{person}{Fatima Hamlaoui}, \bibinfo{person}{Dmitry Idiatov}, {et~al\mbox{.}}} \bibinfo{year}{2016}\natexlab{}.
\newblock \showarticletitle{Breaking the unwritten language barrier: The BULB project}.
\newblock \bibinfo{journal}{\emph{Procedia Computer Science}}  \bibinfo{volume}{81} (\bibinfo{year}{2016}), \bibinfo{pages}{8--14}.
\newblock


\bibitem[Adebara and Abdul-Mageed(2022)]%
        {adebara2022towards}
\bibfield{author}{\bibinfo{person}{Ife Adebara} {and} \bibinfo{person}{Muhammad Abdul-Mageed}.} \bibinfo{year}{2022}\natexlab{}.
\newblock \showarticletitle{Towards afrocentric NLP for African languages: Where we are and where we can go}.
\newblock \bibinfo{journal}{\emph{arXiv preprint arXiv:2203.08351}} (\bibinfo{year}{2022}).
\newblock


\bibitem[Adebara et~al\mbox{.}(2021)]%
        {adebara2021translating}
\bibfield{author}{\bibinfo{person}{Ife Adebara}, \bibinfo{person}{Muhammad Abdul-Mageed}, {and} \bibinfo{person}{Miikka Silfverberg}.} \bibinfo{year}{2021}\natexlab{}.
\newblock \showarticletitle{Translating the unseen? yoruba-english mt in low-resource, morphologically-unmarked settings}.
\newblock \bibinfo{journal}{\emph{arXiv preprint arXiv:2103.04225}} (\bibinfo{year}{2021}).
\newblock


\bibitem[Adebara et~al\mbox{.}(2022)]%
        {adebara2022afrolid}
\bibfield{author}{\bibinfo{person}{Ife Adebara}, \bibinfo{person}{AbdelRahim Elmadany}, \bibinfo{person}{Muhammad Abdul-Mageed}, {and} \bibinfo{person}{Alcides~Alcoba Inciarte}.} \bibinfo{year}{2022}\natexlab{}.
\newblock \showarticletitle{AfroLID: A neural language identification tool for African languages}.
\newblock \bibinfo{journal}{\emph{arXiv preprint arXiv:2210.11744}} (\bibinfo{year}{2022}).
\newblock


\bibitem[Adedjouma et~al\mbox{.}(2013)]%
        {adedjouma2013part}
\bibfield{author}{\bibinfo{person}{S{\`e}miyou~A Adedjouma}, \bibinfo{person}{John~OR Aoga}, {and} \bibinfo{person}{Mamoud~A Igue}.} \bibinfo{year}{2013}\natexlab{}.
\newblock \showarticletitle{Part-of-speech tagging of Yoruba standard, language of niger-congo family}.
\newblock \bibinfo{journal}{\emph{Res. Journal of Computer \& IT Sciences}} \bibinfo{volume}{1}, \bibinfo{number}{1} (\bibinfo{year}{2013}), \bibinfo{pages}{2--5}.
\newblock


\bibitem[Adegbola(2009)]%
        {adegbola2009building}
\bibfield{author}{\bibinfo{person}{Tunde Adegbola}.} \bibinfo{year}{2009}\natexlab{}.
\newblock \showarticletitle{Building capacities in human language technology for African languages}. In \bibinfo{booktitle}{\emph{Proceedings of the First Workshop on Language Technologies for African Languages}}. \bibinfo{pages}{53--58}.
\newblock


\bibitem[Adegbola(2016)]%
        {adegbola2016pattern}
\bibfield{author}{\bibinfo{person}{Tunde Adegbola}.} \bibinfo{year}{2016}\natexlab{}.
\newblock \showarticletitle{Pattern-based unsupervised induction of Yorùbá Morphology}. In \bibinfo{booktitle}{\emph{Proceedings of the 25th International conference companion on World wide web}}. \bibinfo{pages}{599--604}.
\newblock


\bibitem[Adegbola and Odilinye(2012)]%
        {adegbola2012quantifying}
\bibfield{author}{\bibinfo{person}{Tunde Adegbola} {and} \bibinfo{person}{Lydia~Uchechukwu Odilinye}.} \bibinfo{year}{2012}\natexlab{}.
\newblock \showarticletitle{Quantifying the effect of corpus size on the quality of automatic diacritization of Yor{\`u}b{\'a} texts}. In \bibinfo{booktitle}{\emph{Proc. SLTU 2012}}. \bibinfo{pages}{48--53}.
\newblock


\bibitem[Adegbola et~al\mbox{.}(2011)]%
        {adegbola2011localising}
\bibfield{author}{\bibinfo{person}{Tunde Adegbola}, \bibinfo{person}{Kola Owolabi}, {and} \bibinfo{person}{Tunji Odejobi}.} \bibinfo{year}{2011}\natexlab{}.
\newblock \showarticletitle{Localising for Yoruba: Experience, challenges and future direction}. In \bibinfo{booktitle}{\emph{Proceedings of conference on human language technology for development}}. \bibinfo{pages}{3--5}.
\newblock


\bibitem[Adegoke-Elijah et~al\mbox{.}(2023)]%
        {adegokedevelopment}
\bibfield{author}{\bibinfo{person}{A Adegoke-Elijah}, \bibinfo{person}{K Jimoh}, {and} \bibinfo{person}{A Alabi}.} \bibinfo{year}{2023}\natexlab{}.
\newblock \showarticletitle{DEVELOPMENT OF A XML-ENCODED MACHINE-READABLE DICTIONARY FOR YORUBA WORD SENSE DISAMBIGUATION}.
\newblock \bibinfo{journal}{\emph{UNIOSUN Journal of Engineering and Environmental Sciences}} (\bibinfo{year}{2023}).
\newblock


\bibitem[Adegoke-Elijah et~al\mbox{.}(2024)]%
        {adegokeestimating}
\bibfield{author}{\bibinfo{person}{Adenike Adegoke-Elijah}, \bibinfo{person}{Theresa Ojewumi}, {and} \bibinfo{person}{Kudirat Jimoh}.} \bibinfo{year}{2024}\natexlab{}.
\newblock \showarticletitle{ESTIMATING SEMANTIC SIMILARITY IN YORUBA SENTENCES USING PATH-BASED METRICS}.
\newblock  (\bibinfo{year}{2024}).
\newblock


\bibitem[Adegunlehin et~al\mbox{.}(2019)]%
        {adegunlehin2019investigation}
\bibfield{author}{\bibinfo{person}{EA Adegunlehin}, \bibinfo{person}{FO Asahiah}, {and} \bibinfo{person}{MT Onifade}.} \bibinfo{year}{2019}\natexlab{}.
\newblock \showarticletitle{Investigation of Feature Characteristics for Yoruba Named Entity Recognition System}. In \bibinfo{booktitle}{\emph{the Proceedings of 2019 AICTTRA Conference, Nigeria}}.
\newblock


\bibitem[Adelani(2022)]%
        {adelani2022natural}
\bibfield{author}{\bibinfo{person}{David~Ifeoluwa Adelani}.} \bibinfo{year}{2022}\natexlab{}.
\newblock \showarticletitle{Natural language processing for African languages}.
\newblock  (\bibinfo{year}{2022}).
\newblock


\bibitem[Adelani et~al\mbox{.}(2021)]%
        {adelani2021masakhaner}
\bibfield{author}{\bibinfo{person}{David~Ifeoluwa Adelani}, \bibinfo{person}{Jade Abbott}, \bibinfo{person}{Graham Neubig}, \bibinfo{person}{Daniel D’souza}, \bibinfo{person}{Julia Kreutzer}, \bibinfo{person}{Constantine Lignos}, \bibinfo{person}{Chester Palen-Michel}, \bibinfo{person}{Happy Buzaaba}, \bibinfo{person}{Shruti Rijhwani}, \bibinfo{person}{Sebastian Ruder}, {et~al\mbox{.}}} \bibinfo{year}{2021}\natexlab{}.
\newblock \showarticletitle{MasakhaNER: Named entity recognition for African languages}.
\newblock \bibinfo{journal}{\emph{Transactions of the Association for Computational Linguistics}}  \bibinfo{volume}{9} (\bibinfo{year}{2021}), \bibinfo{pages}{1116--1131}.
\newblock


\bibitem[Adelani et~al\mbox{.}(2020)]%
        {adelani2020distant}
\bibfield{author}{\bibinfo{person}{David~Ifeoluwa Adelani}, \bibinfo{person}{Michael~A Hedderich}, \bibinfo{person}{Dawei Zhu}, \bibinfo{person}{Esther van~den Berg}, {and} \bibinfo{person}{Dietrich Klakow}.} \bibinfo{year}{2020}\natexlab{}.
\newblock \showarticletitle{Distant Supervision and Noisy Label Learning for Low Resource Named Entity Recognition: A Study on Hausa and Yor$\backslash$ub$\backslash$'a}.
\newblock \bibinfo{journal}{\emph{arXiv preprint arXiv:2003.08370}} (\bibinfo{year}{2020}).
\newblock


\bibitem[Adelani et~al\mbox{.}(2023)]%
        {adelani2023masakhanews}
\bibfield{author}{\bibinfo{person}{David~Ifeoluwa Adelani}, \bibinfo{person}{Marek Masiak}, \bibinfo{person}{Israel~Abebe Azime}, \bibinfo{person}{Jesujoba Alabi}, \bibinfo{person}{Atnafu~Lambebo Tonja}, \bibinfo{person}{Christine Mwase}, \bibinfo{person}{Odunayo Ogundepo}, \bibinfo{person}{Bonaventure~FP Dossou}, \bibinfo{person}{Akintunde Oladipo}, \bibinfo{person}{Doreen Nixdorf}, {et~al\mbox{.}}} \bibinfo{year}{2023}\natexlab{}.
\newblock \showarticletitle{Masakhanews: News topic classification for african languages}.
\newblock \bibinfo{journal}{\emph{arXiv preprint arXiv:2304.09972}} (\bibinfo{year}{2023}).
\newblock


\bibitem[Adelani et~al\mbox{.}(2024)]%
        {adelani2024irokobench}
\bibfield{author}{\bibinfo{person}{David~Ifeoluwa Adelani}, \bibinfo{person}{Jessica Ojo}, \bibinfo{person}{Israel~Abebe Azime}, \bibinfo{person}{Jian~Yun Zhuang}, \bibinfo{person}{Jesujoba~O Alabi}, \bibinfo{person}{Xuanli He}, \bibinfo{person}{Millicent Ochieng}, \bibinfo{person}{Sara Hooker}, \bibinfo{person}{Andiswa Bukula}, \bibinfo{person}{En-Shiun~Annie Lee}, {et~al\mbox{.}}} \bibinfo{year}{2024}\natexlab{}.
\newblock \showarticletitle{IrokoBench: A New Benchmark for African Languages in the Age of Large Language Models}.
\newblock \bibinfo{journal}{\emph{arXiv preprint arXiv:2406.03368}} (\bibinfo{year}{2024}).
\newblock


\bibitem[Ademusire and Ninan(2023)]%
        {ademusire2023development}
\bibfield{author}{\bibinfo{person}{Adebisi~Joseph Ademusire} {and} \bibinfo{person}{Olufemi~D Ninan}.} \bibinfo{year}{2023}\natexlab{}.
\newblock \showarticletitle{Development of an annotated yoruba text corpus for automatic event extraction}.
\newblock  (\bibinfo{year}{2023}).
\newblock


\bibitem[Adeniji et~al\mbox{.}(2024)]%
        {adeniji2024framework}
\bibfield{author}{\bibinfo{person}{Adedapo~Bolaji Adeniji}, \bibinfo{person}{Taiwo Kolajo}, {and} \bibinfo{person}{Joshua~Babatunde Agbogun}.} \bibinfo{year}{2024}\natexlab{}.
\newblock \showarticletitle{A Framework for Yoruba Sentiment Lexicon Disambiguation}. In \bibinfo{booktitle}{\emph{2024 International Conference on Science, Engineering and Business for Driving Sustainable Development Goals (SEB4SDG)}}. IEEE, \bibinfo{pages}{1--6}.
\newblock


\bibitem[Adewole et~al\mbox{.}(2017)]%
        {adewole2017token}
\bibfield{author}{\bibinfo{person}{Lawrence~B Adewole}, \bibinfo{person}{Adebayo~O Adetunmbi}, \bibinfo{person}{Boniface~K Alese}, {and} \bibinfo{person}{Samuel~A Oluwadare}.} \bibinfo{year}{2017}\natexlab{}.
\newblock \showarticletitle{Token Validation in Automatic Corpus Gathering for Yoruba Language}.
\newblock \bibinfo{journal}{\emph{FUOYE Journal of Engineering and Technology}} \bibinfo{volume}{2}, \bibinfo{number}{1} (\bibinfo{year}{2017}), \bibinfo{pages}{4}.
\newblock


\bibitem[Adewole et~al\mbox{.}(2020)]%
        {adewole2020automatic}
\bibfield{author}{\bibinfo{person}{Lawrence~B Adewole}, \bibinfo{person}{Adebayo~O Adetunmbi}, \bibinfo{person}{Boniface~K Alese}, \bibinfo{person}{Samuel~A Oluwadare}, \bibinfo{person}{Oluwatoyin~B Abiola}, {and} \bibinfo{person}{Olaiya Folorunsho}.} \bibinfo{year}{2020}\natexlab{}.
\newblock \showarticletitle{Automatic Vowel Elision Resolution in Yor{\`u}b{\'a} Language}. In \bibinfo{booktitle}{\emph{Conference of the South African Institute of Computer Scientists and Information Technologists 2020}}. \bibinfo{pages}{126--133}.
\newblock


\bibitem[Adewumi(2022a)]%
        {adewumi2022vector}
\bibfield{author}{\bibinfo{person}{Oluwatosin Adewumi}.} \bibinfo{year}{2022}\natexlab{a}.
\newblock \emph{\bibinfo{title}{Vector representations of idioms in data-driven chatbots for robust assistance}}.
\newblock \bibinfo{thesistype}{Ph.\,D. Dissertation}. \bibinfo{school}{Lule{\aa} University of Technology}.
\newblock


\bibitem[Adewumi(2022b)]%
        {adewumi2022itakuroso}
\bibfield{author}{\bibinfo{person}{Tosin Adewumi}.} \bibinfo{year}{2022}\natexlab{b}.
\newblock \showarticletitle{Itak{\'u}roso: Exploiting cross-lingual transferability for natural language generation of dialogues in low-resource, african languages}. In \bibinfo{booktitle}{\emph{3rd Workshop on African Natural Language Processing}}.
\newblock


\bibitem[Adewumi et~al\mbox{.}(2023)]%
        {adewumi2023afriwoz}
\bibfield{author}{\bibinfo{person}{Tosin Adewumi}, \bibinfo{person}{Mofetoluwa Adeyemi}, \bibinfo{person}{Aremu Anuoluwapo}, \bibinfo{person}{Bukola Peters}, \bibinfo{person}{Happy Buzaaba}, \bibinfo{person}{Oyerinde Samuel}, \bibinfo{person}{Amina~Mardiyyah Rufai}, \bibinfo{person}{Benjamin Ajibade}, \bibinfo{person}{Tajudeen Gwadabe}, \bibinfo{person}{Mory Moussou~Koulibaly Traore}, {et~al\mbox{.}}} \bibinfo{year}{2023}\natexlab{}.
\newblock \showarticletitle{AfriWOZ: Corpus for Exploiting Cross-Lingual Transfer for Dialogue Generation in Low-Resource, African Languages}. In \bibinfo{booktitle}{\emph{2023 International Joint Conference on Neural Networks (IJCNN)}}. IEEE, \bibinfo{pages}{1--8}.
\newblock


\bibitem[Adewumi et~al\mbox{.}(2020)]%
        {adewumi2020challenge}
\bibfield{author}{\bibinfo{person}{Tosin~P Adewumi}, \bibinfo{person}{Foteini Liwicki}, {and} \bibinfo{person}{Marcus Liwicki}.} \bibinfo{year}{2020}\natexlab{}.
\newblock \showarticletitle{The challenge of diacritics in yoruba embeddings}.
\newblock \bibinfo{journal}{\emph{arXiv preprint arXiv:2011.07605}} (\bibinfo{year}{2020}).
\newblock


\bibitem[Adeyemi et~al\mbox{.}(2024)]%
        {adeyemi2024ciral}
\bibfield{author}{\bibinfo{person}{Mofetoluwa Adeyemi}, \bibinfo{person}{Akintunde Oladipo}, \bibinfo{person}{Xinyu Zhang}, \bibinfo{person}{David Alfonso-Hermelo}, \bibinfo{person}{Mehdi Rezagholizadeh}, \bibinfo{person}{Boxing Chen}, \bibinfo{person}{Abdul-Hakeem Omotayo}, \bibinfo{person}{Idris Abdulmumin}, \bibinfo{person}{Naome~A Etori}, \bibinfo{person}{Toyib~Babatunde Musa}, {et~al\mbox{.}}} \bibinfo{year}{2024}\natexlab{}.
\newblock \showarticletitle{CIRAL: A Test Collection for CLIR Evaluations in African Languages}. In \bibinfo{booktitle}{\emph{Proceedings of the 47th International ACM SIGIR Conference on Research and Development in Information Retrieval}}. \bibinfo{pages}{293--302}.
\newblock


\bibitem[Afolabi et~al\mbox{.}(2013)]%
        {afolabi2013development}
\bibfield{author}{\bibinfo{person}{Akin Afolabi}, \bibinfo{person}{Elijah Omidiora}, {and} \bibinfo{person}{Tayo Arulogun}.} \bibinfo{year}{2013}\natexlab{}.
\newblock \showarticletitle{Development of text to speech system for yoruba language}. In \bibinfo{booktitle}{\emph{Innovative Systems Design and Engineering—Special Issue of 2nd International Conference on Engineering and Technology Research}}, Vol.~\bibinfo{volume}{4}. \bibinfo{pages}{1--8}.
\newblock


\bibitem[Afolabi and Wahab(2013)]%
        {afolabi2013implementation}
\bibfield{author}{\bibinfo{person}{AO Afolabi} {and} \bibinfo{person}{AS Wahab}.} \bibinfo{year}{2013}\natexlab{}.
\newblock \showarticletitle{Implementation of Yoruba text-to-speech E-learning system}.
\newblock \bibinfo{journal}{\emph{International Journal of Engineering Research and Technology}} \bibinfo{volume}{2}, \bibinfo{number}{11} (\bibinfo{year}{2013}), \bibinfo{pages}{1055--1064}.
\newblock


\bibitem[Agbeyangi et~al\mbox{.}(2016)]%
        {agbeyangi2016web}
\bibfield{author}{\bibinfo{person}{AO Agbeyangi}, \bibinfo{person}{SI Eludiora}, {and} \bibinfo{person}{OA Popoola}.} \bibinfo{year}{2016}\natexlab{}.
\newblock \showarticletitle{Web-based Yoruba numeral translation system}.
\newblock \bibinfo{journal}{\emph{IAES International Journal of Artificial Intelligence (IJ-AI)}} \bibinfo{volume}{5}, \bibinfo{number}{4} (\bibinfo{year}{2016}), \bibinfo{pages}{127--134}.
\newblock


\bibitem[Ahia et~al\mbox{.}(2024)]%
        {ahia2024voices}
\bibfield{author}{\bibinfo{person}{Orevaoghene Ahia}, \bibinfo{person}{Anuoluwapo Aremu}, \bibinfo{person}{Diana Abagyan}, \bibinfo{person}{Hila Gonen}, \bibinfo{person}{David~Ifeoluwa Adelani}, \bibinfo{person}{Daud Abolade}, \bibinfo{person}{Noah~A Smith}, {and} \bibinfo{person}{Yulia Tsvetkov}.} \bibinfo{year}{2024}\natexlab{}.
\newblock \showarticletitle{Voices Unheard: NLP Resources and Models for Yor$\backslash$ub$\backslash$'a Regional Dialects}.
\newblock \bibinfo{journal}{\emph{arXiv preprint arXiv:2406.19564}} (\bibinfo{year}{2024}).
\newblock


\bibitem[AI4D(2024)]%
        {AI4D2024}
\bibfield{author}{\bibinfo{person}{AI~for Development Funders~Collaborative AI4D}.} \bibinfo{year}{2024}\natexlab{}.
\newblock \showarticletitle{Uniting with local partners to combat inequality through inclusive, responsible AI made by and for communities in the Global South}.
\newblock  (\bibinfo{year}{2024}).
\newblock


\bibitem[Aina and Akinwale(2023)]%
        {aina2023human}
\bibfield{author}{\bibinfo{person}{Akindele~Akanji Aina} {and} \bibinfo{person}{Adeola~Moyosoluwa Akinwale}.} \bibinfo{year}{2023}\natexlab{}.
\newblock \showarticletitle{Human Evaluation of Yorùbá Noun Ontology}.
\newblock \bibinfo{journal}{\emph{NIU Journal of Humanities}} \bibinfo{volume}{8}, \bibinfo{number}{1} (\bibinfo{year}{2023}), \bibinfo{pages}{33--51}.
\newblock


\bibitem[Aina and T{\'a}{\'\i}w{\`o}(2021)]%
        {aina2021ontology}
\bibfield{author}{\bibinfo{person}{Akindele~A Aina} {and} \bibinfo{person}{Oy{\`e}~P T{\'a}{\'\i}w{\`o}}.} \bibinfo{year}{2021}\natexlab{}.
\newblock \showarticletitle{ONTOLOGY ANNOTATION FOR NATURAL LANGUAGE DEVELOPMENT: A YOR{\`U}B{\'A} NOUN PRELIMINARY MODEL.}
\newblock \bibinfo{journal}{\emph{Journal of West African Languages}} \bibinfo{volume}{48}, \bibinfo{number}{1} (\bibinfo{year}{2021}).
\newblock


\bibitem[Ajagbe(2024)]%
        {ajagbe2024developing}
\bibfield{author}{\bibinfo{person}{Sunday~Adeola Ajagbe}.} \bibinfo{year}{2024}\natexlab{}.
\newblock \showarticletitle{Developing Nigeria Multilingual Languages Speech Datasets for Antenatal Orientation}. In \bibinfo{booktitle}{\emph{International Conference on Applied Informatics}}. Springer, \bibinfo{pages}{157--170}.
\newblock


\bibitem[Ajao et~al\mbox{.}(2022)]%
        {ajao2022yoruba}
\bibfield{author}{\bibinfo{person}{Jumoke Ajao}, \bibinfo{person}{Shakirat Yusuff}, {and} \bibinfo{person}{Abdulazeez Ajao}.} \bibinfo{year}{2022}\natexlab{}.
\newblock \showarticletitle{Yor{\`u}b{\'a} character recognition system using convolutional recurrent neural network}.
\newblock \bibinfo{journal}{\emph{Black Sea Journal of Engineering and Science}} \bibinfo{volume}{5}, \bibinfo{number}{4} (\bibinfo{year}{2022}), \bibinfo{pages}{151--157}.
\newblock


\bibitem[Ajao et~al\mbox{.}(2023)]%
        {ajao2023recurrent}
\bibfield{author}{\bibinfo{person}{Jumoke~F Ajao}, \bibinfo{person}{Oluwasogo~A Okunade}, {and} \bibinfo{person}{Abdulalzzez~O Ajao}.} \bibinfo{year}{2023}\natexlab{}.
\newblock \showarticletitle{Recurrent Neural Network (RNN) for Igbo Handwritten Character Recognition}.
\newblock \bibinfo{journal}{\emph{The Pacific Journal of Science and Technology}}  \bibinfo{volume}{24} (\bibinfo{year}{2023}).
\newblock


\bibitem[Akande and Opateye(2026)]%
        {akande2026low}
\bibfield{author}{\bibinfo{person}{Oyebola Akande} {and} \bibinfo{person}{Goodness Opateye}.} \bibinfo{year}{2026}\natexlab{}.
\newblock \showarticletitle{Low-Resource Machine Translation of Yoruba Text to Nigerian Pidgin Using NLLB-200 and Parameter-Efficient Fine-Tuning}.
\newblock \bibinfo{journal}{\emph{Cureus Journals}} \bibinfo{volume}{3}, \bibinfo{number}{1} (\bibinfo{year}{2026}).
\newblock


\bibitem[Akande et~al\mbox{.}(2022)]%
        {akande2022tweerify}
\bibfield{author}{\bibinfo{person}{Oluwatobi~Noah Akande}, \bibinfo{person}{Enemuo~Stephen Nnaemeka}, \bibinfo{person}{Oluwakemi~Christiana Abikoye}, \bibinfo{person}{Hakeem~Babalola Akande}, \bibinfo{person}{Abdullateef Balogun}, {and} \bibinfo{person}{Joyce Ayoola}.} \bibinfo{year}{2022}\natexlab{}.
\newblock \showarticletitle{TWEERIFY: a web-based sentiment analysis system using rule and deep learning techniques}. In \bibinfo{booktitle}{\emph{Proceedings of International Conference on Computational Intelligence and Data Engineering: ICCIDE 2021}}. Springer, \bibinfo{pages}{75--87}.
\newblock


\bibitem[Akinfaderin(2020)]%
        {akinfaderin2020hausamt}
\bibfield{author}{\bibinfo{person}{Adewale Akinfaderin}.} \bibinfo{year}{2020}\natexlab{}.
\newblock \showarticletitle{Hausamt v1. 0: Towards english-hausa neural machine translation}.
\newblock \bibinfo{journal}{\emph{arXiv preprint arXiv:2006.05014}} (\bibinfo{year}{2020}).
\newblock


\bibitem[Akintola and Ibiyemi(2017)]%
        {akintola2017machine}
\bibfield{author}{\bibinfo{person}{Abimbola Akintola} {and} \bibinfo{person}{Tunji Ibiyemi}.} \bibinfo{year}{2017}\natexlab{}.
\newblock \showarticletitle{Machine to Man Communication in Yor{\`u}b{\'a} Language}.
\newblock \bibinfo{journal}{\emph{Annal. Comput. Sci. Ser}} \bibinfo{volume}{15}, \bibinfo{number}{2} (\bibinfo{year}{2017}).
\newblock


\bibitem[Akinwonmi(2024)]%
        {akinwonmi2024rule}
\bibfield{author}{\bibinfo{person}{AE Akinwonmi}.} \bibinfo{year}{2024}\natexlab{}.
\newblock \showarticletitle{Rule-Induced Misanalysis of Nasal Syllables in Yoruba Declarative Syllabification Algorithm}.
\newblock \bibinfo{journal}{\emph{Journal of Sustainable Technology}} \bibinfo{volume}{13}, \bibinfo{number}{1} (\bibinfo{year}{2024}).
\newblock


\bibitem[Akpobi(2024)]%
        {akpobi2024yankari}
\bibfield{author}{\bibinfo{person}{Maro Akpobi}.} \bibinfo{year}{2024}\natexlab{}.
\newblock \showarticletitle{Yankari: A Monolingual Yoruba Dataset}.
\newblock \bibinfo{journal}{\emph{arXiv preprint arXiv:2412.03334}} (\bibinfo{year}{2024}).
\newblock


\bibitem[Alabi et~al\mbox{.}(2022)]%
        {alabi2022adapting}
\bibfield{author}{\bibinfo{person}{Jesujoba Alabi}, \bibinfo{person}{David~Ifeoluwa Adelani}, \bibinfo{person}{Marius Mosbach}, {and} \bibinfo{person}{Dietrich Klakow}.} \bibinfo{year}{2022}\natexlab{}.
\newblock \showarticletitle{Adapting pre-trained language models to African languages via multilingual adaptive fine-tuning}. In \bibinfo{booktitle}{\emph{Proceedings of the 29th international conference on computational linguistics}}. \bibinfo{pages}{4336--4349}.
\newblock


\bibitem[Alabi et~al\mbox{.}(2020)]%
        {alabi2020massive}
\bibfield{author}{\bibinfo{person}{Jesujoba Alabi}, \bibinfo{person}{Kwabena Amponsah-Kaakyire}, \bibinfo{person}{David Adelani}, {and} \bibinfo{person}{Cristina Espana-Bonet}.} \bibinfo{year}{2020}\natexlab{}.
\newblock \showarticletitle{Massive vs. curated embeddings for low-resourced languages: the case of Yor{\`u}b{\'a} and Twi}. In \bibinfo{booktitle}{\emph{Proceedings of the Twelfth Language Resources and Evaluation Conference}}. \bibinfo{pages}{2754--2762}.
\newblock


\bibitem[Alade et~al\mbox{.}(2019)]%
        {alade2019issues}
\bibfield{author}{\bibinfo{person}{SM Alade}, \bibinfo{person}{OD Ninan}, {and} \bibinfo{person}{OA Odejobi}.} \bibinfo{year}{2019}\natexlab{}.
\newblock \showarticletitle{ISSUES IN KNOWLEDGE REPRESENTATION IN YOR{\`U}B{\'A} NARRATIVES}.
\newblock \bibinfo{journal}{\emph{Annals. Computer Science Series}} \bibinfo{volume}{17}, \bibinfo{number}{2} (\bibinfo{year}{2019}).
\newblock


\bibitem[Aliero et~al\mbox{.}(2019)]%
        {aliero10taxonomy}
\bibfield{author}{\bibinfo{person}{Abubakar~Ahmad Aliero}, \bibinfo{person}{Dalhatu Muhammed}, {and} \bibinfo{person}{Abubakar Ibrahim}.} \bibinfo{year}{2019}\natexlab{}.
\newblock \showarticletitle{Taxonomy, Review and Research Challenges Of DNN-Based Text-To-Speech System for Hausa as Under-Resourced Language}.
\newblock \bibinfo{journal}{\emph{vol}}  \bibinfo{volume}{10} (\bibinfo{year}{2019}), \bibinfo{pages}{548--560}.
\newblock


\bibitem[Aliyu et~al\mbox{.}(2024)]%
        {aliyubeyond}
\bibfield{author}{\bibinfo{person}{Saminu~Mohammad Aliyu}, \bibinfo{person}{Gregory~Maksha Wajiga}, \bibinfo{person}{Muhammad Murtala}, {and} \bibinfo{person}{Lukman~Jibril Aliyu}.} \bibinfo{year}{2024}\natexlab{}.
\newblock \showarticletitle{Beyond English: Offensive Language Detection in Low-Resource Nigerian Languages}. In \bibinfo{booktitle}{\emph{5th Workshop on African Natural Language Processing}}.
\newblock


\bibitem[Amuda et~al\mbox{.}(2010)]%
        {amuda2010limited}
\bibfield{author}{\bibinfo{person}{Sulyman Amuda}, \bibinfo{person}{Hynek Bo{\v{r}}il}, \bibinfo{person}{Abhijeet Sangwan}, {and} \bibinfo{person}{John~HL Hansen}.} \bibinfo{year}{2010}\natexlab{}.
\newblock \showarticletitle{Limited resource speech recognition for Nigerian English}. In \bibinfo{booktitle}{\emph{2010 IEEE International Conference on Acoustics, Speech and Signal Processing}}. IEEE, \bibinfo{pages}{5090--5093}.
\newblock


\bibitem[Ana et~al\mbox{.}(2024)]%
        {ana2024ai}
\bibfield{author}{\bibinfo{person}{P Ana}, \bibinfo{person}{GI Emereonye}, \bibinfo{person}{AC Onuora}, \bibinfo{person}{CC Ukaegbu}, \bibinfo{person}{RC Aguwamba}, {and} \bibinfo{person}{PC Sunday}.} \bibinfo{year}{2024}\natexlab{}.
\newblock \showarticletitle{An AI System for the Detection of Hate Speech Encoded in Igbo Native}.
\newblock \bibinfo{journal}{\emph{International Research Journal of Engineering and Technology (IRJET)}} (\bibinfo{year}{2024}).
\newblock


\bibitem[Andrabi et~al\mbox{.}(2021)]%
        {andrabi2021review}
\bibfield{author}{\bibinfo{person}{Syed Abdul~Basit Andrabi} {et~al\mbox{.}}} \bibinfo{year}{2021}\natexlab{}.
\newblock \showarticletitle{A review of machine translation for south asian low resource languages}.
\newblock \bibinfo{journal}{\emph{Turkish Journal of Computer and Mathematics Education (TURCOMAT)}} \bibinfo{volume}{12}, \bibinfo{number}{5} (\bibinfo{year}{2021}), \bibinfo{pages}{1134--1147}.
\newblock


\bibitem[Ann(2002)]%
        {copestake2002implementing}
\bibfield{author}{\bibinfo{person}{Copestake Ann}.} \bibinfo{year}{2002}\natexlab{}.
\newblock \bibinfo{booktitle}{\emph{Implementing typed feature structure grammars}}. Vol.~\bibinfo{volume}{110}.
\newblock \bibinfo{publisher}{CSLI publications Stanford}.
\newblock


\bibitem[Aoga et~al\mbox{.}(2016)]%
        {aoga2016integration}
\bibfield{author}{\bibinfo{person}{John~OR Aoga}, \bibinfo{person}{Theophile~K Dagba}, {and} \bibinfo{person}{Codjo~C Fanou}.} \bibinfo{year}{2016}\natexlab{}.
\newblock \showarticletitle{Integration of Yoruba language into MaryTTS}.
\newblock \bibinfo{journal}{\emph{International Journal of Speech Technology}}  \bibinfo{volume}{19} (\bibinfo{year}{2016}), \bibinfo{pages}{151--158}.
\newblock


\bibitem[Aremu et~al\mbox{.}(2024)]%
        {aremunaijarc}
\bibfield{author}{\bibinfo{person}{Anuoluwapo Aremu}, \bibinfo{person}{Jesujoba~Oluwadara Alabi}, \bibinfo{person}{Daud Abolade}, \bibinfo{person}{Nkechinyere~Faith Aguobi}, \bibinfo{person}{Shamsuddeen~Hassan Muhammad}, {and} \bibinfo{person}{David~Ifeoluwa Adelani}.} \bibinfo{year}{2024}\natexlab{}.
\newblock \showarticletitle{NaijaRC: A Multi-choice Reading Comprehension Dataset for Nigerian Languages}. In \bibinfo{booktitle}{\emph{5th Workshop on African Natural Language Processing}}.
\newblock


\bibitem[Asahiah et~al\mbox{.}(2017)]%
        {asahiah2017restoring}
\bibfield{author}{\bibinfo{person}{Franklin~Oladiipo Asahiah}, \bibinfo{person}{Odetunji~Ajadi Odejobi}, {and} \bibinfo{person}{Emmanuel~Rotimi Adagunodo}.} \bibinfo{year}{2017}\natexlab{}.
\newblock \showarticletitle{Restoring tone-marks in standard Yor{\`u}b{\'a} electronic text: improved model}.
\newblock \bibinfo{journal}{\emph{Computer Science}} \bibinfo{volume}{18}, \bibinfo{number}{3} (\bibinfo{year}{2017}).
\newblock


\bibitem[Asahiah et~al\mbox{.}(2023)]%
        {asahiah2023diacritic}
\bibfield{author}{\bibinfo{person}{Franklin~Ol{\'a}di{\'\i}p{\`o} Asahiah}, \bibinfo{person}{Mary~Taiwo On{\'\i}f{\'a}d{\'e}}, \bibinfo{person}{Adekemisola~Olufunmilayo Asahiah}, \bibinfo{person}{Abayomi~Emmanuel Adegunlehin}, {and} \bibinfo{person}{Adekemi~Olawunmi Amoo}.} \bibinfo{year}{2023}\natexlab{}.
\newblock \showarticletitle{Diacritic-aware yor{\`u}b{\'a} spell checker}.
\newblock \bibinfo{journal}{\emph{Computer Science}}  \bibinfo{volume}{24} (\bibinfo{year}{2023}).
\newblock


\bibitem[Asubiaro(2015)]%
        {asubiaro2015statistical}
\bibfield{author}{\bibinfo{person}{Toluwase Asubiaro}.} \bibinfo{year}{2015}\natexlab{}.
\newblock \showarticletitle{Statistical patterns of diacritized and undiacritized yoruba texts}.
\newblock \bibinfo{journal}{\emph{International Journal of Computational Linguistics Research}} \bibinfo{volume}{6}, \bibinfo{number}{3} (\bibinfo{year}{2015}), \bibinfo{pages}{77--84}.
\newblock


\bibitem[Asubiaro(2021)]%
        {asubiaro2021evaluating}
\bibfield{author}{\bibinfo{person}{Toluwase Asubiaro}.} \bibinfo{year}{2021}\natexlab{}.
\newblock \showarticletitle{Evaluating the Availability of Resources, Research Hubs, and Financial Supports for Nigerian Languages Natural Language Processing Research/{\'E}valuation de la disponibilit{\'e} des ressources, des centres de recherche et du soutien financier pour la recherche sur le traitement du langage naturel des langues nig{\'e}rianes}.
\newblock \bibinfo{journal}{\emph{Canadian Journal of Information and Library Science}} \bibinfo{volume}{43}, \bibinfo{number}{3} (\bibinfo{year}{2021}), \bibinfo{pages}{269--290}.
\newblock


\bibitem[Asubiaro and Igwe(2021)]%
        {asubiaro2021state}
\bibfield{author}{\bibinfo{person}{Toluwase~Victor Asubiaro} {and} \bibinfo{person}{Ebelechukwu~Gloria Igwe}.} \bibinfo{year}{2021}\natexlab{}.
\newblock \showarticletitle{A State-of-the-Art Review of Nigerian Languages Natural Language Processing Research}.
\newblock \bibinfo{journal}{\emph{Developing Countries and Technology Inclusion in the 21st Century Information Society}} (\bibinfo{year}{2021}), \bibinfo{pages}{147--167}.
\newblock


\bibitem[Avetisyan and Broneske(2023)]%
        {avetisyan2023large}
\bibfield{author}{\bibinfo{person}{Hayastan Avetisyan} {and} \bibinfo{person}{David Broneske}.} \bibinfo{year}{2023}\natexlab{}.
\newblock \showarticletitle{Large Language Models and Low-Resource Languages: An Examination of Armenian NLP}.
\newblock \bibinfo{journal}{\emph{Findings of the Association for Computational Linguistics: IJCNLP-AACL 2023 (Findings)}} (\bibinfo{year}{2023}), \bibinfo{pages}{199--210}.
\newblock


\bibitem[Awwalu et~al\mbox{.}(2021)]%
        {awwalu2021corpus}
\bibfield{author}{\bibinfo{person}{Jamilu Awwalu}, \bibinfo{person}{Saleh~Elyakub Abdullahi}, {and} \bibinfo{person}{Abraham~Eseoghene Evwiekpaefe}.} \bibinfo{year}{2021}\natexlab{}.
\newblock \showarticletitle{A corpus based transformation-based learning for Hausa text parts of speech tagging}.
\newblock \bibinfo{journal}{\emph{International Journal of Computing and Digital Systems}}  \bibinfo{volume}{10} (\bibinfo{year}{2021}), \bibinfo{pages}{473--490}.
\newblock


\bibitem[Ayodeji et~al\mbox{.}(2020)]%
        {model2020accent}
\bibfield{author}{\bibinfo{person}{Olalekan~Salau Ayodeji}, \bibinfo{person}{David~Olowoyo Tilewa}, {and} \bibinfo{person}{Oluwole~Akintola Solomon}.} \bibinfo{year}{2020}\natexlab{}.
\newblock \showarticletitle{Accent Classification of the Three Major Nigerian Indigenous Languages Using 1D}.
\newblock \bibinfo{journal}{\emph{Advances in Computational Intelligence Techniques}} (\bibinfo{year}{2020}).
\newblock


\bibitem[Azubuike and Umeh(2024)]%
        {azubuike2024design}
\bibfield{author}{\bibinfo{person}{Prince~C Azubuike} {and} \bibinfo{person}{Innocent~I Umeh}.} \bibinfo{year}{2024}\natexlab{}.
\newblock \showarticletitle{Design and implementation of an automated web-based Igbo text analyzer using natural language processing (NLP) tools}.
\newblock \bibinfo{journal}{\emph{World Journal of Advanced Research and Reviews}} \bibinfo{volume}{23}, \bibinfo{number}{3} (\bibinfo{year}{2024}), \bibinfo{pages}{1036--1045}.
\newblock


\bibitem[Babatunde et~al\mbox{.}(2026a)]%
        {babatunde2026sentiment}
\bibfield{author}{\bibinfo{person}{Akinbowale~Nathaniel Babatunde}, \bibinfo{person}{Jumoke~Falilat Ajao}, \bibinfo{person}{Abdulrahman~Abdullahi Kayode}, \bibinfo{person}{Ronke~Seyi Babatunde}, \bibinfo{person}{Salman~Olatunji Isiaka}, \bibinfo{person}{Roseline~Oluwaseun Ogundokun}, {and} \bibinfo{person}{Cheng-Chi Lee}.} \bibinfo{year}{2026}\natexlab{a}.
\newblock \showarticletitle{Sentiment classification in the Yoruba language using multilingual Bidirectional Encoder Representations from Transformers (mBERT)}.
\newblock \bibinfo{journal}{\emph{International Journal of Speech Technology}} \bibinfo{volume}{29}, \bibinfo{number}{1} (\bibinfo{year}{2026}), \bibinfo{pages}{9}.
\newblock


\bibitem[Babatunde et~al\mbox{.}(2024)]%
        {babatunde2024speech}
\bibfield{author}{\bibinfo{person}{Akinbowale~Nathaniel Babatunde}, \bibinfo{person}{Ronke~Seyi Babatunde}, \bibinfo{person}{Bukola~Fatimah Balogun}, \bibinfo{person}{Emmanuel Umar}, \bibinfo{person}{Shuaib~Babatunde Mohammed}, \bibinfo{person}{Afeez~Adeshina Oke}, {and} \bibinfo{person}{Kolawole~Yusuf Obiwusi}.} \bibinfo{year}{2024}\natexlab{}.
\newblock \showarticletitle{Speech-to-Text Hybrid English to Yoruba SMS Translator}.
\newblock \bibinfo{journal}{\emph{Innovative Computing Review}} \bibinfo{volume}{4}, \bibinfo{number}{1} (\bibinfo{year}{2024}), \bibinfo{pages}{15--36}.
\newblock


\bibitem[Babatunde et~al\mbox{.}(2026b)]%
        {babatunde2026advanced}
\bibfield{author}{\bibinfo{person}{Akinbowale~Nathaniel Babatunde}, \bibinfo{person}{Bukola~Fatimah Balogun}, \bibinfo{person}{Sunday~Adeola Ajagbe}, \bibinfo{person}{Salman~Olatunji Isiaka}, \bibinfo{person}{Anthony~Taiwo Olajide}, \bibinfo{person}{Ijeoma~N Ezeji}, \bibinfo{person}{Pragasen Mudali}, {and} \bibinfo{person}{Matthew~Olusegun Adigun}.} \bibinfo{year}{2026}\natexlab{b}.
\newblock \showarticletitle{An advanced NLP and machine learning-based chatbot system for effective Igbo language communication}.
\newblock \bibinfo{journal}{\emph{Discover Artificial Intelligence}} (\bibinfo{year}{2026}).
\newblock


\bibitem[Bashir et~al\mbox{.}(2017)]%
        {bashir2017automatic}
\bibfield{author}{\bibinfo{person}{Muazzam Bashir}, \bibinfo{person}{Azilawati Rozaimee}, {and} \bibinfo{person}{Wan Malini~Wan Isa}.} \bibinfo{year}{2017}\natexlab{}.
\newblock \showarticletitle{Automatic Hausa Language Text Summarization Based on Feature Extraction using Na{\"\i}ve Bayes Model}.
\newblock \bibinfo{journal}{\emph{World Applied Science Journal}} \bibinfo{volume}{35}, \bibinfo{number}{9} (\bibinfo{year}{2017}), \bibinfo{pages}{2074--2080}.
\newblock


\bibitem[Bashir et~al\mbox{.}(2015)]%
        {bashir2015word}
\bibfield{author}{\bibinfo{person}{Muazzam Bashir}, \bibinfo{person}{Azilawati~Binti Rozaimee}, {and} \bibinfo{person}{Wan Malini Binti~Wan Isa}.} \bibinfo{year}{2015}\natexlab{}.
\newblock \showarticletitle{A word stemming algorithm for Hausa language}.
\newblock \bibinfo{journal}{\emph{no. June}} (\bibinfo{year}{2015}).
\newblock


\bibitem[Bender et~al\mbox{.}(2002)]%
        {bender2002grammar}
\bibfield{author}{\bibinfo{person}{Emily~M Bender}, \bibinfo{person}{Dan Flickinger}, {and} \bibinfo{person}{Stephan Oepen}.} \bibinfo{year}{2002}\natexlab{}.
\newblock \showarticletitle{The grammar matrix: An open-source starter-kit for the rapid development of cross-linguistically consistent broad-coverage precision grammars}. In \bibinfo{booktitle}{\emph{COLING-02: Grammar Engineering and Evaluation}}.
\newblock


\bibitem[Benito-Santiago et~al\mbox{.}(2022)]%
        {benito2022machine}
\bibfield{author}{\bibinfo{person}{Hermilo Benito-Santiago}, \bibinfo{person}{Diana~Margarita C{\'o}rdova-Esparza}, \bibinfo{person}{No{\'e}~Alejandro Castro-S{\'a}nchez}, {and} \bibinfo{person}{Ana-Marcela Herrera-Navarro}.} \bibinfo{year}{2022}\natexlab{}.
\newblock \showarticletitle{Machine Translation of Texts from Languages with Low Digital Resources: A Systematic Review}. In \bibinfo{booktitle}{\emph{Mexican International Conference on Artificial Intelligence}}. Springer, \bibinfo{pages}{41--56}.
\newblock


\bibitem[Benlahbib and Boumhidi(2023)]%
        {benlahbib2023nlp}
\bibfield{author}{\bibinfo{person}{Abdessamad Benlahbib} {and} \bibinfo{person}{Achraf Boumhidi}.} \bibinfo{year}{2023}\natexlab{}.
\newblock \showarticletitle{Nlp-lisac at semeval-2023 task 12: Sentiment analysis for tweets expressed in african languages via transformer-based models}. In \bibinfo{booktitle}{\emph{Proceedings of the 17th International Workshop on Semantic Evaluation (SemEval-2023)}}. \bibinfo{pages}{199--204}.
\newblock


\bibitem[Berthold(2009)]%
        {crysmann2009autosegmental}
\bibfield{author}{\bibinfo{person}{Crysmann Berthold}.} \bibinfo{year}{2009}\natexlab{}.
\newblock \showarticletitle{Autosegmental representations in an HPSG of Hausa}. In \bibinfo{booktitle}{\emph{Proceedings of the 2009 workshop on grammar engineering across frameworks (geaf 2009)}}. \bibinfo{pages}{28--36}.
\newblock


\bibitem[Bichi et~al\mbox{.}(2022)]%
        {bichi2022automatic}
\bibfield{author}{\bibinfo{person}{Abdulkadir~Abubakar Bichi}, \bibinfo{person}{Ruhaidah Samsudin}, {and} \bibinfo{person}{Rohayanti Hassan}.} \bibinfo{year}{2022}\natexlab{}.
\newblock \showarticletitle{Automatic construction of generic stop words list for Hausa text}.
\newblock \bibinfo{journal}{\emph{Indonesian Journal of Electrical Engineering and Computer Science}} \bibinfo{volume}{25}, \bibinfo{number}{3} (\bibinfo{year}{2022}), \bibinfo{pages}{1501--1507}.
\newblock


\bibitem[Bichi et~al\mbox{.}(2024)]%
        {bichi2024integrating}
\bibfield{author}{\bibinfo{person}{Abdulkadir~Abubakar Bichi}, \bibinfo{person}{Ruhaidah Samsudin}, {and} \bibinfo{person}{Rohayanti Hassan}.} \bibinfo{year}{2024}\natexlab{}.
\newblock \showarticletitle{Integrating clustering method with graph-based ranking for Hausa text multi-document summarisation}.
\newblock \bibinfo{journal}{\emph{International Journal of Society Systems Science}} \bibinfo{volume}{15}, \bibinfo{number}{1} (\bibinfo{year}{2024}), \bibinfo{pages}{59--72}.
\newblock


\bibitem[Bichi et~al\mbox{.}(2023)]%
        {bichi2023graph}
\bibfield{author}{\bibinfo{person}{Abdulkadir~Abubakar Bichi}, \bibinfo{person}{Ruhaidah Samsudin}, \bibinfo{person}{Rohayanti Hassan}, \bibinfo{person}{Layla Rasheed~Abdallah Hasan}, {and} \bibinfo{person}{Abubakar Ado~Rogo}.} \bibinfo{year}{2023}\natexlab{}.
\newblock \showarticletitle{Graph-based extractive text summarization method for Hausa text}.
\newblock \bibinfo{journal}{\emph{Plos one}} \bibinfo{volume}{18}, \bibinfo{number}{5} (\bibinfo{year}{2023}), \bibinfo{pages}{e0285376}.
\newblock


\bibitem[Bigi et~al\mbox{.}(2017)]%
        {bigi2017developing}
\bibfield{author}{\bibinfo{person}{Brigitte Bigi}, \bibinfo{person}{Bernard Caron}, {and} \bibinfo{person}{Oyelere~S Abiola}.} \bibinfo{year}{2017}\natexlab{}.
\newblock \showarticletitle{Developing resources for automated speech processing of the african language naija (nigerian pidgin)}. In \bibinfo{booktitle}{\emph{8th Language \& Technology Conference: Human Language Technologies as a Challenge for Computer Science and Linguistics}}. \bibinfo{pages}{441--445}.
\newblock


\bibitem[Bimba et~al\mbox{.}(2016)]%
        {bimba2016stemming}
\bibfield{author}{\bibinfo{person}{Andrew Bimba}, \bibinfo{person}{Norisma Idris}, \bibinfo{person}{Norazlina Khamis}, {and} \bibinfo{person}{Nurul Fazmidar~Mohd Noor}.} \bibinfo{year}{2016}\natexlab{}.
\newblock \showarticletitle{Stemming Hausa text: using affix-stripping rules and reference look-up}.
\newblock \bibinfo{journal}{\emph{Language Resources and Evaluation}}  \bibinfo{volume}{50} (\bibinfo{year}{2016}), \bibinfo{pages}{687--703}.
\newblock


\bibitem[Blasi et~al\mbox{.}(2017)]%
        {blasi2017grammars}
\bibfield{author}{\bibinfo{person}{Dami{\'a}n~E Blasi}, \bibinfo{person}{Susanne~Maria Michaelis}, {and} \bibinfo{person}{Martin Haspelmath}.} \bibinfo{year}{2017}\natexlab{}.
\newblock \showarticletitle{Grammars are robustly transmitted even during the emergence of creole languages}.
\newblock \bibinfo{journal}{\emph{Nature Human Behaviour}} \bibinfo{volume}{1}, \bibinfo{number}{10} (\bibinfo{year}{2017}), \bibinfo{pages}{723--729}.
\newblock


\bibitem[Brugnone et~al\mbox{.}(2024)]%
        {brugnone2024ought}
\bibfield{author}{\bibinfo{person}{Nathan Brugnone}, \bibinfo{person}{Noam Benkler}, \bibinfo{person}{Peter Revay}, \bibinfo{person}{Rebecca Myhre}, \bibinfo{person}{Scott Friedman}, \bibinfo{person}{Sonja Schmer-Galunder}, \bibinfo{person}{Steven Gray}, {and} \bibinfo{person}{James Gentile}.} \bibinfo{year}{2024}\natexlab{}.
\newblock \showarticletitle{Is from ought? A comparison of unsupervised methods for structuring values-based wisdom-of-crowds estimates}.
\newblock \bibinfo{journal}{\emph{Journal of Computational Social Science}} (\bibinfo{year}{2024}), \bibinfo{pages}{1--51}.
\newblock


\bibitem[Caron(2020)]%
        {caron2020methodological}
\bibfield{author}{\bibinfo{person}{Bernard Caron}.} \bibinfo{year}{2020}\natexlab{}.
\newblock \showarticletitle{Methodological and technical challenges of a corpus-based study of Naija}.
\newblock In \bibinfo{booktitle}{\emph{West African languages. Linguistic theory and communication}}. \bibinfo{publisher}{Wydawnictwa Uniwersytetu Warszawskiego}, \bibinfo{pages}{57--75}.
\newblock


\bibitem[Chen et~al\mbox{.}(2021)]%
        {chen2021university}
\bibfield{author}{\bibinfo{person}{Pinzhen Chen}, \bibinfo{person}{Jind{\v{r}}ich Helcl}, \bibinfo{person}{Ulrich Germann}, \bibinfo{person}{Laurie Burchell}, \bibinfo{person}{Nikolay Bogoychev}, \bibinfo{person}{Antonio~Valerio Miceli-Barone}, \bibinfo{person}{Jonas Waldendorf}, \bibinfo{person}{Alexandra Birch}, {and} \bibinfo{person}{Kenneth Heafield}.} \bibinfo{year}{2021}\natexlab{}.
\newblock \showarticletitle{The University of Edinburgh’s English-German and English-Hausa submissions to the WMT21 news translation task}. In \bibinfo{booktitle}{\emph{Proceedings of the Sixth Conference on Machine Translation}}. \bibinfo{pages}{104--109}.
\newblock


\bibitem[Chiarcos et~al\mbox{.}(2011)]%
        {chiarcos2011information}
\bibfield{author}{\bibinfo{person}{Christian Chiarcos}, \bibinfo{person}{Ines Fiedler}, \bibinfo{person}{Mira Grubic}, \bibinfo{person}{Katharina Hartmann}, \bibinfo{person}{Julia Ritz}, \bibinfo{person}{Anne Schwarz}, \bibinfo{person}{Amir Zeldes}, {and} \bibinfo{person}{Malte Zimmermann}.} \bibinfo{year}{2011}\natexlab{}.
\newblock \showarticletitle{Information structure in African languages: corpora and tools}.
\newblock \bibinfo{journal}{\emph{Language resources and evaluation}}  \bibinfo{volume}{45} (\bibinfo{year}{2011}), \bibinfo{pages}{361--374}.
\newblock


\bibitem[Chidiebere et~al\mbox{.}(2020)]%
        {chidiebere2020analysis}
\bibfield{author}{\bibinfo{person}{Ugwu Chidiebere}, \bibinfo{person}{Adegbola Tunde}, {et~al\mbox{.}}} \bibinfo{year}{2020}\natexlab{}.
\newblock \showarticletitle{Analysis and representation of Igbo text document for a text-based system}.
\newblock \bibinfo{journal}{\emph{arXiv preprint arXiv:2009.06376}} (\bibinfo{year}{2020}).
\newblock


\bibitem[Chioma and Nnadozie(2023)]%
        {chiomaweb}
\bibfield{author}{\bibinfo{person}{Odirichukwu~Jacinta Chioma} {and} \bibinfo{person}{Nnamdi~Reginald Nnadozie}.} \bibinfo{year}{2023}\natexlab{}.
\newblock \showarticletitle{Web-Based Igbo Thesaurus with Real-Time Retrieval}.
\newblock \bibinfo{journal}{\emph{Journal of Computer Science Engineering and Software Testing}} (\bibinfo{year}{2023}).
\newblock


\bibitem[Cho(2014)]%
        {cho2014properties}
\bibfield{author}{\bibinfo{person}{Kyunghyun Cho}.} \bibinfo{year}{2014}\natexlab{}.
\newblock \showarticletitle{On the properties of neural machine translation: Encoder-decoder approaches}.
\newblock \bibinfo{journal}{\emph{arXiv preprint arXiv:1409.1259}} (\bibinfo{year}{2014}).
\newblock


\bibitem[Chowdhery et~al\mbox{.}(2023)]%
        {chowdhery2023palm}
\bibfield{author}{\bibinfo{person}{Aakanksha Chowdhery}, \bibinfo{person}{Sharan Narang}, \bibinfo{person}{Jacob Devlin}, \bibinfo{person}{Maarten Bosma}, \bibinfo{person}{Gaurav Mishra}, \bibinfo{person}{Adam Roberts}, \bibinfo{person}{Paul Barham}, \bibinfo{person}{Hyung~Won Chung}, \bibinfo{person}{Charles Sutton}, \bibinfo{person}{Sebastian Gehrmann}, {et~al\mbox{.}}} \bibinfo{year}{2023}\natexlab{}.
\newblock \showarticletitle{Palm: Scaling language modeling with pathways}.
\newblock \bibinfo{journal}{\emph{Journal of Machine Learning Research}} \bibinfo{volume}{24}, \bibinfo{number}{240} (\bibinfo{year}{2023}), \bibinfo{pages}{1--113}.
\newblock


\bibitem[Christianson et~al\mbox{.}(2018)]%
        {christianson2018overview}
\bibfield{author}{\bibinfo{person}{Caitlin Christianson}, \bibinfo{person}{Jason Duncan}, {and} \bibinfo{person}{Boyan Onyshkevych}.} \bibinfo{year}{2018}\natexlab{}.
\newblock \showarticletitle{Overview of the DARPA LORELEI Program}.
\newblock \bibinfo{journal}{\emph{Machine Translation}}  \bibinfo{volume}{32} (\bibinfo{year}{2018}), \bibinfo{pages}{3--9}.
\newblock


\bibitem[Chukwuneke et~al\mbox{.}(2023)]%
        {chukwuneke2023igboner}
\bibfield{author}{\bibinfo{person}{Chiamaka~Ijeoma Chukwuneke}, \bibinfo{person}{Paul Rayson}, \bibinfo{person}{Ignatius Ezeani}, \bibinfo{person}{Mo El-Haj}, \bibinfo{person}{Doris~Chinedu Asogwa}, \bibinfo{person}{Chidimma~Lilian Okpalla}, {and} \bibinfo{person}{Chinedu~Emmanuel Mbonu}.} \bibinfo{year}{2023}\natexlab{}.
\newblock \showarticletitle{IGBONER 2.0: EXPANDING NAMED ENTITY RECOGNITION DATASETS VIA PROJECTION}. In \bibinfo{booktitle}{\emph{4th Workshop on African Natural Language Processing}}.
\newblock


\bibitem[Clark(2011)]%
        {clark2011tonal}
\bibfield{author}{\bibinfo{person}{Mary~M Clark}.} \bibinfo{year}{2011}\natexlab{}.
\newblock \bibinfo{booktitle}{\emph{The tonal system of Igbo}}. Vol.~\bibinfo{volume}{10}.
\newblock \bibinfo{publisher}{Walter de Gruyter}.
\newblock


\bibitem[Conneau(2019)]%
        {conneau2019unsupervised}
\bibfield{author}{\bibinfo{person}{A Conneau}.} \bibinfo{year}{2019}\natexlab{}.
\newblock \showarticletitle{Unsupervised cross-lingual representation learning at scale}.
\newblock \bibinfo{journal}{\emph{arXiv preprint arXiv:1911.02116}} (\bibinfo{year}{2019}).
\newblock


\bibitem[Costa-juss{\`a} et~al\mbox{.}(2022)]%
        {costa2022no}
\bibfield{author}{\bibinfo{person}{Marta~R Costa-juss{\`a}}, \bibinfo{person}{James Cross}, \bibinfo{person}{Onur {\c{C}}elebi}, \bibinfo{person}{Maha Elbayad}, \bibinfo{person}{Kenneth Heafield}, \bibinfo{person}{Kevin Heffernan}, \bibinfo{person}{Elahe Kalbassi}, \bibinfo{person}{Janice Lam}, \bibinfo{person}{Daniel Licht}, \bibinfo{person}{Jean Maillard}, {et~al\mbox{.}}} \bibinfo{year}{2022}\natexlab{}.
\newblock \showarticletitle{No language left behind: Scaling human-centered machine translation}.
\newblock \bibinfo{journal}{\emph{arXiv preprint arXiv:2207.04672}} (\bibinfo{year}{2022}).
\newblock


\bibitem[Crysmann(2012)]%
        {crysmann2012hag}
\bibfield{author}{\bibinfo{person}{Berthold Crysmann}.} \bibinfo{year}{2012}\natexlab{}.
\newblock \showarticletitle{HaG: A computational grammar of Hausa}. In \bibinfo{booktitle}{\emph{Selected proceedings of the 42nd annual conference on african linguistics (acal 42)}}. Citeseer, \bibinfo{pages}{321--337}.
\newblock


\bibitem[Culy(1996)]%
        {culy1996formal}
\bibfield{author}{\bibinfo{person}{Christopher Culy}.} \bibinfo{year}{1996}\natexlab{}.
\newblock \showarticletitle{Formal properties of natural language and linguistic theories}.
\newblock \bibinfo{journal}{\emph{Linguistics and Philosophy}} (\bibinfo{year}{1996}), \bibinfo{pages}{599--617}.
\newblock


\bibitem[Dabre et~al\mbox{.}(2020)]%
        {dabre2020survey}
\bibfield{author}{\bibinfo{person}{Raj Dabre}, \bibinfo{person}{Chenhui Chu}, {and} \bibinfo{person}{Anoop Kunchukuttan}.} \bibinfo{year}{2020}\natexlab{}.
\newblock \showarticletitle{A survey of multilingual neural machine translation}.
\newblock \bibinfo{journal}{\emph{ACM Computing Surveys (CSUR)}} \bibinfo{volume}{53}, \bibinfo{number}{5} (\bibinfo{year}{2020}), \bibinfo{pages}{1--38}.
\newblock


\bibitem[De~Pauw et~al\mbox{.}(2007)]%
        {de2007automatic}
\bibfield{author}{\bibinfo{person}{Guy De~Pauw}, \bibinfo{person}{Peter~W Wagacha}, {and} \bibinfo{person}{Gilles-Maurice De~Schryver}.} \bibinfo{year}{2007}\natexlab{}.
\newblock \showarticletitle{Automatic diacritic restoration for resource-scarce languages}. In \bibinfo{booktitle}{\emph{Text, Speech and Dialogue: 10th International Conference, TSD 2007, Pilsen, Czech Republic, September 3-7, 2007. Proceedings 10}}. Springer, \bibinfo{pages}{170--179}.
\newblock


\bibitem[Dossou and Emezue(2021)]%
        {dossou2021okwugb}
\bibfield{author}{\bibinfo{person}{Bonaventure~FP Dossou} {and} \bibinfo{person}{Chris~C Emezue}.} \bibinfo{year}{2021}\natexlab{}.
\newblock \showarticletitle{OkwuGb$\backslash$'e: End-to-End Speech Recognition for Fon and Igbo}.
\newblock \bibinfo{journal}{\emph{arXiv preprint arXiv:2103.07762}} (\bibinfo{year}{2021}).
\newblock


\bibitem[Dossou et~al\mbox{.}(2022)]%
        {dossou2022afrolm}
\bibfield{author}{\bibinfo{person}{Bonaventure~FP Dossou}, \bibinfo{person}{Atnafu~Lambebo Tonja}, \bibinfo{person}{Oreen Yousuf}, \bibinfo{person}{Salomey Osei}, \bibinfo{person}{Abigail Oppong}, \bibinfo{person}{Iyanuoluwa Shode}, \bibinfo{person}{Oluwabusayo~Olufunke Awoyomi}, {and} \bibinfo{person}{Chris~Chinenye Emezue}.} \bibinfo{year}{2022}\natexlab{}.
\newblock \showarticletitle{AfroLM: A self-active learning-based multilingual pretrained language model for 23 African languages}.
\newblock \bibinfo{journal}{\emph{arXiv preprint arXiv:2211.03263}} (\bibinfo{year}{2022}).
\newblock


\bibitem[Duong(2017)]%
        {duong2017natural}
\bibfield{author}{\bibinfo{person}{Long Duong}.} \bibinfo{year}{2017}\natexlab{}.
\newblock \showarticletitle{Natural language processing for resource-poor languages}.
\newblock \bibinfo{journal}{\emph{University of Melbourne}} (\bibinfo{year}{2017}).
\newblock


\bibitem[Edwards and Sefara(2023)]%
        {edwards2023text}
\bibfield{author}{\bibinfo{person}{Gareth~Reeve Edwards} {and} \bibinfo{person}{Tshephisho~Joseph Sefara}.} \bibinfo{year}{2023}\natexlab{}.
\newblock \showarticletitle{Text Summarisation for Low-resourced Languages, A review}. In \bibinfo{booktitle}{\emph{International Conference on Speech and Language Technologies for Low-resource Languages}}. Springer, \bibinfo{pages}{283--296}.
\newblock


\bibitem[Ekpenyong et~al\mbox{.}(2022)]%
        {ekpenyong2022towards}
\bibfield{author}{\bibinfo{person}{Moses~E Ekpenyong}, \bibinfo{person}{Aminu~A Suleiman}, {and} \bibinfo{person}{Murtala Salihu}.} \bibinfo{year}{2022}\natexlab{}.
\newblock \showarticletitle{Towards Massive Parallel Corpus Creation for Hausa-to-English Machine Translation}.
\newblock In \bibinfo{booktitle}{\emph{Current Issues in Descriptive Linguistics and Digital Humanities: A Festschrift in Honor of Professor Eno-Abasi Essien Urua}}. \bibinfo{publisher}{Springer}, \bibinfo{pages}{501--550}.
\newblock


\bibitem[Eludiora and Ayemonisan(2018)]%
        {eludiora2018computational}
\bibfield{author}{\bibinfo{person}{Safiriyu~Ijiyemi Eludiora} {and} \bibinfo{person}{OR Ayemonisan}.} \bibinfo{year}{2018}\natexlab{}.
\newblock \showarticletitle{Computational Morphological Analysis of Yor{\`u}b{\'a} Language Words}.
\newblock \bibinfo{journal}{\emph{IAES International Journal of Artificial Intelligence}} \bibinfo{volume}{7}, \bibinfo{number}{1} (\bibinfo{year}{2018}), \bibinfo{pages}{11}.
\newblock


\bibitem[Eludiora and Odejobi(2016)]%
        {eludiora2016development}
\bibfield{author}{\bibinfo{person}{Safiriyu~I Eludiora} {and} \bibinfo{person}{Odetunji~A Odejobi}.} \bibinfo{year}{2016}\natexlab{}.
\newblock \showarticletitle{Development of an English to Yoruba machine translator}.
\newblock \bibinfo{journal}{\emph{International Journal of Modern Education and Computer Science}} \bibinfo{volume}{8}, \bibinfo{number}{11} (\bibinfo{year}{2016}), \bibinfo{pages}{8}.
\newblock


\bibitem[Emmanuel(2021)]%
        {akinwonmi2021development}
\bibfield{author}{\bibinfo{person}{Akinwonmi~Akintoba Emmanuel}.} \bibinfo{year}{2021}\natexlab{}.
\newblock \showarticletitle{Development of a prosodic read speech syllabic corpus of the Yoruba language}.
\newblock \bibinfo{journal}{\emph{Development}} \bibinfo{volume}{7}, \bibinfo{number}{36} (\bibinfo{year}{2021}).
\newblock


\bibitem[Emmanuel and Andrew(2024)]%
        {emmanuelcurrent}
\bibfield{author}{\bibinfo{person}{Chesire Emmanuel} {and} \bibinfo{person}{Kipkebut Andrew}.} \bibinfo{year}{2024}\natexlab{}.
\newblock \showarticletitle{CURRENT STATE, CHALLENGES AND OPPORTUNITIES FOR NATURAL LANGUAGE PROCESSING RESEARCH AND DEVELOPMENT IN AFRICA: ASYSTEMATIC RE}.
\newblock  (\bibinfo{year}{2024}).
\newblock


\bibitem[Enguehard and Mangeot(2014)]%
        {enguehard2014computerization}
\bibfield{author}{\bibinfo{person}{Chantal Enguehard} {and} \bibinfo{person}{Mathieu Mangeot}.} \bibinfo{year}{2014}\natexlab{}.
\newblock \showarticletitle{Computerization of African languages-French dictionaries}.
\newblock \bibinfo{journal}{\emph{arXiv preprint arXiv:1405.5893}} (\bibinfo{year}{2014}).
\newblock


\bibitem[Ethnologue(2026)]%
        {globallang2025}
\bibfield{author}{\bibinfo{person}{Ethnologue}.} \bibinfo{year}{01-08-2026}\natexlab{}.
\newblock \showarticletitle{Languages of the World}.
\newblock  (\bibinfo{year}{01-08-2026}).
\newblock


\bibitem[Eze(1997)]%
        {eze1997aspects}
\bibfield{author}{\bibinfo{person}{Bethrand~Ejike Eze}.} \bibinfo{year}{1997}\natexlab{}.
\newblock \bibinfo{booktitle}{\emph{Aspects of language contact: A variationist perspective on codeswitching and borrowing in Igbo-English bilingual discourse.}}
\newblock \bibinfo{publisher}{University of Ottawa (Canada)}.
\newblock


\bibitem[Ezeani et~al\mbox{.}(2016)]%
        {ezeani2016automatic}
\bibfield{author}{\bibinfo{person}{Ignatius Ezeani}, \bibinfo{person}{Mark Hepple}, {and} \bibinfo{person}{Ikechukwu Onyenwe}.} \bibinfo{year}{2016}\natexlab{}.
\newblock \showarticletitle{Automatic restoration of diacritics for Igbo language}. In \bibinfo{booktitle}{\emph{Text, Speech, and Dialogue: 19th International Conference, TSD 2016, Brno, Czech Republic, September 12-16, 2016, Proceedings 19}}. Springer, \bibinfo{pages}{198--205}.
\newblock


\bibitem[Ezeani et~al\mbox{.}(2017)]%
        {ezeani2017lexical}
\bibfield{author}{\bibinfo{person}{Ignatius Ezeani}, \bibinfo{person}{MR Hepple}, {and} \bibinfo{person}{Ikechukwu Onyenwe}.} \bibinfo{year}{2017}\natexlab{}.
\newblock \showarticletitle{Lexical disambiguation of Igbo using diacritic restoration}. In \bibinfo{booktitle}{\emph{Proceedings of the 1st Workshop on Sense, Concept and Entity Representations and their Applications}}. Association for Computational Linguistics, \bibinfo{pages}{53--60}.
\newblock


\bibitem[Ezeani et~al\mbox{.}(2018a)]%
        {ezeani2018igbo}
\bibfield{author}{\bibinfo{person}{Ignatius Ezeani}, \bibinfo{person}{Mark Hepple}, \bibinfo{person}{Ikechukwu Onyenwe}, {and} \bibinfo{person}{Enemouh Chioma}.} \bibinfo{year}{2018}\natexlab{a}.
\newblock \showarticletitle{Igbo diacritic restoration using embedding models}. In \bibinfo{booktitle}{\emph{Proceedings of the 2018 Conference of the North American Chapter of the Association for Computational Linguistics: Student Research Workshop}}. \bibinfo{pages}{54--60}.
\newblock


\bibitem[Ezeani et~al\mbox{.}(2018b)]%
        {ezeani2018transferred}
\bibfield{author}{\bibinfo{person}{Ignatius Ezeani}, \bibinfo{person}{Ikechukwu Onyenwe}, {and} \bibinfo{person}{Mark Hepple}.} \bibinfo{year}{2018}\natexlab{b}.
\newblock \showarticletitle{Transferred embeddings for igbo similarity, analogy, and diacritic restoration tasks}. In \bibinfo{booktitle}{\emph{Proceedings of the Third Workshop on Semantic Deep Learning}}. \bibinfo{pages}{30--38}.
\newblock


\bibitem[Ezugwu et~al\mbox{.}(2023)]%
        {ezugwu2023machine}
\bibfield{author}{\bibinfo{person}{Absalom~E Ezugwu}, \bibinfo{person}{Olaide~N Oyelade}, \bibinfo{person}{Abiodun~M Ikotun}, \bibinfo{person}{Jeffery~O Agushaka}, {and} \bibinfo{person}{Yuh-Shan Ho}.} \bibinfo{year}{2023}\natexlab{}.
\newblock \showarticletitle{Machine learning research trends in Africa: a 30 years overview with bibliometric analysis review}.
\newblock \bibinfo{journal}{\emph{Archives of Computational Methods in Engineering}} \bibinfo{volume}{30}, \bibinfo{number}{7} (\bibinfo{year}{2023}), \bibinfo{pages}{4177--4207}.
\newblock


\bibitem[Fadaee et~al\mbox{.}(2017)]%
        {fadaee2017data}
\bibfield{author}{\bibinfo{person}{Marzieh Fadaee}, \bibinfo{person}{Arianna Bisazza}, {and} \bibinfo{person}{Christof Monz}.} \bibinfo{year}{2017}\natexlab{}.
\newblock \showarticletitle{Data augmentation for low-resource neural machine translation}.
\newblock \bibinfo{journal}{\emph{arXiv preprint arXiv:1705.00440}} (\bibinfo{year}{2017}).
\newblock


\bibitem[Fagbolu et~al\mbox{.}(2015)]%
        {fagbolu2015digital}
\bibfield{author}{\bibinfo{person}{Olutola Fagbolu}, \bibinfo{person}{Akinwale Ojoawo}, \bibinfo{person}{Kayode Ajibade}, {and} \bibinfo{person}{Boniface Alese}.} \bibinfo{year}{2015}\natexlab{}.
\newblock \showarticletitle{Digital yoruba corpus}.
\newblock \bibinfo{journal}{\emph{International Journal of Innovative Science, Engineering and Technology}} (\bibinfo{year}{2015}), \bibinfo{pages}{2348--7968}.
\newblock


\bibitem[Fagbolu et~al\mbox{.}(2016)]%
        {fagbolu2016applying}
\bibfield{author}{\bibinfo{person}{Olutola~Olaide Fagbolu}, \bibinfo{person}{Babatunde~Sunday Obalalu}, \bibinfo{person}{Samuel~S Udoh}, {and} \bibinfo{person}{Ibadan~Abeokuta Uyo}.} \bibinfo{year}{2016}\natexlab{}.
\newblock \showarticletitle{Applying rough set theory to Yorub{\'a} language translation}. In \bibinfo{booktitle}{\emph{International Conference on Advanced Trends in ICT and Management (ICAITM) 28th}}.
\newblock


\bibitem[Fan et~al\mbox{.}(2021)]%
        {fan2021beyond}
\bibfield{author}{\bibinfo{person}{Angela Fan}, \bibinfo{person}{Shruti Bhosale}, \bibinfo{person}{Holger Schwenk}, \bibinfo{person}{Zhiyi Ma}, \bibinfo{person}{Ahmed El-Kishky}, \bibinfo{person}{Siddharth Goyal}, \bibinfo{person}{Mandeep Baines}, \bibinfo{person}{Onur Celebi}, \bibinfo{person}{Guillaume Wenzek}, \bibinfo{person}{Vishrav Chaudhary}, {et~al\mbox{.}}} \bibinfo{year}{2021}\natexlab{}.
\newblock \showarticletitle{Beyond english-centric multilingual machine translation}.
\newblock \bibinfo{journal}{\emph{Journal of Machine Learning Research}} \bibinfo{volume}{22}, \bibinfo{number}{107} (\bibinfo{year}{2021}), \bibinfo{pages}{1--48}.
\newblock


\bibitem[Ferroggiaro(2018)]%
        {ferroggiaro2018social}
\bibfield{author}{\bibinfo{person}{W Ferroggiaro}.} \bibinfo{year}{2018}\natexlab{}.
\newblock \bibinfo{title}{Social media and conflict in Nigeria: A lexicon of hate speech terms}.
\newblock


\bibitem[Finkel and Odejobi(2009)]%
        {finkel2009computational}
\bibfield{author}{\bibinfo{person}{Raphael Finkel} {and} \bibinfo{person}{Odetunji~Ajadi Odejobi}.} \bibinfo{year}{2009}\natexlab{}.
\newblock \showarticletitle{A computational approach to Yorub{\'a} morphology}. In \bibinfo{booktitle}{\emph{Proceedings of the First Workshop on Language Technologies for African Languages}}. \bibinfo{pages}{25--31}.
\newblock


\bibitem[Folajimi and Omonayin(2012)]%
        {folajimi2012using}
\bibfield{author}{\bibinfo{person}{Yetunde~O Folajimi} {and} \bibinfo{person}{Isaac Omonayin}.} \bibinfo{year}{2012}\natexlab{}.
\newblock \showarticletitle{Using Statistical Machine Translation (SMT) as a Language Translation Tool for Understanding Yoruba Language}.
\newblock  (\bibinfo{year}{2012}).
\newblock


\bibitem[Frohmann et~al\mbox{.}(2024)]%
        {frohmann2024segment}
\bibfield{author}{\bibinfo{person}{Markus Frohmann}, \bibinfo{person}{Igor Sterner}, \bibinfo{person}{Ivan Vuli{\'c}}, \bibinfo{person}{Benjamin Minixhofer}, {and} \bibinfo{person}{Markus Schedl}.} \bibinfo{year}{2024}\natexlab{}.
\newblock \showarticletitle{Segment Any Text: A universal approach for robust, efficient and adaptable sentence segmentation}.
\newblock \bibinfo{journal}{\emph{arXiv preprint arXiv:2406.16678}} (\bibinfo{year}{2024}).
\newblock


\bibitem[Gasparini et~al\mbox{.}(2026)]%
        {gasparini2026automated}
\bibfield{author}{\bibinfo{person}{Loretta Gasparini}, \bibinfo{person}{Daniel Shaw}, \bibinfo{person}{Thayabaran Kathiresan}, \bibinfo{person}{Devon Borysiewicz}, \bibinfo{person}{Geoffrey~W Stuart}, \bibinfo{person}{Adam~P Vogel}, \bibinfo{person}{Cross linguistic POS~Consortium}, {et~al\mbox{.}}} \bibinfo{year}{2026}\natexlab{}.
\newblock \showarticletitle{Automated parts-of-speech tagging in 52 language varieties: benchmarking towards establishing suitability for multilingual clinical language analysis}.
\newblock \bibinfo{journal}{\emph{Natural Language Processing Journal}} (\bibinfo{year}{2026}), \bibinfo{pages}{100226}.
\newblock


\bibitem[Gebru et~al\mbox{.}(2021)]%
        {gebru2021datasheets}
\bibfield{author}{\bibinfo{person}{Timnit Gebru}, \bibinfo{person}{Jamie Morgenstern}, \bibinfo{person}{Briana Vecchione}, \bibinfo{person}{Jennifer~Wortman Vaughan}, \bibinfo{person}{Hanna Wallach}, \bibinfo{person}{Hal~Daum{\'e} Iii}, {and} \bibinfo{person}{Kate Crawford}.} \bibinfo{year}{2021}\natexlab{}.
\newblock \showarticletitle{Datasheets for datasets}.
\newblock \bibinfo{journal}{\emph{Commun. ACM}} \bibinfo{volume}{64}, \bibinfo{number}{12} (\bibinfo{year}{2021}), \bibinfo{pages}{86--92}.
\newblock


\bibitem[Ghafoor et~al\mbox{.}(2021)]%
        {ghafoor2021impact}
\bibfield{author}{\bibinfo{person}{Abdul Ghafoor}, \bibinfo{person}{Ali~Shariq Imran}, \bibinfo{person}{Sher~Muhammad Daudpota}, \bibinfo{person}{Zenun Kastrati}, \bibinfo{person}{Rakhi Batra}, \bibinfo{person}{Mudasir~Ahmad Wani}, {et~al\mbox{.}}} \bibinfo{year}{2021}\natexlab{}.
\newblock \showarticletitle{The impact of translating resource-rich datasets to low-resource languages through multi-lingual text processing}.
\newblock \bibinfo{journal}{\emph{IEEE Access}}  \bibinfo{volume}{9} (\bibinfo{year}{2021}), \bibinfo{pages}{124478--124490}.
\newblock


\bibitem[Gibbon et~al\mbox{.}(2012)]%
        {gibbonmarketspeak}
\bibfield{author}{\bibinfo{person}{Dafydd Gibbon}, \bibinfo{person}{Ugonna Duruibe}, {and} \bibinfo{person}{Jolanta Bachan}.} \bibinfo{year}{2012}\natexlab{}.
\newblock \showarticletitle{‘MARKETSPEAK’IN IGBO: ASpeech SYNTHESIS TRAINING PROJECT}.
\newblock \bibinfo{journal}{\emph{Vili, Zulu, Xhosa, Afrikaans, English, Swati, Ndebele, Punu, Shona, Tswana, Sotho, Sepedi, Obamba}} (\bibinfo{year}{2012}), \bibinfo{pages}{339}.
\newblock


\bibitem[Girija et~al\mbox{.}(2023)]%
        {girija2023analysis}
\bibfield{author}{\bibinfo{person}{VR Girija}, \bibinfo{person}{T Sudha}, {and} \bibinfo{person}{Riboy Cheriyan}.} \bibinfo{year}{2023}\natexlab{}.
\newblock \showarticletitle{Analysis of Sentiments in Low Resource Languages: Challenges and Solutions}. In \bibinfo{booktitle}{\emph{2023 IEEE International Conference on Recent Advances in Systems Science and Engineering (RASSE)}}. IEEE, \bibinfo{pages}{1--6}.
\newblock


\bibitem[Godslove and Nayak(2024)]%
        {godslove2024trilingual}
\bibfield{author}{\bibinfo{person}{Julius~Femi Godslove} {and} \bibinfo{person}{Ajit~Kumar Nayak}.} \bibinfo{year}{2024}\natexlab{}.
\newblock \showarticletitle{Trilingual conversational intent decoding for response retrieval}.
\newblock \bibinfo{journal}{\emph{Knowledge and Information Systems}} \bibinfo{volume}{66}, \bibinfo{number}{1} (\bibinfo{year}{2024}), \bibinfo{pages}{535--556}.
\newblock


\bibitem[Goldsmith(1976)]%
        {goldsmith1976autosegmental}
\bibfield{author}{\bibinfo{person}{JOHN Goldsmith}.} \bibinfo{year}{1976}\natexlab{}.
\newblock \showarticletitle{Autosegmental phonology}.
\newblock \bibinfo{journal}{\emph{Indiana University}} (\bibinfo{year}{1976}).
\newblock


\bibitem[Goyal et~al\mbox{.}(2022)]%
        {goyal2022flores}
\bibfield{author}{\bibinfo{person}{Naman Goyal}, \bibinfo{person}{Cynthia Gao}, \bibinfo{person}{Vishrav Chaudhary}, \bibinfo{person}{Peng-Jen Chen}, \bibinfo{person}{Guillaume Wenzek}, \bibinfo{person}{Da Ju}, \bibinfo{person}{Sanjana Krishnan}, \bibinfo{person}{Marc’Aurelio Ranzato}, \bibinfo{person}{Francisco Guzm{\'a}n}, {and} \bibinfo{person}{Angela Fan}.} \bibinfo{year}{2022}\natexlab{}.
\newblock \showarticletitle{The flores-101 evaluation benchmark for low-resource and multilingual machine translation}.
\newblock \bibinfo{journal}{\emph{Transactions of the Association for Computational Linguistics}}  \bibinfo{volume}{10} (\bibinfo{year}{2022}), \bibinfo{pages}{522--538}.
\newblock


\bibitem[Graff and Finch(1994)]%
        {graff1994multilingual}
\bibfield{author}{\bibinfo{person}{David Graff} {and} \bibinfo{person}{Rebecca Finch}.} \bibinfo{year}{1994}\natexlab{}.
\newblock \showarticletitle{Multilingual text resources at the linguistic data consortium}. In \bibinfo{booktitle}{\emph{Human Language Technology: Proceedings of a Workshop held at Plainsboro, New Jersey, March 8-11, 1994}}.
\newblock


\bibitem[G{\"u}nther and Rinaldi(2022)]%
        {gunther2022language}
\bibfield{author}{\bibinfo{person}{Fritz G{\"u}nther} {and} \bibinfo{person}{Luca Rinaldi}.} \bibinfo{year}{2022}\natexlab{}.
\newblock \showarticletitle{Language statistics as a window into mental representations}.
\newblock \bibinfo{journal}{\emph{Scientific Reports}} \bibinfo{volume}{12}, \bibinfo{number}{1} (\bibinfo{year}{2022}), \bibinfo{pages}{8043}.
\newblock


\bibitem[Gururangan et~al\mbox{.}(2020)]%
        {gururangan2020don}
\bibfield{author}{\bibinfo{person}{Suchin Gururangan}, \bibinfo{person}{Ana Marasovi{\'c}}, \bibinfo{person}{Swabha Swayamdipta}, \bibinfo{person}{Kyle Lo}, \bibinfo{person}{Iz Beltagy}, \bibinfo{person}{Doug Downey}, {and} \bibinfo{person}{Noah~A Smith}.} \bibinfo{year}{2020}\natexlab{}.
\newblock \showarticletitle{Don't stop pretraining: Adapt language models to domains and tasks}.
\newblock \bibinfo{journal}{\emph{arXiv preprint arXiv:2004.10964}} (\bibinfo{year}{2020}).
\newblock


\bibitem[Gutkin et~al\mbox{.}(2020)]%
        {gutkin2020developing}
\bibfield{author}{\bibinfo{person}{Alexander Gutkin}, \bibinfo{person}{Isin Demirsahin}, \bibinfo{person}{Oddur Kjartansson}, \bibinfo{person}{Clara~E Rivera}, {and} \bibinfo{person}{K{\'o}l{\'a} T{\'u}b{\`o}s{\'u}n}.} \bibinfo{year}{2020}\natexlab{}.
\newblock \showarticletitle{Developing an open-source corpus of yoruba speech}.
\newblock  (\bibinfo{year}{2020}).
\newblock


\bibitem[Haddow et~al\mbox{.}(2022)]%
        {haddow2022survey}
\bibfield{author}{\bibinfo{person}{Barry Haddow}, \bibinfo{person}{Rachel Bawden}, \bibinfo{person}{Antonio Valerio~Miceli Barone}, \bibinfo{person}{Jind{\v{r}}ich Helcl}, {and} \bibinfo{person}{Alexandra Birch}.} \bibinfo{year}{2022}\natexlab{}.
\newblock \showarticletitle{Survey of low-resource machine translation}.
\newblock \bibinfo{journal}{\emph{Computational Linguistics}} \bibinfo{volume}{48}, \bibinfo{number}{3} (\bibinfo{year}{2022}), \bibinfo{pages}{673--732}.
\newblock


\bibitem[Haruna et~al\mbox{.}(2021)]%
        {haruna2021hausa}
\bibfield{author}{\bibinfo{person}{Usman Haruna}, \bibinfo{person}{Umar~Sunusi Maitalata}, \bibinfo{person}{Murtala Mohammed}, {and} \bibinfo{person}{Jaafar~Zubairu Maitama}.} \bibinfo{year}{2021}\natexlab{}.
\newblock \showarticletitle{Hausa Intelligence Chatbot System}. In \bibinfo{booktitle}{\emph{Information and Communication Technology and Applications: Third International Conference, ICTA 2020, Minna, Nigeria, November 24--27, 2020, Revised Selected Papers 3}}. Springer, \bibinfo{pages}{206--219}.
\newblock


\bibitem[Hatami et~al\mbox{.}(2024)]%
        {hatami2024english}
\bibfield{author}{\bibinfo{person}{Ali Hatami}, \bibinfo{person}{Shubhanker Banerjee}, \bibinfo{person}{Mihael Arcan}, \bibinfo{person}{Bharathi Chakravarthi}, \bibinfo{person}{Paul Buitelaar}, {and} \bibinfo{person}{John Mccrae}.} \bibinfo{year}{2024}\natexlab{}.
\newblock \showarticletitle{English-to-low-resource translation: A multimodal approach for hindi, malayalam, bengali, and hausa}. In \bibinfo{booktitle}{\emph{Proceedings of the Ninth Conference on Machine Translation}}. \bibinfo{pages}{815--822}.
\newblock


\bibitem[Hedderich(2022)]%
        {hedderich2022weak}
\bibfield{author}{\bibinfo{person}{Michael~Aloys Hedderich}.} \bibinfo{year}{2022}\natexlab{}.
\newblock \showarticletitle{Weak supervision and label noise handling for Natural language processing in low-resource scenarios}.
\newblock  (\bibinfo{year}{2022}).
\newblock


\bibitem[Hedderich et~al\mbox{.}(2020)]%
        {hedderich2020survey}
\bibfield{author}{\bibinfo{person}{Michael~A Hedderich}, \bibinfo{person}{Lukas Lange}, \bibinfo{person}{Heike Adel}, \bibinfo{person}{Jannik Str{\"o}tgen}, {and} \bibinfo{person}{Dietrich Klakow}.} \bibinfo{year}{2020}\natexlab{}.
\newblock \showarticletitle{A survey on recent approaches for natural language processing in low-resource scenarios}.
\newblock \bibinfo{journal}{\emph{arXiv preprint arXiv:2010.12309}} (\bibinfo{year}{2020}).
\newblock


\bibitem[Henry et~al\mbox{.}(2020)]%
        {henry2020performance}
\bibfield{author}{\bibinfo{person}{Odikwa Henry} {et~al\mbox{.}}} \bibinfo{year}{2020}\natexlab{}.
\newblock \showarticletitle{The Performance Evaluation of an Igbo Text-Based Intelligent System}.
\newblock \bibinfo{journal}{\emph{International Journal of Applied Information Systems}} \bibinfo{volume}{12}, \bibinfo{number}{30} (\bibinfo{year}{2020}), \bibinfo{pages}{1--5}.
\newblock


\bibitem[Hu et~al\mbox{.}(2024)]%
        {hu2024review}
\bibfield{author}{\bibinfo{person}{Songbo Hu}, \bibinfo{person}{Abigail Oppong}, \bibinfo{person}{Ebele Mogo}, \bibinfo{person}{Anna Barford}, \bibinfo{person}{Giulia Occhini}, \bibinfo{person}{Charlotte Collins}, {and} \bibinfo{person}{Anna Korhonen}.} \bibinfo{year}{2024}\natexlab{}.
\newblock \showarticletitle{Review Protocol: A Scoping Review of Natural Language Processing Technologies for Public Health in Africa}.
\newblock \bibinfo{journal}{\emph{medRxiv}} (\bibinfo{year}{2024}), \bibinfo{pages}{2024--07}.
\newblock


\bibitem[Hughes et~al\mbox{.}(2023)]%
        {hughes2023bhattacharya_lab}
\bibfield{author}{\bibinfo{person}{Nathaniel Hughes}, \bibinfo{person}{Kevan Baker}, \bibinfo{person}{Aditya Singh}, \bibinfo{person}{Aryavardhan Singh}, \bibinfo{person}{Tharalillah Dauda}, {and} \bibinfo{person}{Sutanu Bhattacharya}.} \bibinfo{year}{2023}\natexlab{}.
\newblock \showarticletitle{Bhattacharya\_Lab at SemEval-2023 task 12: A transformer-based language model for sentiment classification for low resource African languages: Nigerian Pidgin and Yoruba}. In \bibinfo{booktitle}{\emph{Proceedings of the 17th international workshop on semantic evaluation (SemEval-2023)}}. \bibinfo{pages}{1502--1507}.
\newblock


\bibitem[Ibrahim and Abdulmumin(2022)]%
        {ibrahimnecat}
\bibfield{author}{\bibinfo{person}{Rabiu~Abdullahi Ibrahim} {and} \bibinfo{person}{Idris Abdulmumin}.} \bibinfo{year}{2022}\natexlab{}.
\newblock \showarticletitle{NECAT-CLWE: A Simple But Efficient Parallel Data Generation Approach for Unsupervised and Semi-Supervised Neural Machine Translation}. In \bibinfo{booktitle}{\emph{3rd Workshop on African Natural Language Processing}}. \bibinfo{publisher}{AfricaNLP Workshop at ICLR}.
\newblock


\bibitem[Ibrahim et~al\mbox{.}(2024)]%
        {ibrahim2024deep}
\bibfield{author}{\bibinfo{person}{Umar Ibrahim}, \bibinfo{person}{Abubakar~Yakubu Zandam}, \bibinfo{person}{Fatima~Muhammad Adam}, {and} \bibinfo{person}{Aminu Musa}.} \bibinfo{year}{2024}\natexlab{}.
\newblock \showarticletitle{A Deep Convolutional Neural Network-based Model for Aspect and Polarity Classification in Hausa Movie Reviews}.
\newblock \bibinfo{journal}{\emph{arXiv preprint arXiv:2405.19575}} (\bibinfo{year}{2024}).
\newblock


\bibitem[Ibrahim et~al\mbox{.}(2022a)]%
        {ibrahim2022graphic}
\bibfield{author}{\bibinfo{person}{Umar~Adam Ibrahim}, \bibinfo{person}{Moussa~Mahamat Boukar}, {and} \bibinfo{person}{Muhammed~Aliyu Suleiman}.} \bibinfo{year}{2022}\natexlab{a}.
\newblock \showarticletitle{Graphic User Interface for Hausa Text-to-Speech System}. In \bibinfo{booktitle}{\emph{2022 2nd International Conference on Computing and Machine Intelligence (ICMI)}}. IEEE, \bibinfo{pages}{1--4}.
\newblock


\bibitem[Ibrahim et~al\mbox{.}(2022b)]%
        {ibrahim2022framework}
\bibfield{author}{\bibinfo{person}{Umar~Adam Ibrahim}, \bibinfo{person}{Moussa~Boukar Mahatma}, {and} \bibinfo{person}{Muhammed~Aliyu Suleiman}.} \bibinfo{year}{2022}\natexlab{b}.
\newblock \showarticletitle{Framework for Hausa Speech Recognition}. In \bibinfo{booktitle}{\emph{2022 5th Information Technology for Education and Development (ITED)}}. IEEE, \bibinfo{pages}{1--4}.
\newblock


\bibitem[Ibrahim and Umar(2021)]%
        {ibrahim2021convolutional}
\bibfield{author}{\bibinfo{person}{Umar~Adam Ibrahim} {and} \bibinfo{person}{Rukkaya Umar}.} \bibinfo{year}{2021}\natexlab{}.
\newblock \showarticletitle{Convolutional Neural Network for Nigerian Traditional Male Attire Classification}. In \bibinfo{booktitle}{\emph{2021 1st International Conference on Multidisciplinary Engineering and Applied Science (ICMEAS)}}. IEEE, \bibinfo{pages}{1--5}.
\newblock


\bibitem[Idris et~al\mbox{.}(2022)]%
        {scholar2022development}
\bibfield{author}{\bibinfo{person}{Ya’u~Idris Idris}, \bibinfo{person}{Yusuf Suberu}, \bibinfo{person}{Ifeyinwa Madu}, \bibinfo{person}{Agabus Aminu}, \bibinfo{person}{Yahaya Amatullah~Aliyu}, \bibinfo{person}{Garba Lawan}, \bibinfo{person}{Adamu Yamusa~Idris}, \bibinfo{person}{Dantata Sunusi~Abdulhamid}, {and} \bibinfo{person}{John Okonkwo~Ogochukwo}.} \bibinfo{year}{2022}\natexlab{}.
\newblock \showarticletitle{Development of Multilingual Dictionary for English and the three popular Nigerian Languages (Hausa, Yoruba, and Igbo)}.
\newblock \bibinfo{journal}{\emph{Scholar, African}} (\bibinfo{year}{2022}).
\newblock


\bibitem[Iheanetu and Oha(2017)]%
        {iheanetu2017some}
\bibfield{author}{\bibinfo{person}{OU Iheanetu} {and} \bibinfo{person}{O Oha}.} \bibinfo{year}{2017}\natexlab{}.
\newblock \showarticletitle{Some salient issues in the unsupervised learning of Igbo morphology}. In \bibinfo{booktitle}{\emph{Proceedings of the World Congress on Engineering and Computer Science}}, Vol.~\bibinfo{volume}{1}.
\newblock


\bibitem[Iheanetu and Oha(2019)]%
        {iheanetu2019addressing}
\bibfield{author}{\bibinfo{person}{Olamma~U Iheanetu} {and} \bibinfo{person}{Obododimma Oha}.} \bibinfo{year}{2019}\natexlab{}.
\newblock \showarticletitle{Addressing the Challenges of Igbo Computational Morphological Studies Using Frequent Pattern-Based Induction}. In \bibinfo{booktitle}{\emph{Transactions on Engineering Technologies: World Congress on Engineering and Computer Science 2017}}. Springer, \bibinfo{pages}{143--156}.
\newblock


\bibitem[Imam et~al\mbox{.}(2022)]%
        {imam2022first}
\bibfield{author}{\bibinfo{person}{Sukairaj~Hafiz Imam}, \bibinfo{person}{Abubakar~Ahmad Musa}, {and} \bibinfo{person}{Ankur Choudhary}.} \bibinfo{year}{2022}\natexlab{}.
\newblock \showarticletitle{The First Corpus for Detecting Fake News in Hausa Language}.
\newblock In \bibinfo{booktitle}{\emph{Emerging Technologies for Computing, Communication and Smart Cities: Proceedings of ETCCS 2021}}. \bibinfo{publisher}{Springer}, \bibinfo{pages}{563--576}.
\newblock


\bibitem[Inuwa-Dutse(2021)]%
        {inuwa2021first}
\bibfield{author}{\bibinfo{person}{Isa Inuwa-Dutse}.} \bibinfo{year}{2021}\natexlab{}.
\newblock \showarticletitle{The first large scale collection of diverse Hausa language datasets}.
\newblock \bibinfo{journal}{\emph{arXiv preprint arXiv:2102.06991}} (\bibinfo{year}{2021}).
\newblock


\bibitem[Issaka et~al\mbox{.}(2026)]%
        {issaka2026african}
\bibfield{author}{\bibinfo{person}{Sheriff Issaka}, \bibinfo{person}{Keyi Wang}, \bibinfo{person}{Yinka Ajibola}, \bibinfo{person}{Oluwatumininu Samuel-Ipaye}, \bibinfo{person}{Zhaoyi Zhang}, \bibinfo{person}{Nicte~Aguillon Jimenez}, \bibinfo{person}{Evans~Kofi Agyei}, \bibinfo{person}{Abraham Lin}, \bibinfo{person}{Rohan Ramachandran}, \bibinfo{person}{Sadick~Abdul Mumin}, {et~al\mbox{.}}} \bibinfo{year}{2026}\natexlab{}.
\newblock \showarticletitle{The African Languages Lab: A Collaborative Approach to Advancing Low-Resource African NLP}. In \bibinfo{booktitle}{\emph{Proceedings of the 64th Annual Meeting of the Association for Computational Linguistics (Volume 1: Long Papers)}}. \bibinfo{pages}{42460--42477}.
\newblock


\bibitem[Iyanda(2015)]%
        {iyanda2015statistical}
\bibfield{author}{\bibinfo{person}{AR Iyanda}.} \bibinfo{year}{2015}\natexlab{}.
\newblock \showarticletitle{Statistical Text Analysis for Yor{\`u}b{\'a} Speech Generation Using Zipf’s Law}.
\newblock \bibinfo{journal}{\emph{Ife Journal of Technology}} \bibinfo{volume}{23}, \bibinfo{number}{2} (\bibinfo{year}{2015}), \bibinfo{pages}{40--44}.
\newblock


\bibitem[Jiang et~al\mbox{.}(2023)]%
        {jiang2023low}
\bibfield{author}{\bibinfo{person}{Zhiying Jiang}, \bibinfo{person}{Matthew Yang}, \bibinfo{person}{Mikhail Tsirlin}, \bibinfo{person}{Raphael Tang}, \bibinfo{person}{Yiqin Dai}, {and} \bibinfo{person}{Jimmy Lin}.} \bibinfo{year}{2023}\natexlab{}.
\newblock \showarticletitle{“Low-resource” text classification: A parameter-free classification method with compressors}. In \bibinfo{booktitle}{\emph{Findings of the Association for Computational Linguistics: ACL 2023}}. \bibinfo{pages}{6810--6828}.
\newblock


\bibitem[Joshi et~al\mbox{.}(2020)]%
        {joshi2020state}
\bibfield{author}{\bibinfo{person}{Pratik Joshi}, \bibinfo{person}{Sebastin Santy}, \bibinfo{person}{Amar Budhiraja}, \bibinfo{person}{Kalika Bali}, {and} \bibinfo{person}{Monojit Choudhury}.} \bibinfo{year}{2020}\natexlab{}.
\newblock \showarticletitle{The state and fate of linguistic diversity and inclusion in the NLP world}.
\newblock \bibinfo{journal}{\emph{arXiv preprint arXiv:2004.09095}} (\bibinfo{year}{2020}).
\newblock


\bibitem[Kaur et~al\mbox{.}(2021)]%
        {kaur2021automatic}
\bibfield{author}{\bibinfo{person}{Jaspreet Kaur}, \bibinfo{person}{Amitoj Singh}, {and} \bibinfo{person}{Virender Kadyan}.} \bibinfo{year}{2021}\natexlab{}.
\newblock \showarticletitle{Automatic speech recognition system for tonal languages: State-of-the-art survey}.
\newblock \bibinfo{journal}{\emph{Archives of Computational Methods in Engineering}}  \bibinfo{volume}{28} (\bibinfo{year}{2021}), \bibinfo{pages}{1039--1068}.
\newblock


\bibitem[Kolak and Resnik(2005)]%
        {kolak2005ocr}
\bibfield{author}{\bibinfo{person}{Okan Kolak} {and} \bibinfo{person}{Philip Resnik}.} \bibinfo{year}{2005}\natexlab{}.
\newblock \showarticletitle{OCR post-processing for low density languages}. In \bibinfo{booktitle}{\emph{Proceedings of Human Language Technology Conference and Conference on Empirical Methods in Natural Language Processing}}. \bibinfo{pages}{867--874}.
\newblock


\bibitem[Krasadakis et~al\mbox{.}(2024)]%
        {krasadakis2024survey}
\bibfield{author}{\bibinfo{person}{Panteleimon Krasadakis}, \bibinfo{person}{Evangelos Sakkopoulos}, {and} \bibinfo{person}{Vassilios~S Verykios}.} \bibinfo{year}{2024}\natexlab{}.
\newblock \showarticletitle{A survey on challenges and advances in natural language processing with a focus on legal informatics and low-resource languages}.
\newblock \bibinfo{journal}{\emph{Electronics}} \bibinfo{volume}{13}, \bibinfo{number}{3} (\bibinfo{year}{2024}), \bibinfo{pages}{648}.
\newblock


\bibitem[Kumolalo et~al\mbox{.}(2010)]%
        {kumolalo2010development}
\bibfield{author}{\bibinfo{person}{FO Kumolalo}, \bibinfo{person}{ER Adagunodo}, {and} \bibinfo{person}{OA Odejobi}.} \bibinfo{year}{2010}\natexlab{}.
\newblock \showarticletitle{Development of a syllabicator for Yor{\`u}b{\'a} language}.
\newblock \bibinfo{journal}{\emph{Proceedings of OAU TekConf}} (\bibinfo{year}{2010}), \bibinfo{pages}{47--51}.
\newblock


\bibitem[Lawal et~al\mbox{.}(2024)]%
        {lawalcontextual}
\bibfield{author}{\bibinfo{person}{Olanrewaju~Israel Lawal}, \bibinfo{person}{Olubayo Adekanmbi}, {and} \bibinfo{person}{Anthony Soronnadi}.} \bibinfo{year}{2024}\natexlab{}.
\newblock \showarticletitle{Contextual Evaluation of LLM’s Performance on Primary Education Science Learning Contents in the Yoruba Language}. In \bibinfo{booktitle}{\emph{5th Workshop on African Natural Language Processing}}.
\newblock


\bibitem[Liu and Yin(2023)]%
        {liu2023graphmax}
\bibfield{author}{\bibinfo{person}{Bin Liu} {and} \bibinfo{person}{Guosheng Yin}.} \bibinfo{year}{2023}\natexlab{}.
\newblock \showarticletitle{Graphmax for Text Generation}.
\newblock \bibinfo{journal}{\emph{Journal of Artificial Intelligence Research}}  \bibinfo{volume}{78} (\bibinfo{year}{2023}), \bibinfo{pages}{823--848}.
\newblock


\bibitem[Lu et~al\mbox{.}(2016)]%
        {lu2016multi}
\bibfield{author}{\bibinfo{person}{Di Lu}, \bibinfo{person}{Xiaoman Pan}, \bibinfo{person}{Nima Pourdamghani}, \bibinfo{person}{Shih-Fu Chang}, \bibinfo{person}{Heng Ji}, {and} \bibinfo{person}{Kevin Knight}.} \bibinfo{year}{2016}\natexlab{}.
\newblock \showarticletitle{A multi-media approach to cross-lingual entity knowledge transfer}. In \bibinfo{booktitle}{\emph{Proceedings of the 54th Annual Meeting of the Association for Computational Linguistics (Volume 1: Long Papers)}}. \bibinfo{pages}{54--65}.
\newblock


\bibitem[Luka et~al\mbox{.}(2012)]%
        {luka2012neural}
\bibfield{author}{\bibinfo{person}{Matthew~K Luka}, \bibinfo{person}{F Ibikunle}, {and} \bibinfo{person}{O Gregory}.} \bibinfo{year}{2012}\natexlab{}.
\newblock \showarticletitle{Neural network based Hausa language speech recognition}.
\newblock \bibinfo{journal}{\emph{(IJARAI) International Journal of Advanced Research in Artificial Intelligence}} \bibinfo{volume}{1}, \bibinfo{number}{2} (\bibinfo{year}{2012}), \bibinfo{pages}{39--44}.
\newblock


\bibitem[Mabokela et~al\mbox{.}(2022)]%
        {mabokela2022multilingual}
\bibfield{author}{\bibinfo{person}{Koena~Ronny Mabokela}, \bibinfo{person}{Turgay Celik}, {and} \bibinfo{person}{Mpho Raborife}.} \bibinfo{year}{2022}\natexlab{}.
\newblock \showarticletitle{Multilingual sentiment analysis for under-resourced languages: a systematic review of the landscape}.
\newblock \bibinfo{journal}{\emph{IEEE Access}}  \bibinfo{volume}{11} (\bibinfo{year}{2022}), \bibinfo{pages}{15996--16020}.
\newblock


\bibitem[Mabrouk et~al\mbox{.}(2021)]%
        {mabrouk2021multilingual}
\bibfield{author}{\bibinfo{person}{Aymen Ben~Elhaj Mabrouk}, \bibinfo{person}{Moez Ben~Haj Hmida}, \bibinfo{person}{Chayma Fourati}, \bibinfo{person}{Hatem Haddad}, {and} \bibinfo{person}{Abir Messaoudi}.} \bibinfo{year}{2021}\natexlab{}.
\newblock \showarticletitle{A multilingual african embedding for FAQ chatbots}.
\newblock \bibinfo{journal}{\emph{arXiv preprint arXiv:2103.09185}} (\bibinfo{year}{2021}).
\newblock


\bibitem[Maddu and Sanapala(2024)]%
        {maddu2024survey}
\bibfield{author}{\bibinfo{person}{Sandeep Maddu} {and} \bibinfo{person}{Viziananda~Row Sanapala}.} \bibinfo{year}{2024}\natexlab{}.
\newblock \showarticletitle{A survey on NLP tasks, resources and techniques for low-resource Telugu-English code-mixed text}.
\newblock \bibinfo{journal}{\emph{ACM Transactions on Asian and Low-Resource Language Information Processing}} (\bibinfo{year}{2024}).
\newblock


\bibitem[Magueresse et~al\mbox{.}(2020)]%
        {magueresse2020low}
\bibfield{author}{\bibinfo{person}{Alexandre Magueresse}, \bibinfo{person}{Vincent Carles}, {and} \bibinfo{person}{Evan Heetderks}.} \bibinfo{year}{2020}\natexlab{}.
\newblock \showarticletitle{Low-resource languages: A review of past work and future challenges}.
\newblock \bibinfo{journal}{\emph{arXiv preprint arXiv:2006.07264}} (\bibinfo{year}{2020}).
\newblock


\bibitem[Mahata et~al\mbox{.}(2020)]%
        {mahata2020performance}
\bibfield{author}{\bibinfo{person}{Sainik~Kumar Mahata}, \bibinfo{person}{Subhabrata Dutta}, \bibinfo{person}{Dipankar Das}, {and} \bibinfo{person}{Sivaji Bandyopadhyay}.} \bibinfo{year}{2020}\natexlab{}.
\newblock \showarticletitle{Performance Gain in Low Resource MT with Transfer Learning: An Analysis concerning Language Families}. In \bibinfo{booktitle}{\emph{Proceedings of the 12th Annual Meeting of the Forum for Information Retrieval Evaluation}}. \bibinfo{pages}{58--61}.
\newblock


\bibitem[Mahmud et~al\mbox{.}(2023)]%
        {mahmud2023cyberbullying}
\bibfield{author}{\bibinfo{person}{Tanjim Mahmud}, \bibinfo{person}{Michal Ptaszynski}, \bibinfo{person}{Juuso Eronen}, {and} \bibinfo{person}{Fumito Masui}.} \bibinfo{year}{2023}\natexlab{}.
\newblock \showarticletitle{Cyberbullying detection for low-resource languages and dialects: Review of the state of the art}.
\newblock \bibinfo{journal}{\emph{Information Processing \& Management}} \bibinfo{volume}{60}, \bibinfo{number}{5} (\bibinfo{year}{2023}), \bibinfo{pages}{103454}.
\newblock


\bibitem[Maitama et~al\mbox{.}(2014)]%
        {maitama2014text}
\bibfield{author}{\bibinfo{person}{Jaafar~Zubairu Maitama}, \bibinfo{person}{Usman Haruna}, \bibinfo{person}{Abdullahi~Ya'u Gambo}, \bibinfo{person}{Bimba~Andrew Thomas}, \bibinfo{person}{Norisma~Binti Idris}, \bibinfo{person}{Abdulsalam~Ya'u Gital}, {and} \bibinfo{person}{Adamu~I Abubakar}.} \bibinfo{year}{2014}\natexlab{}.
\newblock \showarticletitle{Text normalization algorithm for Facebook chats in Hausa language}. In \bibinfo{booktitle}{\emph{The 5th International Conference on Information and Communication Technology for The Muslim World (ICT4M)}}. IEEE, \bibinfo{pages}{1--4}.
\newblock


\bibitem[Maryann et~al\mbox{.}(2022)]%
        {maryann2022enhanced}
\bibfield{author}{\bibinfo{person}{Orji~Ifeoma Maryann}, \bibinfo{person}{Sylvanus~Okwudili Anigbogu}, \bibinfo{person}{Ekwealor~Oluchukwu Uzoamaka}, {and} \bibinfo{person}{Chidi~Ukamaka Betrand}.} \bibinfo{year}{2022}\natexlab{}.
\newblock \showarticletitle{Enhanced Machine Learning Algorithm for Translation of English to Igbo Language}.
\newblock \bibinfo{journal}{\emph{Machine Learning Research}} \bibinfo{volume}{7}, \bibinfo{number}{1} (\bibinfo{year}{2022}), \bibinfo{pages}{8--14}.
\newblock


\bibitem[Maryann et~al\mbox{.}(2021)]%
        {maryann2021machine}
\bibfield{author}{\bibinfo{person}{Orji~Ifeoma Maryann}, \bibinfo{person}{Sylvanus~Okwudili Anigbogu}, \bibinfo{person}{Ekwelaro~Oluchukwu Uzoamaka}, {and} \bibinfo{person}{Asogwa~Doris Chinedu}.} \bibinfo{year}{2021}\natexlab{}.
\newblock \showarticletitle{Machine Learning Translation of English into Igbo Language: A Review}.
\newblock \bibinfo{journal}{\emph{International Journal of Intelligent Information Systems}} \bibinfo{volume}{10}, \bibinfo{number}{5} (\bibinfo{year}{2021}), \bibinfo{pages}{104}.
\newblock


\bibitem[Masethe et~al\mbox{.}(2024)]%
        {masethe2024word}
\bibfield{author}{\bibinfo{person}{Hlaudi~Daniel Masethe}, \bibinfo{person}{Mosima~Anna Masethe}, \bibinfo{person}{Sunday~Olusegun Ojo}, \bibinfo{person}{Fausto Giunchiglia}, {and} \bibinfo{person}{Pius~Adewale Owolawi}.} \bibinfo{year}{2024}\natexlab{}.
\newblock \showarticletitle{Word Sense Disambiguation for Morphologically Rich Low-Resourced Languages: A Systematic Literature Review and Meta-Analysis.}
\newblock \bibinfo{journal}{\emph{Information (2078-2489)}} \bibinfo{volume}{15}, \bibinfo{number}{9} (\bibinfo{year}{2024}).
\newblock


\bibitem[Mbonu et~al\mbox{.}(2022)]%
        {mbonu2022igbosum1500}
\bibfield{author}{\bibinfo{person}{Chinedu~Emmanuel Mbonu}, \bibinfo{person}{Chiamaka~Ijeoma Chukwuneke}, \bibinfo{person}{Roseline~Uzoamaka Paul}, \bibinfo{person}{Ignatius Ezeani}, {and} \bibinfo{person}{Ikechukwu Onyenwe}.} \bibinfo{year}{2022}\natexlab{}.
\newblock \showarticletitle{Igbosum1500-introducing the igbo text summarization dataset}. In \bibinfo{booktitle}{\emph{3rd Workshop on African Natural Language Processing}}.
\newblock


\bibitem[Mehari et~al\mbox{.}(2024)]%
        {mehari2024semi}
\bibfield{author}{\bibinfo{person}{Yohannes~Hailemariam Mehari}, \bibinfo{person}{Steven Lynden}, \bibinfo{person}{Toshiyuki Amagasa}, {and} \bibinfo{person}{Akiyoshi Matono}.} \bibinfo{year}{2024}\natexlab{}.
\newblock \showarticletitle{Semi-supervised Named Entity Recognition for Low-Resource Languages Using Dual PLMs}. In \bibinfo{booktitle}{\emph{International Conference on Applications of Natural Language to Information Systems}}. Springer, \bibinfo{pages}{166--180}.
\newblock


\bibitem[Mijinguini(2003)]%
        {mijinguini2003ƙaramin}
\bibfield{author}{\bibinfo{person}{Abdou Mijinguini}.} \bibinfo{year}{2003}\natexlab{}.
\newblock \bibinfo{title}{Ƙaramin Ƙamus: Dictionnaire El{\'e}mentaire Hausa-Fran{\c{c}}ais}.
\newblock


\bibitem[Mohammed and Prasad(2023)]%
        {mohammed2023building}
\bibfield{author}{\bibinfo{person}{Idi Mohammed} {and} \bibinfo{person}{Rajesh Prasad}.} \bibinfo{year}{2023}\natexlab{}.
\newblock \showarticletitle{Building lexicon-based sentiment analysis model for low-resource languages}.
\newblock \bibinfo{journal}{\emph{MethodsX}}  \bibinfo{volume}{11} (\bibinfo{year}{2023}), \bibinfo{pages}{102460}.
\newblock


\bibitem[Mohammed and Prasad(2024)]%
        {mohammed2024lexicon}
\bibfield{author}{\bibinfo{person}{Idi Mohammed} {and} \bibinfo{person}{Rajesh Prasad}.} \bibinfo{year}{2024}\natexlab{}.
\newblock \showarticletitle{Lexicon dataset for the Hausa language}.
\newblock \bibinfo{journal}{\emph{Data in Brief}}  \bibinfo{volume}{53} (\bibinfo{year}{2024}), \bibinfo{pages}{110124}.
\newblock


\bibitem[Muhammad et~al\mbox{.}(2025c)]%
        {muhammad2025dataset}
\bibfield{author}{\bibinfo{person}{Fatima~Adam Muhammad}, \bibinfo{person}{Shamsuddeen~Muhammad Hassan}, {and} \bibinfo{person}{Isa Inuwa-Dutse}.} \bibinfo{year}{2025}\natexlab{c}.
\newblock \showarticletitle{Dataset Creation and Baseline Models for Sexism Detection in Hausa}.
\newblock \bibinfo{journal}{\emph{arXiv preprint arXiv:2510.27038}} (\bibinfo{year}{2025}).
\newblock


\bibitem[Muhammad et~al\mbox{.}(2025a)]%
        {muhammad2025afrihate}
\bibfield{author}{\bibinfo{person}{Shamsuddeen~Hassan Muhammad}, \bibinfo{person}{Idris Abdulmumin}, \bibinfo{person}{Abinew~Ali Ayele}, \bibinfo{person}{David~Ifeoluwa Adelani}, \bibinfo{person}{Ibrahim~Said Ahmad}, \bibinfo{person}{Saminu~Mohammad Aliyu}, \bibinfo{person}{Nelson~Odhiambo Onyango}, \bibinfo{person}{Lilian~DA Wanzare}, \bibinfo{person}{Samuel Rutunda}, \bibinfo{person}{Lukman~Jibril Aliyu}, {et~al\mbox{.}}} \bibinfo{year}{2025}\natexlab{a}.
\newblock \showarticletitle{AfriHate: A Multilingual Collection of Hate Speech and Abusive Language Datasets for African Languages}.
\newblock \bibinfo{journal}{\emph{arXiv preprint arXiv:2501.08284}} (\bibinfo{year}{2025}).
\newblock


\bibitem[Muhammad et~al\mbox{.}(2022)]%
        {muhammad2022naijasenti}
\bibfield{author}{\bibinfo{person}{Shamsuddeen~Hassan Muhammad}, \bibinfo{person}{David~Ifeoluwa Adelani}, \bibinfo{person}{Sebastian Ruder}, \bibinfo{person}{Ibrahim~Said Ahmad}, \bibinfo{person}{Idris Abdulmumin}, \bibinfo{person}{Bello~Shehu Bello}, \bibinfo{person}{Monojit Choudhury}, \bibinfo{person}{Chris~Chinenye Emezue}, \bibinfo{person}{Saheed~Salahudeen Abdullahi}, \bibinfo{person}{Anuoluwapo Aremu}, {et~al\mbox{.}}} \bibinfo{year}{2022}\natexlab{}.
\newblock \showarticletitle{Naijasenti: A nigerian twitter sentiment corpus for multilingual sentiment analysis}.
\newblock \bibinfo{journal}{\emph{arXiv preprint arXiv:2201.08277}} (\bibinfo{year}{2022}).
\newblock


\bibitem[Muhammad et~al\mbox{.}(2025b)]%
        {muhammad2025hausanlp}
\bibfield{author}{\bibinfo{person}{Shamsuddeen~Hassan Muhammad}, \bibinfo{person}{Ibrahim~Said Ahmad}, \bibinfo{person}{Idris Abdulmumin}, \bibinfo{person}{Falalu~Ibrahim Lawan}, \bibinfo{person}{Sukairaj~Hafiz Imam}, \bibinfo{person}{Yusuf Aliyu}, \bibinfo{person}{Sani~Abdullahi Sani}, \bibinfo{person}{Ali~Usman Umar}, \bibinfo{person}{Tajuddeen Gwadabe}, \bibinfo{person}{Kenneth Church}, {et~al\mbox{.}}} \bibinfo{year}{2025}\natexlab{b}.
\newblock \showarticletitle{Hausanlp: Current status, challenges and future directions for hausa natural language processing}. In \bibinfo{booktitle}{\emph{Proceedings of the Sixth Workshop on African Natural Language Processing (AfricaNLP 2025)}}. \bibinfo{pages}{176--191}.
\newblock


\bibitem[Muller et~al\mbox{.}(2020)]%
        {muller2020being}
\bibfield{author}{\bibinfo{person}{Benjamin Muller}, \bibinfo{person}{Antonis Anastasopoulos}, \bibinfo{person}{Beno{\^\i}t Sagot}, {and} \bibinfo{person}{Djam{\'e} Seddah}.} \bibinfo{year}{2020}\natexlab{}.
\newblock \showarticletitle{When being unseen from mBERT is just the beginning: Handling new languages with multilingual language models}.
\newblock \bibinfo{journal}{\emph{arXiv preprint arXiv:2010.12858}} (\bibinfo{year}{2020}).
\newblock


\bibitem[Musa et~al\mbox{.}(2022)]%
        {musa2022improved}
\bibfield{author}{\bibinfo{person}{Sirajo Musa}, \bibinfo{person}{GN Obunadike}, {and} \bibinfo{person}{Muhammad~Muntasir Yakubu}.} \bibinfo{year}{2022}\natexlab{}.
\newblock \showarticletitle{AN IMPROVED HAUSA WORD STEMMING ALGORITHM}.
\newblock \bibinfo{journal}{\emph{FUDMA JOURNAL OF SCIENCES}} \bibinfo{volume}{6}, \bibinfo{number}{1} (\bibinfo{year}{2022}), \bibinfo{pages}{291--295}.
\newblock


\bibitem[Mustapha et~al\mbox{.}(2023)]%
        {mustapha2023automated}
\bibfield{author}{\bibinfo{person}{Rabi Mustapha}, \bibinfo{person}{Falalu~Ibrahim Lawal}, {and} \bibinfo{person}{Muhammad~Aminu Ahmad}.} \bibinfo{year}{2023}\natexlab{}.
\newblock \showarticletitle{Automated conversion of numeral to words in Hausa language}.
\newblock \bibinfo{journal}{\emph{Gadau Journal of Pure and Allied Sciences}} \bibinfo{volume}{2}, \bibinfo{number}{2} (\bibinfo{year}{2023}), \bibinfo{pages}{140--145}.
\newblock


\bibitem[NBS(2024)]%
        {internetusengn2024}
\bibfield{author}{\bibinfo{person}{National Bureau of~Statistics NBS}.} \bibinfo{year}{2024}\natexlab{}.
\newblock \showarticletitle{Telecoms Data: Active Voice and Internet, Porting and Tariff Information}.
\newblock  (\bibinfo{year}{2024}).
\newblock


\bibitem[Nganga and Achebe(2020)]%
        {nganga2020spoken}
\bibfield{author}{\bibinfo{person}{Wanjiku Nganga} {and} \bibinfo{person}{Ikechukwu Achebe}.} \bibinfo{year}{2020}\natexlab{}.
\newblock \showarticletitle{Spoken word corpus and dictionary definition for an African language}.
\newblock \bibinfo{journal}{\emph{Journal of Data Mining \& Digital Humanities}} \bibinfo{number}{Digital humanities in languages} (\bibinfo{year}{2020}).
\newblock


\bibitem[Ngueajio and Washington(2022)]%
        {ngueajio2022hey}
\bibfield{author}{\bibinfo{person}{Mikel~K Ngueajio} {and} \bibinfo{person}{Gloria Washington}.} \bibinfo{year}{2022}\natexlab{}.
\newblock \showarticletitle{Hey ASR system! Why aren’t you more inclusive? Automatic speech recognition systems’ bias and proposed bias mitigation techniques. A literature review}. In \bibinfo{booktitle}{\emph{International Conference on Human-Computer Interaction}}. Springer, \bibinfo{pages}{421--440}.
\newblock


\bibitem[Nigatu et~al\mbox{.}(2024)]%
        {nigatu2024zeno}
\bibfield{author}{\bibinfo{person}{Hellina~Hailu Nigatu}, \bibinfo{person}{Atnafu~Lambebo Tonja}, \bibinfo{person}{Benjamin Rosman}, \bibinfo{person}{Thamar Solorio}, {and} \bibinfo{person}{Monojit Choudhury}.} \bibinfo{year}{2024}\natexlab{}.
\newblock \showarticletitle{The Zeno's Paradox ofLow-Resource'Languages}.
\newblock \bibinfo{journal}{\emph{arXiv preprint arXiv:2410.20817}} (\bibinfo{year}{2024}).
\newblock


\bibitem[Nkechi et~al\mbox{.}(2021)]%
        {ifeanyin}
\bibfield{author}{\bibinfo{person}{Ifeanyi-Reuben Nkechi}, \bibinfo{person}{Odikwa Ndubuisi}, {and} \bibinfo{person}{Ugwu Chidiebere}.} \bibinfo{year}{2021}\natexlab{}.
\newblock \showarticletitle{N-Gram and K-Nearest Neighbour Based Igbo Text Classification Model}.
\newblock \bibinfo{journal}{\emph{International Journal of Innovative Science and Research Technology}} (\bibinfo{year}{2021}).
\newblock


\bibitem[Nkechi et~al\mbox{.}(2020)]%
        {ifeanyi2020comparative}
\bibfield{author}{\bibinfo{person}{Ifeanyi-Reuben Nkechi}, \bibinfo{person}{Chidiebere Ugwu}, {et~al\mbox{.}}} \bibinfo{year}{2020}\natexlab{}.
\newblock \showarticletitle{Comparative Analysis of N-gram Text Representation on Igbo Text Document Similarity}.
\newblock \bibinfo{journal}{\emph{arXiv preprint arXiv:2004.00375}} (\bibinfo{year}{2020}).
\newblock


\bibitem[Nkechi and Usip(2022)]%
        {ifeanyi2022semantic}
\bibfield{author}{\bibinfo{person}{Ifeanyi-Reuben Nkechi} {and} \bibinfo{person}{Patience~Usoro Usip}.} \bibinfo{year}{2022}\natexlab{}.
\newblock \showarticletitle{Semantic Representation of Igbo Text Using Knowledge Graph.}. In \bibinfo{booktitle}{\emph{KGSWC Workshops}}. \bibinfo{pages}{37--51}.
\newblock


\bibitem[Nowakowski and Dwojak(2021)]%
        {nowakowski2021adam}
\bibfield{author}{\bibinfo{person}{Artur Nowakowski} {and} \bibinfo{person}{Tomasz Dwojak}.} \bibinfo{year}{2021}\natexlab{}.
\newblock \showarticletitle{Adam mickiewicz university’s english-hausa submissions to the wmt 2021 news translation task}. In \bibinfo{booktitle}{\emph{Proceedings of the Sixth Conference on Machine Translation}}. \bibinfo{pages}{167--171}.
\newblock


\bibitem[Nwankwegu(2021)]%
        {nwankwegu2021leveraging}
\bibfield{author}{\bibinfo{person}{Jeremiah~Anene Nwankwegu}.} \bibinfo{year}{2021}\natexlab{}.
\newblock \showarticletitle{LEVERAGING DIGITAL TECHNOLOGY FOR IGBO LANGUAGE DEVELOPMENT}.
\newblock \bibinfo{journal}{\emph{IDEAL INTERNATIONAL JOURNAL}} \bibinfo{volume}{14}, \bibinfo{number}{2} (\bibinfo{year}{2021}).
\newblock


\bibitem[Obiajulu et~al\mbox{.}(2019)]%
        {obiajuludigital}
\bibfield{author}{\bibinfo{person}{Emejulu~Augustine Obiajulu}, \bibinfo{person}{Okpala~Izunna Udebuana}, {and} \bibinfo{person}{Nwakanma~Ifeanyi Cosmas}.} \bibinfo{year}{2019}\natexlab{}.
\newblock \showarticletitle{Digital Language Mining Platform for Nigerian Languages (DLMP)}.
\newblock \bibinfo{journal}{\emph{International Journal on Data Science and Technology}} (\bibinfo{year}{2019}).
\newblock


\bibitem[Ochu(2019)]%
        {ochu2019corpus}
\bibfield{author}{\bibinfo{person}{Michael~Chima Ochu}.} \bibinfo{year}{2019}\natexlab{}.
\newblock \showarticletitle{A CORPUS-DRIVEN STUDY OF MULTIWORD EXPRESSIONS IN IGBO.}
\newblock \bibinfo{journal}{\emph{Journal of West African Languages}} \bibinfo{volume}{46}, \bibinfo{number}{2} (\bibinfo{year}{2019}).
\newblock


\bibitem[Od{\'e}lọb{\'\i}(2008)]%
        {odelobi2008recognition}
\bibfield{author}{\bibinfo{person}{Ọd{\'e}t{\'u}nj{\'\i}~{\`A}j{\`a}d{\'\i} Od{\'e}lọb{\'\i}}.} \bibinfo{year}{2008}\natexlab{}.
\newblock \showarticletitle{Recognition of tones in Yoruba speech: experiments with artificial neural networks}.
\newblock In \bibinfo{booktitle}{\emph{Speech, Audio, Image and Biomedical Signal Processing using Neural Networks}}. \bibinfo{publisher}{Springer}, \bibinfo{pages}{23--47}.
\newblock


\bibitem[Odoje(2014)]%
        {odoje2014investigating}
\bibfield{author}{\bibinfo{person}{Clement Odoje}.} \bibinfo{year}{2014}\natexlab{}.
\newblock \showarticletitle{Investigating language in the machine translation: exploring Yor{\`u}b{\'a}-English machine translation as a case study}.
\newblock \bibinfo{journal}{\emph{Language. Text. Society}} \bibinfo{volume}{4}, \bibinfo{number}{1} (\bibinfo{year}{2014}), \bibinfo{pages}{17--26}.
\newblock


\bibitem[Odoje(2016)]%
        {odoje201612}
\bibfield{author}{\bibinfo{person}{Clement Odoje}.} \bibinfo{year}{2016}\natexlab{}.
\newblock \showarticletitle{The Peculiar Challenges of SMT to African Languages}.
\newblock \bibinfo{journal}{\emph{ICT, Globalisation and the Study of Languages and Linguistics in Africa}} (\bibinfo{year}{2016}), \bibinfo{pages}{223}.
\newblock


\bibitem[Oduwole et~al\mbox{.}(2026)]%
        {oduwole2026africaption}
\bibfield{author}{\bibinfo{person}{Mardiyyah Oduwole}, \bibinfo{person}{Prince Mireku}, \bibinfo{person}{Fatimo Adebanjo}, \bibinfo{person}{Oluwatosin Olajide}, \bibinfo{person}{Mahi~Aminu Aliyu}, {and} \bibinfo{person}{Jekaterina Novikova}.} \bibinfo{year}{2026}\natexlab{}.
\newblock \showarticletitle{AFRICAPTION: Establishing a New Paradigm for Image Captioning in African Languages}. In \bibinfo{booktitle}{\emph{Proceedings of the 7th Workshop on African Natural Language Processing (AfricaNLP 2026)}}. \bibinfo{pages}{44--55}.
\newblock


\bibitem[Ogbuju and Onyesolu(2020)]%
        {ogbuju2020development}
\bibfield{author}{\bibinfo{person}{Emeka Ogbuju} {and} \bibinfo{person}{Moses Onyesolu}.} \bibinfo{year}{2020}\natexlab{}.
\newblock \showarticletitle{Development of a general purpose sentiment lexicon for Igbo language}.
\newblock \bibinfo{journal}{\emph{arXiv preprint arXiv:2004.14176}} (\bibinfo{year}{2020}).
\newblock


\bibitem[Ogheneruemu(2022)]%
        {ogheneruemu2022development}
\bibfield{author}{\bibinfo{person}{Kingsley Lucky~Ogheneovo Ogheneruemu}.} \bibinfo{year}{2022}\natexlab{}.
\newblock \emph{\bibinfo{title}{Development of Yoruba Diacritic Restoration for Under Dot and Diacritic Mark for Yoruba Text Using Deep Learning Model}}.
\newblock \bibinfo{thesistype}{Master's\ thesis}. \bibinfo{school}{Kwara State University (Nigeria)}.
\newblock


\bibitem[Ogueji et~al\mbox{.}(2021)]%
        {ogueji2021small}
\bibfield{author}{\bibinfo{person}{Kelechi Ogueji}, \bibinfo{person}{Yuxin Zhu}, {and} \bibinfo{person}{Jimmy Lin}.} \bibinfo{year}{2021}\natexlab{}.
\newblock \showarticletitle{Small data? no problem! exploring the viability of pretrained multilingual language models for low-resourced languages}. In \bibinfo{booktitle}{\emph{Proceedings of the 1st workshop on multilingual representation learning}}. \bibinfo{pages}{116--126}.
\newblock


\bibitem[Ogundepo(2023)]%
        {ogundepo2023enabling}
\bibfield{author}{\bibinfo{person}{Odunayo Ogundepo}.} \bibinfo{year}{2023}\natexlab{}.
\newblock \emph{\bibinfo{title}{Enabling Cross-lingual Information Retrieval for African Languages}}.
\newblock \bibinfo{thesistype}{Master's\ thesis}. \bibinfo{school}{University of Waterloo}.
\newblock


\bibitem[Ogunleye et~al\mbox{.}(2023)]%
        {ogunleye2023using}
\bibfield{author}{\bibinfo{person}{Bayode Ogunleye}, \bibinfo{person}{Teresa Brunsdon}, \bibinfo{person}{Tonderai Maswera}, \bibinfo{person}{Laurence Hirsch}, {and} \bibinfo{person}{Jotham Gaudoin}.} \bibinfo{year}{2023}\natexlab{}.
\newblock \showarticletitle{Using Opinionated-Objective Terms to Improve Lexicon-Based Sentiment Analysis}. In \bibinfo{booktitle}{\emph{International conference on soft computing for problem-solving}}. Springer, \bibinfo{pages}{1--23}.
\newblock


\bibitem[Ogunrinde and Yusuf(2026)]%
        {ogunrinde2026culturally}
\bibfield{author}{\bibinfo{person}{Mutiat~A Ogunrinde} {and} \bibinfo{person}{Yusuf~B Yusuf}.} \bibinfo{year}{2026}\natexlab{}.
\newblock \showarticletitle{A culturally aware NLP approach for fake-news detection in Nigerian online media using BERT}.
\newblock \bibinfo{journal}{\emph{African Scientific Reports}} (\bibinfo{year}{2026}), \bibinfo{pages}{351--351}.
\newblock


\bibitem[Ohuoba et~al\mbox{.}(2024)]%
        {ohuoba2024quantifying}
\bibfield{author}{\bibinfo{person}{Adaeze Ohuoba}, \bibinfo{person}{Serge Sharoff}, {and} \bibinfo{person}{Callum Walker}.} \bibinfo{year}{2024}\natexlab{}.
\newblock \showarticletitle{Quantifying the Contribution of MWEs and Polysemy in Translation Errors for English--Igbo MT}. In \bibinfo{booktitle}{\emph{Proceedings of the 25th Annual Conference of the European Association for Machine Translation (Volume 1)}}. \bibinfo{pages}{537--547}.
\newblock


\bibitem[Ojo et~al\mbox{.}(2023)]%
        {ojo2023good}
\bibfield{author}{\bibinfo{person}{Jessica Ojo}, \bibinfo{person}{Kelechi Ogueji}, \bibinfo{person}{Pontus Stenetorp}, {and} \bibinfo{person}{David~Ifeoluwa Adelani}.} \bibinfo{year}{2023}\natexlab{}.
\newblock \showarticletitle{How good are Large Language Models on African Languages?}
\newblock \bibinfo{journal}{\emph{arXiv preprint arXiv:2311.07978}} (\bibinfo{year}{2023}).
\newblock


\bibitem[Okediya et~al\mbox{.}(2019)]%
        {okediya2019building}
\bibfield{author}{\bibinfo{person}{Theresa Okediya}, \bibinfo{person}{Ibukun Afolabi}, \bibinfo{person}{Olamma Iheanetu}, {and} \bibinfo{person}{Sunday Ojo}.} \bibinfo{year}{2019}\natexlab{}.
\newblock \showarticletitle{Building Ontology for Yor{\`u}b{\'a} Language}. In \bibinfo{booktitle}{\emph{Proceedings of the First International Workshop on NLP Solutions for Under Resourced Languages (NSURL 2019) co-located with ICNLSP 2019-Short Papers}}. \bibinfo{pages}{124--130}.
\newblock


\bibitem[Okoloegbo et~al\mbox{.}(2022)]%
        {okoloegbo2022multilingual}
\bibfield{author}{\bibinfo{person}{CA Okoloegbo}, \bibinfo{person}{UF Eze}, \bibinfo{person}{GA Chukwudebe}, {and} \bibinfo{person}{OC Nwokonkwo}.} \bibinfo{year}{2022}\natexlab{}.
\newblock \bibinfo{title}{Multilingual Cyberbullying Detector (CD) Application for Nigerian Pidgin and Igbo Language Corpus. In 2022 5th Information Technology for Education and Development (ITED), 1-6}.
\newblock


\bibitem[Okoloegbo et~al\mbox{.}(2024)]%
        {okoloegbomultilingual}
\bibfield{author}{\bibinfo{person}{Christiana~Amaka Okoloegbo}, \bibinfo{person}{Udoka~Felista Eze}, \bibinfo{person}{Gloria~A Chukwudebe}, {and} \bibinfo{person}{Obi Chukwuemeka}.} \bibinfo{year}{2024}\natexlab{}.
\newblock \showarticletitle{Multilingual Cyberbullying Detection System in Nigerian Languages}.
\newblock \bibinfo{journal}{\emph{International Journal of Computer Science and Information Security (IJCSIS)}} (\bibinfo{year}{2024}).
\newblock


\bibitem[Oladipo et~al\mbox{.}(2024)]%
        {oladipo2024backbones}
\bibfield{author}{\bibinfo{person}{Akintunde Oladipo}, \bibinfo{person}{Mofetoluwa Adeyemi}, {and} \bibinfo{person}{Jimmy Lin}.} \bibinfo{year}{2024}\natexlab{}.
\newblock \showarticletitle{On Backbones and Training Regimes for Dense Retrieval in African Languages}. In \bibinfo{booktitle}{\emph{Proceedings of the 47th International ACM SIGIR Conference on Research and Development in Information Retrieval}}. \bibinfo{pages}{2564--2568}.
\newblock


\bibitem[Olalekan et~al\mbox{.}(2022)]%
        {olalekan2022machine}
\bibfield{author}{\bibinfo{person}{Awoniran Olalekan}, \bibinfo{person}{Ogundiran Daniel}, \bibinfo{person}{Ozichi~N Emuoyibofarhe}, {et~al\mbox{.}}} \bibinfo{year}{2022}\natexlab{}.
\newblock \showarticletitle{MACHINE LEARNING ALGORITHMS FOR MULTILINGUAL TEXT CLASSIFICATION OF NIGERIAN LOCAL LANGUAGES.}
\newblock \bibinfo{journal}{\emph{Computer Science \& Telecommunications}} \bibinfo{volume}{62}, \bibinfo{number}{2} (\bibinfo{year}{2022}).
\newblock


\bibitem[Olamma et~al\mbox{.}(2019)]%
        {olamma2019hidden}
\bibfield{author}{\bibinfo{person}{Ihenaetu Olamma}, \bibinfo{person}{Michael Kingsley}, {and} \bibinfo{person}{Sunday Ojo}.} \bibinfo{year}{2019}\natexlab{}.
\newblock \showarticletitle{Hidden Markov-based Part-of-Speech Tagger for Igbo Resource-Scarce African Language}. In \bibinfo{booktitle}{\emph{Proceedings of the First International Workshop on NLP Solutions for Under Resourced Languages (NSURL 2019) co-located with ICNLSP 2019-Short Papers}}. \bibinfo{pages}{118--123}.
\newblock


\bibitem[Olarewaju(2024)]%
        {aremu2024utilising}
\bibfield{author}{\bibinfo{person}{Aremu~Abdulahi Olarewaju}.} \bibinfo{year}{2024}\natexlab{}.
\newblock \showarticletitle{Utilising AI-powered Chatbots for Learning Endangered Nigerian Languages and Considerations for Their Development}.
\newblock \bibinfo{journal}{\emph{Journal of Communication, Language and Culture}} \bibinfo{volume}{4}, \bibinfo{number}{2} (\bibinfo{year}{2024}), \bibinfo{pages}{41--56}.
\newblock


\bibitem[Oluwaseyi et~al\mbox{.}(2024)]%
        {oluwaseyi2024automatic}
\bibfield{author}{\bibinfo{person}{Erinfolami Oluwaseyi}, \bibinfo{person}{Oguntimilehin Abiodun}, \bibinfo{person}{Bukola Badeji-Ajisafe}, {et~al\mbox{.}}} \bibinfo{year}{2024}\natexlab{}.
\newblock \showarticletitle{Automatic Spelling Corrector for Yor{\`u}b{\'a} Language Using Edit Distance and N-Gram Language Models}. In \bibinfo{booktitle}{\emph{2024 International Conference on Science, Engineering and Business for Driving Sustainable Development Goals (SEB4SDG)}}. IEEE, \bibinfo{pages}{1--6}.
\newblock


\bibitem[Oluwatoyin and Opeyemi(2018)]%
        {oluwatoyinstochastic}
\bibfield{author}{\bibinfo{person}{Enikuomehin~A Oluwatoyin} {and} \bibinfo{person}{Adewumi~O Opeyemi}.} \bibinfo{year}{2018}\natexlab{}.
\newblock \showarticletitle{A Stochastic Collocation Algorithm Method for Processing the Yoruba Language Using the Data Context Approach Based on Text, Lexicon, and Grammar}.
\newblock \bibinfo{journal}{\emph{The Pacific Journal of Science and Technology}} (\bibinfo{year}{2018}).
\newblock


\bibitem[Omolaoye(2020)]%
        {omolaoye2020proverb}
\bibfield{author}{\bibinfo{person}{Victor~Adelakun Omolaoye}.} \bibinfo{year}{2020}\natexlab{}.
\newblock \emph{\bibinfo{title}{Proverb representation using semantic technologies: a case study of Nigerian Yoruba proverbs}}.
\newblock \bibinfo{thesistype}{Ph.\,D. Dissertation}. \bibinfo{school}{Namibia University of Science and Technology}.
\newblock


\bibitem[Omotayo et~al\mbox{.}(2024)]%
        {omotayo2024state}
\bibfield{author}{\bibinfo{person}{Abdul-Hakeem Omotayo}, \bibinfo{person}{Ashery Mbilinyi}, \bibinfo{person}{Lukman Ismaila}, \bibinfo{person}{Houcemeddine Turki}, \bibinfo{person}{Mahmoud Abdien}, \bibinfo{person}{Karim Gamal}, \bibinfo{person}{Idriss Tondji}, \bibinfo{person}{Yvan Pimi}, \bibinfo{person}{Naome~A Etori}, \bibinfo{person}{Marwa~M Matar}, {et~al\mbox{.}}} \bibinfo{year}{2024}\natexlab{}.
\newblock \showarticletitle{The State of Computer Vision Research in Africa}.
\newblock \bibinfo{journal}{\emph{Journal of Artificial Intelligence Research}}  \bibinfo{volume}{81} (\bibinfo{year}{2024}), \bibinfo{pages}{43--69}.
\newblock


\bibitem[Onuora et~al\mbox{.}(2024)]%
        {onuora2024machine}
\bibfield{author}{\bibinfo{person}{AC Onuora}, \bibinfo{person}{P Ana}, \bibinfo{person}{AO Otiko}, \bibinfo{person}{RC Aguwamba}, {and} \bibinfo{person}{NE Maidoh}.} \bibinfo{year}{2024}\natexlab{}.
\newblock \showarticletitle{Machine Learning Architecture for Combating Hate Speech in Igbo Language}. In \bibinfo{booktitle}{\emph{International conference 2024, March}}, Vol.~\bibinfo{volume}{27}. \bibinfo{pages}{29}.
\newblock


\bibitem[Onyemaechi and Ojiako(2023)]%
        {onyemaechi2023some}
\bibfield{author}{\bibinfo{person}{Chinaecherem~Marycynthia Onyemaechi} {and} \bibinfo{person}{Chinwe~Doris Ojiako}.} \bibinfo{year}{2023}\natexlab{}.
\newblock \showarticletitle{Some Linguistic Features of Chi-Prefixed Igbo Personal Names}.
\newblock \bibinfo{journal}{\emph{JOURNAL OF LINGUISTICS, LANGUAGE AND IGBO STUDIES (JoLLIS)}} \bibinfo{volume}{4}, \bibinfo{number}{1} (\bibinfo{year}{2023}).
\newblock


\bibitem[Onyenwe et~al\mbox{.}(2015)]%
        {onyenwe2015use}
\bibfield{author}{\bibinfo{person}{Ikechukwu Onyenwe}, \bibinfo{person}{Mark Hepple}, \bibinfo{person}{Chinedu Uchechukwu}, {and} \bibinfo{person}{Ignatius Ezeani}.} \bibinfo{year}{2015}\natexlab{}.
\newblock \showarticletitle{Use of transformation-based learning in annotation pipeline of igbo, an african language}. In \bibinfo{booktitle}{\emph{Proceedings of the Joint Workshop on Language Technology for Closely Related Languages, Varieties and Dialects}}. Association for Computational Linguistics, \bibinfo{pages}{24--33}.
\newblock


\bibitem[Onyenwe and Hepple(2016)]%
        {onyenwe2016predicting}
\bibfield{author}{\bibinfo{person}{Ikechukwu~E Onyenwe} {and} \bibinfo{person}{Mark Hepple}.} \bibinfo{year}{2016}\natexlab{}.
\newblock \showarticletitle{Predicting morphologically-complex unknown words in Igbo}. In \bibinfo{booktitle}{\emph{Text, Speech, and Dialogue: 19th International Conference, TSD 2016, Brno, Czech Republic, September 12-16, 2016, Proceedings 19}}. Springer, \bibinfo{pages}{206--214}.
\newblock


\bibitem[Onyenwe et~al\mbox{.}(2018)]%
        {onyenwe2018basic}
\bibfield{author}{\bibinfo{person}{Ikechukwu~E Onyenwe}, \bibinfo{person}{Mark Hepple}, \bibinfo{person}{Uchechukwu Chinedu}, {and} \bibinfo{person}{Ignatius Ezeani}.} \bibinfo{year}{2018}\natexlab{}.
\newblock \showarticletitle{A Basic Language Resource Kit Implementation for the Igbo NLP Project}.
\newblock \bibinfo{journal}{\emph{ACM Transactions on Asian and Low-Resource Language Information Processing (TALLIP)}} \bibinfo{volume}{17}, \bibinfo{number}{2} (\bibinfo{year}{2018}), \bibinfo{pages}{1--23}.
\newblock


\bibitem[Onyenwe et~al\mbox{.}(2019a)]%
        {onyenwe2019toward}
\bibfield{author}{\bibinfo{person}{Ikechukwu~E Onyenwe}, \bibinfo{person}{Mark Hepple}, \bibinfo{person}{Uchechukwu Chinedu}, {and} \bibinfo{person}{Ignatius Ezeani}.} \bibinfo{year}{2019}\natexlab{a}.
\newblock \showarticletitle{Toward an effective igbo part-of-speech tagger}.
\newblock \bibinfo{journal}{\emph{ACM Transactions on Asian and Low-Resource Language Information Processing (TALLIP)}} \bibinfo{volume}{18}, \bibinfo{number}{4} (\bibinfo{year}{2019}), \bibinfo{pages}{1--26}.
\newblock


\bibitem[Onyenwe et~al\mbox{.}(2019b)]%
        {anbootstrapping}
\bibfield{author}{\bibinfo{person}{Ikechukwu~E Onyenwe}, \bibinfo{person}{Ebele~G Onyedinma}, \bibinfo{person}{Godwin~E Aniegwu}, {and} \bibinfo{person}{Ignatius~M Ezeani}.} \bibinfo{year}{2019}\natexlab{b}.
\newblock \showarticletitle{BOOTSTRAPPING METHOD FOR DEVELOPING PART-OF-SPEECH TAGGED CORPUS IN LOW RESOURCE LANGUAGES TAGSET-AFocus ON AN AFRICAN IGBO}.
\newblock \bibinfo{journal}{\emph{International Journal on Natural Language Computing (IJNLC)}} (\bibinfo{year}{2019}).
\newblock


\bibitem[Onyenwe et~al\mbox{.}(2014)]%
        {onyenwe2014part}
\bibfield{author}{\bibinfo{person}{Ikechukwu~E Onyenwe}, \bibinfo{person}{Chinedu Uchechukwu}, {and} \bibinfo{person}{Mark Hepple}.} \bibinfo{year}{2014}\natexlab{}.
\newblock \showarticletitle{Part-of-speech tagset and corpus development for igbo, an african}. In \bibinfo{booktitle}{\emph{Proceedings of LAW VIII-The 8th Linguistic Annotation Workshop}}. Association for Computational Linguistics and Dublin City University, \bibinfo{pages}{93--98}.
\newblock


\bibitem[Orife(2018)]%
        {orife2018attentive}
\bibfield{author}{\bibinfo{person}{Iroro Orife}.} \bibinfo{year}{2018}\natexlab{}.
\newblock \showarticletitle{Attentive Sequence-to-Sequence Learning for Diacritic Restoration of Yor$\backslash$ub$\backslash$'a Language Text}.
\newblock \bibinfo{journal}{\emph{arXiv preprint arXiv:1804.00832}} (\bibinfo{year}{2018}).
\newblock


\bibitem[Orife et~al\mbox{.}(2020)]%
        {orife2020improving}
\bibfield{author}{\bibinfo{person}{Iroro Orife}, \bibinfo{person}{David~I Adelani}, \bibinfo{person}{Timi Fasubaa}, \bibinfo{person}{Victor Williamson}, \bibinfo{person}{Wuraola~Fisayo Oyewusi}, \bibinfo{person}{Olamilekan Wahab}, {and} \bibinfo{person}{Kola Tubosun}.} \bibinfo{year}{2020}\natexlab{}.
\newblock \showarticletitle{Improving Yor$\backslash$ub$\backslash$'a Diacritic Restoration}.
\newblock \bibinfo{journal}{\emph{arXiv preprint arXiv:2003.10564}} (\bibinfo{year}{2020}).
\newblock


\bibitem[Orji and Korie(2024)]%
        {orji2024igbo}
\bibfield{author}{\bibinfo{person}{Dereck-MA Orji} {and} \bibinfo{person}{Ijeoma~Jennifer Korie}.} \bibinfo{year}{2024}\natexlab{}.
\newblock \showarticletitle{The Igbo Language in an AI World: Assessing its Current Status}.
\newblock \bibinfo{journal}{\emph{OHAZURUME-Unizik Journal of Culture and Civilization}} \bibinfo{volume}{3}, \bibinfo{number}{1} (\bibinfo{year}{2024}).
\newblock


\bibitem[Orok et~al\mbox{.}(2024)]%
        {orok2024pharmacy}
\bibfield{author}{\bibinfo{person}{Edidiong Orok}, \bibinfo{person}{Chidera Okaramee}, \bibinfo{person}{Bethel Egboro}, \bibinfo{person}{Esther Egbochukwu}, \bibinfo{person}{Khairat Bello}, \bibinfo{person}{Samuel Etukudo}, \bibinfo{person}{Mark-Solomon Ogologo}, \bibinfo{person}{Precious Onyeka}, \bibinfo{person}{Obinna Etukokwu}, \bibinfo{person}{Mesileya Kolawole}, {et~al\mbox{.}}} \bibinfo{year}{2024}\natexlab{}.
\newblock \showarticletitle{Pharmacy students’ perception and knowledge of chat-based artificial intelligence tools at a Nigerian University}.
\newblock \bibinfo{journal}{\emph{BMC Medical Education}} \bibinfo{volume}{24}, \bibinfo{number}{1} (\bibinfo{year}{2024}), \bibinfo{pages}{1237}.
\newblock


\bibitem[Ortega et~al\mbox{.}(2021)]%
        {ortega2021love}
\bibfield{author}{\bibinfo{person}{John~E Ortega}, \bibinfo{person}{Richard Alexander~Castro Mamani}, {and} \bibinfo{person}{Jaime Rafael~Montoya Samame}.} \bibinfo{year}{2021}\natexlab{}.
\newblock \showarticletitle{Love thy neighbor: combining two neighboring low-resource languages for translation}. In \bibinfo{booktitle}{\emph{Proceedings of the 4th Workshop on Technologies for MT of Low Resource Languages (LoResMT2021)}}. \bibinfo{pages}{44--51}.
\newblock


\bibitem[Ousidhoum et~al\mbox{.}(2024)]%
        {ousidhoum-etal-2024-semeval}
\bibfield{author}{\bibinfo{person}{Nedjma Ousidhoum}, \bibinfo{person}{Shamsuddeen~Hassan Muhammad}, \bibinfo{person}{Mohamed Abdalla}, \bibinfo{person}{Idris Abdulmumin}, \bibinfo{person}{Ibrahim~Said Ahmad}, \bibinfo{person}{Sanchit Ahuja}, \bibinfo{person}{Alham~Fikri Aji}, \bibinfo{person}{Vladimir Araujo}, \bibinfo{person}{Meriem Beloucif}, \bibinfo{person}{Christine De~Kock}, \bibinfo{person}{Oumaima Hourrane}, \bibinfo{person}{Manish Shrivastava}, \bibinfo{person}{Thamar Solorio}, \bibinfo{person}{Nirmal Surange}, \bibinfo{person}{Krishnapriya Vishnubhotla}, \bibinfo{person}{Seid~Muhie Yimam}, {and} \bibinfo{person}{Saif~M. Mohammad}.} \bibinfo{year}{2024}\natexlab{}.
\newblock \showarticletitle{{S}em{E}val Task 1: Semantic Textual Relatedness for {A}frican and {A}sian Languages}. In \bibinfo{booktitle}{\emph{Proceedings of the 18th International Workshop on Semantic Evaluation (SemEval-2024)}}, \bibfield{editor}{\bibinfo{person}{Atul~Kr. Ojha}, \bibinfo{person}{A.~Seza Do\u~gru\"oz}, \bibinfo{person}{Harish Tayyar~Madabushi}, \bibinfo{person}{Giovanni Da~San~Martino}, \bibinfo{person}{Sara Rosenthal}, {and} \bibinfo{person}{Aiala Ros\'a}} (Eds.). \bibinfo{publisher}{Association for Computational Linguistics}, \bibinfo{address}{Mexico City, Mexico}, \bibinfo{pages}{1963--1978}.
\newblock
\href{https://doi.org/10.18653/v1/2024.semeval-1.272}{doi:\nolinkurl{10.18653/v1/2024.semeval-1.272}}


\bibitem[Oyekanmi et~al\mbox{.}(2013)]%
        {oyekanmi2013intelligent}
\bibfield{author}{\bibinfo{person}{Ezekiel~Olufunminiyi Oyekanmi}, \bibinfo{person}{Samuel~Adebayo Oluwadare}, {and} \bibinfo{person}{BK Alese}.} \bibinfo{year}{2013}\natexlab{}.
\newblock \showarticletitle{Intelligent system learning and understanding of Yor{\`u}b{\'a} language}.
\newblock \bibinfo{journal}{\emph{International Journal of Computer and Information Technology}} \bibinfo{volume}{2}, \bibinfo{number}{5} (\bibinfo{year}{2013}), \bibinfo{pages}{993--997}.
\newblock


\bibitem[Oyesanmi and Olukanmi(2024)]%
        {oyesanmi2024towards}
\bibfield{author}{\bibinfo{person}{Fiyinfoluwa Oyesanmi} {and} \bibinfo{person}{Peter Olukanmi}.} \bibinfo{year}{2024}\natexlab{}.
\newblock \showarticletitle{Towards Yoruba-Speaking Google Maps Navigation}.
\newblock  (\bibinfo{year}{2024}).
\newblock


\bibitem[Oyewusi et~al\mbox{.}(2021)]%
        {oyewusi2021naijaner}
\bibfield{author}{\bibinfo{person}{Wuraola~Fisayo Oyewusi}, \bibinfo{person}{Olubayo Adekanmbi}, \bibinfo{person}{Ifeoma Okoh}, \bibinfo{person}{Vitus Onuigwe}, \bibinfo{person}{Mary~Idera Salami}, \bibinfo{person}{Opeyemi Osakuade}, \bibinfo{person}{Sharon Ibejih}, {and} \bibinfo{person}{Usman~Abdullahi Musa}.} \bibinfo{year}{2021}\natexlab{}.
\newblock \showarticletitle{Naijaner: Comprehensive named entity recognition for 5 nigerian languages}.
\newblock \bibinfo{journal}{\emph{arXiv preprint arXiv:2105.00810}} (\bibinfo{year}{2021}).
\newblock


\bibitem[Oyinloye and Odejobi(2015)]%
        {oyinloye2015issues}
\bibfield{author}{\bibinfo{person}{Olufunke~A Oyinloye} {and} \bibinfo{person}{Odetunji~A Odejobi}.} \bibinfo{year}{2015}\natexlab{}.
\newblock \showarticletitle{Issues in Computational System for Morphological Analysis of Standard Yor{\`u}b{\'a} (SY) Verbs}. In \bibinfo{booktitle}{\emph{Proceedings of the World Congress on Engineering}}, Vol.~\bibinfo{volume}{1}.
\newblock


\bibitem[Parida et~al\mbox{.}(2023)]%
        {parida2023havqa}
\bibfield{author}{\bibinfo{person}{Shantipriya Parida}, \bibinfo{person}{Idris Abdulmumin}, \bibinfo{person}{Shamsuddeen~Hassan Muhammad}, \bibinfo{person}{Aneesh Bose}, \bibinfo{person}{Guneet~Singh Kohli}, \bibinfo{person}{Ibrahim~Said Ahmad}, \bibinfo{person}{Ketan Kotwal}, \bibinfo{person}{Sayan~Deb Sarkar}, \bibinfo{person}{Ond{\v{r}}ej Bojar}, {and} \bibinfo{person}{Habeebah~Adamu Kakudi}.} \bibinfo{year}{2023}\natexlab{}.
\newblock \showarticletitle{HaVQA: A Dataset for Visual Question Answering and Multimodal Research in Hausa Language}.
\newblock \bibinfo{journal}{\emph{arXiv preprint arXiv:2305.17690}} (\bibinfo{year}{2023}).
\newblock


\bibitem[Porter(1980)]%
        {porter1980algorithm}
\bibfield{author}{\bibinfo{person}{Martin~F Porter}.} \bibinfo{year}{1980}\natexlab{}.
\newblock \showarticletitle{An algorithm for suffix stripping}.
\newblock \bibinfo{journal}{\emph{Program}} \bibinfo{volume}{14}, \bibinfo{number}{3} (\bibinfo{year}{1980}), \bibinfo{pages}{130--137}.
\newblock


\bibitem[Rakhmanov and Schlippe(2022)]%
        {rakhmanov2022sentiment}
\bibfield{author}{\bibinfo{person}{Ochilbek Rakhmanov} {and} \bibinfo{person}{Tim Schlippe}.} \bibinfo{year}{2022}\natexlab{}.
\newblock \showarticletitle{Sentiment analysis for Hausa: Classifying students’ comments}. In \bibinfo{booktitle}{\emph{Proceedings of the 1st Annual Meeting of the ELRA/ISCA Special Interest Group on Under-Resourced Languages}}. \bibinfo{pages}{98--105}.
\newblock


\bibitem[Ramanathan et~al\mbox{.}(2023)]%
        {ramanathan2023techssn}
\bibfield{author}{\bibinfo{person}{Nishaanth Ramanathan}, \bibinfo{person}{Rajalakshmi Sivanaiah}, \bibinfo{person}{Mirnalinee Thanka~Nadar Thanagathai}, {et~al\mbox{.}}} \bibinfo{year}{2023}\natexlab{}.
\newblock \showarticletitle{TechSSN at SemEval-2023 Task 12: Monolingual Sentiment Classification in Hausa Tweets}. In \bibinfo{booktitle}{\emph{Proceedings of the 17th International Workshop on Semantic Evaluation (SemEval-2023)}}. \bibinfo{pages}{1190--1194}.
\newblock


\bibitem[Randell et~al\mbox{.}(1998)]%
        {randell1998hausar}
\bibfield{author}{\bibinfo{person}{R Randell}, \bibinfo{person}{A Bature}, {and} \bibinfo{person}{R Schuh}.} \bibinfo{year}{1998}\natexlab{}.
\newblock \bibinfo{title}{Hausar Baka - Gani Ya Kori Ji}.
\newblock


\bibitem[Raychawdhary et~al\mbox{.}(2025)]%
        {raychawdhary2025empowering}
\bibfield{author}{\bibinfo{person}{Nilanjana Raychawdhary}, \bibinfo{person}{Sutanu Bhattacharya}, \bibinfo{person}{Cheryl Seals}, {and} \bibinfo{person}{Gerry Dozier}.} \bibinfo{year}{2025}\natexlab{}.
\newblock \showarticletitle{Empowering Sentiment Analysis in African Low-Resource Languages through Transformer Models and Strategic Language Selection}.
\newblock \bibinfo{journal}{\emph{IEEE Access}} (\bibinfo{year}{2025}).
\newblock


\bibitem[Raychawdhary et~al\mbox{.}(2024a)]%
        {raychawdhary2024enhancing}
\bibfield{author}{\bibinfo{person}{Nilanjana Raychawdhary}, \bibinfo{person}{Amit Das}, \bibinfo{person}{Sutanu Bhattacharya}, \bibinfo{person}{Gerry Dozier}, {and} \bibinfo{person}{Cheryl~D Seals}.} \bibinfo{year}{2024}\natexlab{a}.
\newblock \showarticletitle{Enhancing Monolingual Sentiment Classification: Pioneering Strategies in Tailored Language Training and Analytical Techniques}. In \bibinfo{booktitle}{\emph{2024 IEEE 3rd International Conference on Computing and Machine Intelligence (ICMI)}}. IEEE, \bibinfo{pages}{1--5}.
\newblock


\bibitem[Raychawdhary et~al\mbox{.}(2024b)]%
        {raychawdhary2024optimizing}
\bibfield{author}{\bibinfo{person}{Nilanjana Raychawdhary}, \bibinfo{person}{Amit Das}, \bibinfo{person}{Sutanu Bhattacharya}, \bibinfo{person}{Gerry Dozier}, {and} \bibinfo{person}{Cheryl~D Seals}.} \bibinfo{year}{2024}\natexlab{b}.
\newblock \showarticletitle{Optimizing Multilingual Sentiment Analysis in Low-Resource Languages with Adaptive Pretraining and Strategic Language Selection}. In \bibinfo{booktitle}{\emph{2024 IEEE 3rd International Conference on Computing and Machine Intelligence (ICMI)}}. IEEE, \bibinfo{pages}{1--5}.
\newblock


\bibitem[Raychawdhary et~al\mbox{.}(2023a)]%
        {raychawdhary2023seals_lab}
\bibfield{author}{\bibinfo{person}{Nilanjana Raychawdhary}, \bibinfo{person}{Amit Das}, \bibinfo{person}{Gerry Dozier}, {and} \bibinfo{person}{Cheryl~D Seals}.} \bibinfo{year}{2023}\natexlab{a}.
\newblock \showarticletitle{Seals\_Lab at SemEval-2023 task 12: Sentiment analysis for low-resource African languages, Hausa and Igbo}. In \bibinfo{booktitle}{\emph{Proceedings of the 17th International Workshop on Semantic Evaluation (SemEval-2023)}}. \bibinfo{pages}{1508--1517}.
\newblock


\bibitem[Raychawdhary et~al\mbox{.}(2023b)]%
        {raychawdhary2023transformer}
\bibfield{author}{\bibinfo{person}{Nilanjana Raychawdhary}, \bibinfo{person}{Nathaniel Hughes}, \bibinfo{person}{Sutanu Bhattacharya}, \bibinfo{person}{Gerry Dozier}, {and} \bibinfo{person}{Cheryl~D Seals}.} \bibinfo{year}{2023}\natexlab{b}.
\newblock \showarticletitle{A Transformer-Based Language Model for Sentiment Classification and Cross-Linguistic Generalization: Empowering Low-Resource African Languages}. In \bibinfo{booktitle}{\emph{2023 IEEE International Conference on Artificial Intelligence, Blockchain, and Internet of Things (AIBThings)}}. IEEE, \bibinfo{pages}{1--5}.
\newblock


\bibitem[Remy et~al\mbox{.}(2024)]%
        {remy2024trans}
\bibfield{author}{\bibinfo{person}{Fran{\c{c}}ois Remy}, \bibinfo{person}{Pieter Delobelle}, \bibinfo{person}{Hayastan Avetisyan}, \bibinfo{person}{Alfiya Khabibullina}, \bibinfo{person}{Miryam de Lhoneux}, {and} \bibinfo{person}{Thomas Demeester}.} \bibinfo{year}{2024}\natexlab{}.
\newblock \showarticletitle{Trans-tokenization and cross-lingual vocabulary transfers: Language adaptation of LLMs for low-resource NLP}.
\newblock \bibinfo{journal}{\emph{arXiv preprint arXiv:2408.04303}} (\bibinfo{year}{2024}).
\newblock


\bibitem[Ren et~al\mbox{.}(2016)]%
        {ren2016automatic}
\bibfield{author}{\bibinfo{person}{Xiang Ren}, \bibinfo{person}{Ahmed El-Kishky}, \bibinfo{person}{Heng Ji}, {and} \bibinfo{person}{Jiawei Han}.} \bibinfo{year}{2016}\natexlab{}.
\newblock \showarticletitle{Automatic entity recognition and typing in massive text data}. In \bibinfo{booktitle}{\emph{Proceedings of the 2016 International Conference on Management of Data}}. \bibinfo{pages}{2235--2239}.
\newblock


\bibitem[Rhoda(2017)]%
        {rhoda2017computational}
\bibfield{author}{\bibinfo{person}{{\.I}yanda~Abimbola Rhoda}.} \bibinfo{year}{2017}\natexlab{}.
\newblock \showarticletitle{Computational analysis of Igbo numerals in a number-to-text conversion system}.
\newblock \bibinfo{journal}{\emph{Journal of Computer and Education Research}} \bibinfo{volume}{5}, \bibinfo{number}{10} (\bibinfo{year}{2017}), \bibinfo{pages}{241--254}.
\newblock


\bibitem[Rigdon(2017)]%
        {rigdon2017english}
\bibfield{author}{\bibinfo{person}{J Rigdon}.} \bibinfo{year}{2017}\natexlab{}.
\newblock \bibinfo{title}{English/Hausa Dictionary: Kamus na Hausa Zura Turanci}.
\newblock


\bibitem[Robertson and D{\'\i}az(2022)]%
        {robertson2022understanding}
\bibfield{author}{\bibinfo{person}{Samantha Robertson} {and} \bibinfo{person}{Mark D{\'\i}az}.} \bibinfo{year}{2022}\natexlab{}.
\newblock \showarticletitle{Understanding and Being Understood: User Strategies for Identifying and Recovering From Mistranslations in Machine Translation-Mediated Chat}. In \bibinfo{booktitle}{\emph{Proceedings of the 2022 ACM Conference on Fairness, Accountability, and Transparency}}. \bibinfo{pages}{2223--2238}.
\newblock


\bibitem[Safiriyu et~al\mbox{.}(2015)]%
        {eludiora2015development}
\bibfield{author}{\bibinfo{person}{I~Eludiora Safiriyu}, \bibinfo{person}{O~Agbeyangi Abayomi}, {and} \bibinfo{person}{O~I Fatusin}.} \bibinfo{year}{2015}\natexlab{}.
\newblock \showarticletitle{Development of an English to Yorùbá Machine Translation System for Yorùbá Verbs’ Tone Changing}.
\newblock \bibinfo{journal}{\emph{International Journal Computer Application, USA}} \bibinfo{volume}{129}, \bibinfo{number}{10} (\bibinfo{year}{2015}), \bibinfo{pages}{12--17}.
\newblock


\bibitem[Salahudeen et~al\mbox{.}(2024)]%
        {salahudeen2024hausanlp}
\bibfield{author}{\bibinfo{person}{Saheed~Abdullahi Salahudeen}, \bibinfo{person}{Falalu~Ibrahim Lawan}, \bibinfo{person}{Yusuf Aliyu}, \bibinfo{person}{Amina Abubakar}, \bibinfo{person}{Lukman Aliyu}, \bibinfo{person}{Nur Rabiu}, \bibinfo{person}{Mahmoud Ahmad}, \bibinfo{person}{Aliyu~Rabiu Shuaibu}, {and} \bibinfo{person}{Alamin Musa}.} \bibinfo{year}{2024}\natexlab{}.
\newblock \showarticletitle{Hausanlp at semeval-2024 task 1: Textual relatedness analysis for semantic representation of sentences}. In \bibinfo{booktitle}{\emph{Proceedings of the 18th International Workshop on Semantic Evaluation (SemEval-2024)}}. \bibinfo{pages}{188--192}.
\newblock


\bibitem[Salifou and Naroua(2014)]%
        {salifou2014design}
\bibfield{author}{\bibinfo{person}{Lawaly Salifou} {and} \bibinfo{person}{Harouna Naroua}.} \bibinfo{year}{2014}\natexlab{}.
\newblock \showarticletitle{Design of a spell corrector for Hausa language}.
\newblock \bibinfo{journal}{\emph{international Journal of computational Linguistics (IJCL)}} \bibinfo{volume}{5}, \bibinfo{number}{2} (\bibinfo{year}{2014}), \bibinfo{pages}{14--26}.
\newblock


\bibitem[Sani et~al\mbox{.}(2022)]%
        {sani2022sentiment}
\bibfield{author}{\bibinfo{person}{Muhammad Sani}, \bibinfo{person}{Abubakar Ahmad}, {and} \bibinfo{person}{Hadiza~S Abdulazeez}.} \bibinfo{year}{2022}\natexlab{}.
\newblock \showarticletitle{Sentiment Analysis of Hausa Language Tweet Using Machine Learning Approach}.
\newblock \bibinfo{journal}{\emph{Journal of Research in Applied Mathematics}} \bibinfo{volume}{8}, \bibinfo{number}{9} (\bibinfo{year}{2022}), \bibinfo{pages}{07--16}.
\newblock


\bibitem[Scannell(2011)]%
        {scannell2011statistical}
\bibfield{author}{\bibinfo{person}{Kevin~P Scannell}.} \bibinfo{year}{2011}\natexlab{}.
\newblock \showarticletitle{Statistical unicodification of African languages}.
\newblock \bibinfo{journal}{\emph{Language resources and evaluation}} \bibinfo{volume}{45}, \bibinfo{number}{3} (\bibinfo{year}{2011}), \bibinfo{pages}{375--386}.
\newblock


\bibitem[Shehu et~al\mbox{.}(2024)]%
        {shehu2024unveiling}
\bibfield{author}{\bibinfo{person}{Harisu~Abdullahi Shehu}, \bibinfo{person}{Kaloma~Usman Majikumna}, \bibinfo{person}{Aminu~Bashir Suleiman}, \bibinfo{person}{Stephen Luka}, \bibinfo{person}{Md~Haidar Sharif}, \bibinfo{person}{Rabie~A Ramadan}, {and} \bibinfo{person}{Huseyin Kusetogullari}.} \bibinfo{year}{2024}\natexlab{}.
\newblock \showarticletitle{Unveiling Sentiments: A Deep Dive into Sentiment Analysis for Low-Resource Languages--A Case Study on Hausa Texts}.
\newblock \bibinfo{journal}{\emph{IEEE Access}} (\bibinfo{year}{2024}).
\newblock


\bibitem[Shi et~al\mbox{.}(2022)]%
        {shi2022low}
\bibfield{author}{\bibinfo{person}{Shumin Shi}, \bibinfo{person}{Xing Wu}, \bibinfo{person}{Rihai Su}, {and} \bibinfo{person}{Heyan Huang}.} \bibinfo{year}{2022}\natexlab{}.
\newblock \showarticletitle{Low-resource neural machine translation: Methods and trends}.
\newblock \bibinfo{journal}{\emph{ACM Transactions on Asian and Low-Resource Language Information Processing}} \bibinfo{volume}{21}, \bibinfo{number}{5} (\bibinfo{year}{2022}), \bibinfo{pages}{1--22}.
\newblock


\bibitem[Shode et~al\mbox{.}(2022)]%
        {shode2022yosm}
\bibfield{author}{\bibinfo{person}{Iyanuoluwa Shode}, \bibinfo{person}{David~Ifeoluwa Adelani}, {and} \bibinfo{person}{Anna Feldman}.} \bibinfo{year}{2022}\natexlab{}.
\newblock \showarticletitle{yosm: A new yoruba sentiment corpus for movie reviews}.
\newblock \bibinfo{journal}{\emph{arXiv preprint arXiv:2204.09711}} (\bibinfo{year}{2022}).
\newblock


\bibitem[Sigurd(1980)]%
        {sigurd1980numbers}
\bibfield{author}{\bibinfo{person}{Bengt Sigurd}.} \bibinfo{year}{1980}\natexlab{}.
\newblock \showarticletitle{From numbers to numerals and vice versa}. In \bibinfo{booktitle}{\emph{Computational and mathematical linguistics: Proceedings of the International Conference on Computational Linguistics: Pisa, 27/viii-1/ix 1973: vol. II.-(Biblioteca dell'Archivum Romanicum. Serie II: Linguistica; 37)}}. LS Olschki, \bibinfo{pages}{429--455}.
\newblock


\bibitem[Soronnadi et~al\mbox{.}(2024)]%
        {soronnadienhancing}
\bibfield{author}{\bibinfo{person}{Anthony Soronnadi}, \bibinfo{person}{Olubayo Adekanmbi}, \bibinfo{person}{Chinazo Anebelundu}, {and} \bibinfo{person}{David~Ifeoluwa Adelani}.} \bibinfo{year}{2024}\natexlab{}.
\newblock \showarticletitle{Enhancing Transformer Models for Igbo Language Processing: A Critical Comparative Study}. In \bibinfo{booktitle}{\emph{5th Workshop on African Natural Language Processing}}.
\newblock


\bibitem[Srivastava et~al\mbox{.}(2022)]%
        {srivastava2022beyond}
\bibfield{author}{\bibinfo{person}{Aarohi Srivastava}, \bibinfo{person}{Abhinav Rastogi}, \bibinfo{person}{Abhishek Rao}, \bibinfo{person}{Abu Awal~Md Shoeb}, \bibinfo{person}{Abubakar Abid}, \bibinfo{person}{Adam Fisch}, \bibinfo{person}{Adam~R Brown}, \bibinfo{person}{Adam Santoro}, \bibinfo{person}{Aditya Gupta}, \bibinfo{person}{Adri{\`a} Garriga-Alonso}, {et~al\mbox{.}}} \bibinfo{year}{2022}\natexlab{}.
\newblock \showarticletitle{Beyond the imitation game: Quantifying and extrapolating the capabilities of language models}.
\newblock \bibinfo{journal}{\emph{arXiv preprint arXiv:2206.04615}} (\bibinfo{year}{2022}).
\newblock


\bibitem[Sunday et~al\mbox{.}(2020)]%
        {sundaydevelopment}
\bibfield{author}{\bibinfo{person}{Adewale~Olumide Sunday}, \bibinfo{person}{Agbonifo~Oluwatoyin Catherine}, {and} \bibinfo{person}{Olaniyan Julius}.} \bibinfo{year}{2020}\natexlab{}.
\newblock \showarticletitle{Development of Bi-Directional English To Yoruba Translator for Real-Time Mobile Chatting}.
\newblock \bibinfo{journal}{\emph{International Journal of Computational Linguistics (IJCL)}} (\bibinfo{year}{2020}).
\newblock


\bibitem[Teze and Nazaruka(2024)]%
        {teze2024future}
\bibfield{author}{\bibinfo{person}{Vitalijs Teze} {and} \bibinfo{person}{Erika Nazaruka}.} \bibinfo{year}{2024}\natexlab{}.
\newblock \showarticletitle{Future Directions in Defence NLP: Investigating Research Gaps for Low-Resource Languages}. In \bibinfo{booktitle}{\emph{International Baltic Conference on Digital Business and Intelligent Systems}}. Springer, \bibinfo{pages}{93--105}.
\newblock


\bibitem[Timothy et~al\mbox{.}(2024)]%
        {timothybilingual}
\bibfield{author}{\bibinfo{person}{Adeboje~Olawale Timothy}, \bibinfo{person}{Adetunmbi~Olusola Adebayo}, \bibinfo{person}{Arome~Gabriel Junior}, {and} \bibinfo{person}{Akinyede~Raphael Olufemi}.} \bibinfo{year}{2024}\natexlab{}.
\newblock \showarticletitle{Bilingual Neural Machine Translation From English To Yoruba Using A Transformer Model}.
\newblock \bibinfo{journal}{\emph{International Journal of Innovative Science and Research Technology}} (\bibinfo{year}{2024}).
\newblock


\bibitem[Tonneau et~al\mbox{.}(2024)]%
        {tonneau2024naijahate}
\bibfield{author}{\bibinfo{person}{Manuel Tonneau}, \bibinfo{person}{Pedro~Quinta De~Castro}, \bibinfo{person}{Karim Lasri}, \bibinfo{person}{Ibrahim Farouq}, \bibinfo{person}{Lakshmi Subramanian}, \bibinfo{person}{Victor Orozco-Olvera}, {and} \bibinfo{person}{Samuel Fraiberger}.} \bibinfo{year}{2024}\natexlab{}.
\newblock \showarticletitle{NaijaHate: Evaluating hate speech detection on Nigerian Twitter using representative data}. In \bibinfo{booktitle}{\emph{Proceedings of the 62nd Annual Meeting of the Association for Computational Linguistics (Volume 1: Long Papers)}}. \bibinfo{pages}{9020--9040}.
\newblock


\bibitem[Toyin et~al\mbox{.}(2024)]%
        {toyin2024hidden}
\bibfield{author}{\bibinfo{person}{Okebule Toyin}, \bibinfo{person}{Christianah~O Akinduyite}, {et~al\mbox{.}}} \bibinfo{year}{2024}\natexlab{}.
\newblock \showarticletitle{A Hidden Markov Model-Based Parts-of-Speech Tagger for Yoruba Language}. In \bibinfo{booktitle}{\emph{2024 International Conference on Science, Engineering and Business for Driving Sustainable Development Goals (SEB4SDG)}}. IEEE, \bibinfo{pages}{1--6}.
\newblock


\bibitem[Tran et~al\mbox{.}(2021)]%
        {tran2021facebook}
\bibfield{author}{\bibinfo{person}{Chau Tran}, \bibinfo{person}{Shruti Bhosale}, \bibinfo{person}{James Cross}, \bibinfo{person}{Philipp Koehn}, \bibinfo{person}{Sergey Edunov}, {and} \bibinfo{person}{Angela Fan}.} \bibinfo{year}{2021}\natexlab{}.
\newblock \showarticletitle{Facebook ai wmt21 news translation task submission}.
\newblock \bibinfo{journal}{\emph{arXiv preprint arXiv:2108.03265}} (\bibinfo{year}{2021}).
\newblock


\bibitem[Tresner-Kirsch et~al\mbox{.}(2023)]%
        {tresner2023intent}
\bibfield{author}{\bibinfo{person}{David Tresner-Kirsch}, \bibinfo{person}{Amanda~Azari Mikkelson}, \bibinfo{person}{Chika Yinka-Banjo}, \bibinfo{person}{Mary Akinyemi}, {and} \bibinfo{person}{Siddhartha Goyal}.} \bibinfo{year}{2023}\natexlab{}.
\newblock \showarticletitle{Intent Recognition on Low-Resource Language Messages in a Health Marketplace Chatbot}. In \bibinfo{booktitle}{\emph{2023 IEEE 11th International Conference on Healthcare Informatics (ICHI)}}. IEEE, \bibinfo{pages}{457--459}.
\newblock


\bibitem[Tukur et~al\mbox{.}(2024)]%
        {tukur2024towards}
\bibfield{author}{\bibinfo{person}{A Tukur}, \bibinfo{person}{A Jibrin}, {and} \bibinfo{person}{U Inuwa}.} \bibinfo{year}{2024}\natexlab{}.
\newblock \showarticletitle{TOWARDS EFFICIENT PART-OF-SPEECH TAGGING FOR THE KANURI LANGUAGE: A HIDDEN MARKOV MODEL-BASED SOLUTION}.
\newblock \bibinfo{journal}{\emph{Nigerian Journal of Engineering Science and Technology Research}} \bibinfo{volume}{10}, \bibinfo{number}{2} (\bibinfo{year}{2024}), \bibinfo{pages}{115--123}.
\newblock


\bibitem[Tukur et~al\mbox{.}(2019a)]%
        {tukurcorpus}
\bibfield{author}{\bibinfo{person}{Aminu Tukur}, \bibinfo{person}{Kabir Umar}, {and} \bibinfo{person}{Anas~Sa’idu Muhammad}.} \bibinfo{year}{2019}\natexlab{a}.
\newblock \showarticletitle{A Corpus-Based Approach to Parts-of-Speech Tagging of the Hausa-Based Texts Using Hidden Markov and Unigram Models}.
\newblock \bibinfo{journal}{\emph{Algaita}} (\bibinfo{year}{2019}).
\newblock


\bibitem[Tukur et~al\mbox{.}(2019b)]%
        {tukur2019tagging}
\bibfield{author}{\bibinfo{person}{Aminu Tukur}, \bibinfo{person}{Kabir Umar}, {and} \bibinfo{person}{Anas~Saidu Muhammad}.} \bibinfo{year}{2019}\natexlab{b}.
\newblock \showarticletitle{Tagging part of speech in hausa sentences}. In \bibinfo{booktitle}{\emph{2019 15th International Conference on Electronics, Computer and Computation (ICECCO)}}. IEEE, \bibinfo{pages}{1--6}.
\newblock


\bibitem[Tukur et~al\mbox{.}(2020)]%
        {tukur2020parts}
\bibfield{author}{\bibinfo{person}{Aminu Tukur}, \bibinfo{person}{Kabir Umar}, {and} \bibinfo{person}{S Muhammad}.} \bibinfo{year}{2020}\natexlab{}.
\newblock \showarticletitle{Parts-of-speech tagging of Hausa-based texts using hidden Markov model}.
\newblock \bibinfo{journal}{\emph{Dutse Journal of Pure and Applied Sciences}}  \bibinfo{volume}{6} (\bibinfo{year}{2020}), \bibinfo{pages}{303--13}.
\newblock


\bibitem[T{\'u}nj{\'\i} and B{\'i}(2011)]%
        {tunji2011design}
\bibfield{author}{\bibinfo{person}{{\`A}j{\`a}d{\'\i} O~d{\'e} T{\'u}nj{\'\i}} {and} \bibinfo{person}{Od{\'e}jo B{\'i}}.} \bibinfo{year}{2011}\natexlab{}.
\newblock \showarticletitle{Design of a Text Markup System for Yor{\`u}b{\'a} Text-to-Speech Synthesis Applications}.
\newblock \bibinfo{journal}{\emph{Conference on Human Language Technology for Development}} (\bibinfo{year}{2011}).
\newblock


\bibitem[Turki et~al\mbox{.}(2024)]%
        {turki2024text}
\bibfield{author}{\bibinfo{person}{Houcemeddine Turki}, \bibinfo{person}{Naome~A Etori}, \bibinfo{person}{Mohamed Ali~Hadj Taieb}, \bibinfo{person}{Abdul-Hakeem Omotayo}, \bibinfo{person}{Chris~Chinenye Emezue}, \bibinfo{person}{Mohamed~Ben Aouicha}, \bibinfo{person}{Ayodele Awokoya}, \bibinfo{person}{Falalu~Ibrahim Lawan}, {and} \bibinfo{person}{Doreen Nixdorf}.} \bibinfo{year}{2024}\natexlab{}.
\newblock \showarticletitle{Text Categorization Can Enhance Domain-Agnostic Stopword Extraction}.
\newblock \bibinfo{journal}{\emph{arXiv preprint arXiv:2401.13398}} (\bibinfo{year}{2024}).
\newblock


\bibitem[UCLA({[n.\,d.]})]%
        {hausapoetry}
\bibfield{author}{\bibinfo{person}{UCLA}.} \bibinfo{year}{[n.\,d.]}\natexlab{}.
\newblock \showarticletitle{Hausa Poetry and Songs - An Introduction to Hausa Poetry and Song}.
\newblock  (\bibinfo{year}{[n.\,d.]}).
\newblock


\bibitem[Udeze and Voigt(2026)]%
        {udeze2026language}
\bibfield{author}{\bibinfo{person}{Nneoma~C Udeze} {and} \bibinfo{person}{Rob Voigt}.} \bibinfo{year}{2026}\natexlab{}.
\newblock \showarticletitle{Language Choice in Nigerian Social Media Hate Speech}. In \bibinfo{booktitle}{\emph{Proceedings of the 7th Workshop on African Natural Language Processing (AfricaNLP 2026)}}. \bibinfo{pages}{88--102}.
\newblock


\bibitem[Ugwu et~al\mbox{.}(2024)]%
        {ugwu2024part}
\bibfield{author}{\bibinfo{person}{Chukwuemeka~Christian Ugwu}, \bibinfo{person}{Abisola~Rukayat Oyewole}, \bibinfo{person}{Olugbemiga~Solomon Popoola}, \bibinfo{person}{Adebayo~Olusola Adetunmbi}, {and} \bibinfo{person}{Ayo Elebute}.} \bibinfo{year}{2024}\natexlab{}.
\newblock \showarticletitle{A part of speech tagger for Yoruba language text using deep neural network}.
\newblock \bibinfo{journal}{\emph{Franklin Open}}  \bibinfo{volume}{9} (\bibinfo{year}{2024}), \bibinfo{pages}{100185}.
\newblock


\bibitem[Umar(2021)]%
        {umaraspects}
\bibfield{author}{\bibinfo{person}{Ali~Usman Umar}.} \bibinfo{year}{2021}\natexlab{}.
\newblock \showarticletitle{Aspects of Meaning by Collocation and the Degree of Collocability in Hausa Recurrent Word Combinations}.
\newblock  (\bibinfo{year}{2021}).
\newblock


\bibitem[Usip et~al\mbox{.}(2023)]%
        {usip2023text}
\bibfield{author}{\bibinfo{person}{Patience~U Usip}, \bibinfo{person}{Funebi~F Ijebu}, \bibinfo{person}{Ifiok~J Udo}, {and} \bibinfo{person}{Ikechukwu~K Ollawa}.} \bibinfo{year}{2023}\natexlab{}.
\newblock \showarticletitle{Text-Based Emergency Alert Framework for Under-Resourced Languages in Southern Nigeria}.
\newblock In \bibinfo{booktitle}{\emph{Semantic AI in Knowledge Graphs}}. \bibinfo{publisher}{CRC Press}, \bibinfo{pages}{111--126}.
\newblock


\bibitem[Van~Valin and Foley(1980)]%
        {van1980role}
\bibfield{author}{\bibinfo{person}{Robert~D Van~Valin} {and} \bibinfo{person}{William~A Foley}.} \bibinfo{year}{1980}\natexlab{}.
\newblock \showarticletitle{Role and reference grammar}.
\newblock In \bibinfo{booktitle}{\emph{Current approaches to syntax}}. \bibinfo{publisher}{Brill}, \bibinfo{pages}{329--352}.
\newblock


\bibitem[Varab and Schluter(2021)]%
        {varab2021massivesumm}
\bibfield{author}{\bibinfo{person}{Daniel Varab} {and} \bibinfo{person}{Natalie Schluter}.} \bibinfo{year}{2021}\natexlab{}.
\newblock \showarticletitle{MassiveSumm: a very large-scale, very multilingual, news summarisation dataset}. In \bibinfo{booktitle}{\emph{Proceedings of the 2021 Conference on Empirical Methods in Natural Language Processing}}. \bibinfo{pages}{10150--10161}.
\newblock


\bibitem[Vargas et~al\mbox{.}(2024)]%
        {vargas2024hausahate}
\bibfield{author}{\bibinfo{person}{Francielle~Alves Vargas}, \bibinfo{person}{Samuel Guimar{\~a}es}, \bibinfo{person}{Shamsuddeen~Hassan Muhammad}, \bibinfo{person}{Diego Alves}, \bibinfo{person}{Ibrahim~Said Ahmad}, \bibinfo{person}{Idris Abdulmumin}, \bibinfo{person}{Diallo Mohamed}, \bibinfo{person}{Thiago Alexandre~Salgueiro Pardo}, {and} \bibinfo{person}{Fabr{\'\i}cio Benevenuto}.} \bibinfo{year}{2024}\natexlab{}.
\newblock \showarticletitle{HausaHate: an expert annotated corpus for hausa hate speech detection}. In \bibinfo{booktitle}{\emph{Proceedings}}.
\newblock


\bibitem[Wali and Nisioi(2026)]%
        {wali2026automatic}
\bibfield{author}{\bibinfo{person}{Ahmad~Mustapha Wali} {and} \bibinfo{person}{Sergiu Nisioi}.} \bibinfo{year}{2026}\natexlab{}.
\newblock \showarticletitle{Automatic Correction of Writing Anomalies in Hausa Texts}. In \bibinfo{booktitle}{\emph{Proceedings of the 64th Annual Meeting of the Association for Computational Linguistics (Volume 1: Long Papers)}}. \bibinfo{pages}{9514--9528}.
\newblock


\bibitem[Wang et~al\mbox{.}(2023)]%
        {wang2023noise}
\bibfield{author}{\bibinfo{person}{Song Wang}, \bibinfo{person}{Zhen Tan}, \bibinfo{person}{Ruocheng Guo}, {and} \bibinfo{person}{Jundong Li}.} \bibinfo{year}{2023}\natexlab{}.
\newblock \showarticletitle{Noise-robust fine-tuning of pretrained language models via external guidance}.
\newblock \bibinfo{journal}{\emph{arXiv preprint arXiv:2311.01108}} (\bibinfo{year}{2023}).
\newblock


\bibitem[Wang et~al\mbox{.}(2024)]%
        {wang2024monolingual}
\bibfield{author}{\bibinfo{person}{Xinyu Wang}, \bibinfo{person}{Wenbo Zhang}, {and} \bibinfo{person}{Sarah Rajtmajer}.} \bibinfo{year}{2024}\natexlab{}.
\newblock \showarticletitle{Monolingual and Multilingual Misinformation Detection for Low-Resource Languages: A Comprehensive Survey}.
\newblock \bibinfo{journal}{\emph{arXiv preprint arXiv:2410.18390}} (\bibinfo{year}{2024}).
\newblock


\bibitem[Watt et~al\mbox{.}(2023)]%
        {watt2023edge}
\bibfield{author}{\bibinfo{person}{Tess Watt}, \bibinfo{person}{Christos Chrysoulas}, {and} \bibinfo{person}{Dimitra Gkatzia}.} \bibinfo{year}{2023}\natexlab{}.
\newblock \showarticletitle{Edge NLP for Efficient Machine Translation in Low Connectivity Areas}. In \bibinfo{booktitle}{\emph{2023 IEEE 9th World Forum on Internet of Things (WF-IoT)}}. IEEE, \bibinfo{pages}{1--6}.
\newblock


\bibitem[Wolff(2014)]%
        {britannica2014}
\bibfield{author}{\bibinfo{person}{H.~Ekkehard Wolff}.} \bibinfo{year}{2014}\natexlab{}.
\newblock \showarticletitle{Chadic Languages}.
\newblock  (\bibinfo{year}{2014}).
\newblock


\bibitem[Wu et~al\mbox{.}(2018)]%
        {wu2018starspace}
\bibfield{author}{\bibinfo{person}{Ledell Wu}, \bibinfo{person}{Adam Fisch}, \bibinfo{person}{Sumit Chopra}, \bibinfo{person}{Keith Adams}, \bibinfo{person}{Antoine Bordes}, {and} \bibinfo{person}{Jason Weston}.} \bibinfo{year}{2018}\natexlab{}.
\newblock \showarticletitle{Starspace: Embed all the things!}. In \bibinfo{booktitle}{\emph{Proceedings of the AAAI conference on artificial intelligence}}, Vol.~\bibinfo{volume}{32}.
\newblock


\bibitem[Yazar et~al\mbox{.}(2023)]%
        {yazar2023low}
\bibfield{author}{\bibinfo{person}{B{\i}lge~Ka{\u{g}}an Yazar}, \bibinfo{person}{Durmu{\c{s}}~{\"O}zkan {\c{S}}ah{\i}n}, {and} \bibinfo{person}{Erdal Kili{\c{c}}}.} \bibinfo{year}{2023}\natexlab{}.
\newblock \showarticletitle{Low-resource neural machine translation: A systematic literature review}.
\newblock \bibinfo{journal}{\emph{IEEE Access}}  \bibinfo{volume}{11} (\bibinfo{year}{2023}), \bibinfo{pages}{131775--131813}.
\newblock


\bibitem[Yousuf et~al\mbox{.}(2024)]%
        {yousuf2024improving}
\bibfield{author}{\bibinfo{person}{Oreen Yousuf}, \bibinfo{person}{Gongbo Tang}, {and} \bibinfo{person}{Zeying Jin}.} \bibinfo{year}{2024}\natexlab{}.
\newblock \showarticletitle{Improving BERTScore for Machine Translation Evaluation Through Contrastive Learning}.
\newblock \bibinfo{journal}{\emph{IEEE Access}} (\bibinfo{year}{2024}).
\newblock


\bibitem[Yusof et~al\mbox{.}(2013)]%
        {yusof2013review}
\bibfield{author}{\bibinfo{person}{Shahrul Azmi~Mohd Yusof}, \bibinfo{person}{Abdulwahab~Funsho Atanda}, {and} \bibinfo{person}{M Hariharan}.} \bibinfo{year}{2013}\natexlab{}.
\newblock \showarticletitle{A Review of Yor{\`u}b{\'a} Automatic Speech Recognition}. In \bibinfo{booktitle}{\emph{2013 IEEE 3rd International Conference on System Engineering and Technology}}. IEEE, \bibinfo{pages}{242--247}.
\newblock


\bibitem[Yusuf et~al\mbox{.}(2023)]%
        {yusuf2023fine}
\bibfield{author}{\bibinfo{person}{Aliyu Yusuf}, \bibinfo{person}{Aliza Sarlan}, \bibinfo{person}{Kamaluddeen~Usman Danyaro}, {and} \bibinfo{person}{Abdullahi Sani~BA Rahman}.} \bibinfo{year}{2023}\natexlab{}.
\newblock \showarticletitle{Fine-tuning Multilingual Transformers for Hausa-English Sentiment Analysis}. In \bibinfo{booktitle}{\emph{2023 13th International Conference on Information Technology in Asia (CITA)}}. IEEE, \bibinfo{pages}{13--18}.
\newblock


\bibitem[Yusuf et~al\mbox{.}(2024)]%
        {yusuf2024sentiment}
\bibfield{author}{\bibinfo{person}{Aliyu Yusuf}, \bibinfo{person}{Aliza Sarlan}, \bibinfo{person}{Kamaluddeen~Usman Danyaro}, \bibinfo{person}{Abdullahi Sani~BA Rahman}, {and} \bibinfo{person}{Mujaheed Abdullahi}.} \bibinfo{year}{2024}\natexlab{}.
\newblock \showarticletitle{Sentiment Analysis in Low-Resource Settings: A Comprehensive Review of Approaches, Languages, and Data Sources}.
\newblock \bibinfo{journal}{\emph{IEEE Access}} (\bibinfo{year}{2024}).
\newblock


\bibitem[Zakari et~al\mbox{.}(2021)]%
        {zakari2021systematic}
\bibfield{author}{\bibinfo{person}{Rufai~Yusuf Zakari}, \bibinfo{person}{Zaharaddeen~Karami Lawal}, {and} \bibinfo{person}{Idris Abdulmumin}.} \bibinfo{year}{2021}\natexlab{}.
\newblock \showarticletitle{A systematic literature review of hausa natural language processing}.
\newblock \bibinfo{journal}{\emph{International Journal of Computer and Information Technology (2279-0764)}} \bibinfo{volume}{10}, \bibinfo{number}{4} (\bibinfo{year}{2021}).
\newblock


\bibitem[Zandam et~al\mbox{.}(2023)]%
        {zandam2023online}
\bibfield{author}{\bibinfo{person}{Abubakar~Yakubu Zandam}, \bibinfo{person}{Fatima~Adam Muhammad}, {and} \bibinfo{person}{Isa Inuwa-Dutse}.} \bibinfo{year}{2023}\natexlab{}.
\newblock \showarticletitle{Online threats detection in hausa language}. In \bibinfo{booktitle}{\emph{4th Workshop on African Natural Language Processing}}.
\newblock


\bibitem[Zhang et~al\mbox{.}(2017)]%
        {zhang2017embracing}
\bibfield{author}{\bibinfo{person}{Boliang Zhang}, \bibinfo{person}{Di Lu}, \bibinfo{person}{Xiaoman Pan}, \bibinfo{person}{Ying Lin}, \bibinfo{person}{Halidanmu Abudukelimu}, \bibinfo{person}{Heng Ji}, {and} \bibinfo{person}{Kevin Knight}.} \bibinfo{year}{2017}\natexlab{}.
\newblock \showarticletitle{Embracing non-traditional linguistic resources for low-resource language name tagging}. In \bibinfo{booktitle}{\emph{Proceedings of the Eighth International Joint Conference on Natural Language Processing (Volume 1: Long Papers)}}. \bibinfo{pages}{362--372}.
\newblock


\bibitem[Zhou et~al\mbox{.}(2024)]%
        {zhou2024character}
\bibfield{author}{\bibinfo{person}{Rui Zhou}, \bibinfo{person}{Akinori Ito}, {and} \bibinfo{person}{Takashi Nose}.} \bibinfo{year}{2024}\natexlab{}.
\newblock \showarticletitle{Character Expressions in Meta-Learning for Extremely Low Resource Language Speech Recognition}. In \bibinfo{booktitle}{\emph{Proceedings of the 2024 16th International Conference on Machine Learning and Computing}}. \bibinfo{pages}{525--529}.
\newblock


\end{thebibliography}

\appendix

\begin{table}[h!]
    \centering
    \renewcommand{\arraystretch}{1.5}
    \resizebox{\textwidth}{!}{ 
    \begin{tabular}{@{}p{4cm}p{12cm}@{}}
        \toprule
        \textbf{Category} & \textbf{Search Terms} \\ 
        \midrule
        \textbf{Language-Related Terms} & 
        \begin{itemize}
            \item[-] \texttt{"Hausa" OR "Hausa Language"}
            \item[-] \texttt{"Igbo" OR "Igbo Language"}
            \item[-] \texttt{"Yorùbá" OR "Yorùbá Language"}
        \end{itemize} \\ 
        
        \textbf{NLP-Related Terms} & 
        \begin{itemize}
            \item[-] \texttt{"natural language processing" OR "deep learning" OR "machine learning" OR "artificial intelligence"}
            \item[-] \texttt{"low-resource language" OR "low-resource languages" OR "low-resource NLP"}
            \item[-] \texttt{"sentiment analysis" OR "machine translation" OR "part-of-speech tagging" OR "named entity recognition"}
            \item[-] \texttt{"automatic speech recognition" OR "text-to-speech" OR "speech recognition"}
        \end{itemize} \\ 
        
        \textbf{Combined Search Queries} & 
        \begin{itemize}
            \item[-] \texttt{("Hausa" OR "Hausa Language") AND ("natural language processing" OR "deep learning" OR "machine learning" OR "artificial intelligence")}
            \item[-] \texttt{("Hausa" OR "Hausa Language") AND ("low-resource languages" OR "low-resource languages" OR "low-resource NLP")}
            \item[-] \texttt{("Yorùbá" OR "Yoruba Language") AND ("natural language processing" OR "deep learning" OR "machine learning" OR "artificial intelligence")}
            \item[-] \texttt{("Yorùbá" OR "Yoruba Language") AND ("low-resource languages" OR "low-resource languages" OR "low-resource NLP")}
            \item[-] \texttt{("Igbo" OR "Igbo Language") AND ("natural language processing" OR "deep learning" OR "machine learning" OR "artificial intelligence")}
            \item[-] \texttt{("Igbo" OR "Igbo Language") AND ("low-resource languages" OR "low-resource languages" OR "low-resource NLP")}
        \end{itemize} \\ 
        \bottomrule
    \end{tabular}
    } 
    \caption{An example of the systematic search strategy for NaijaNLP review. Terms were structured to address (1) language-specific terms (Hausa, Igbo, Yorùbá), (2) NLP-related, and (3) example of the combined integrated search queries.} 
    \label{tab:search-terms}
\end{table}

\end{document}